\documentclass[
  journal=large,
  manuscript= Research Article
]{cup-journal}

\usepackage[dvipsnames]{xcolor}
\usepackage{amsmath}
\usepackage[nopatch]{microtype}
\usepackage{booktabs}
\usepackage{longtable}
\usepackage{graphicx}
\usepackage{subcaption}
\usepackage{multicol}
\usepackage{multirow}
\usepackage{float}
\usepackage{array}
\usepackage{colortbl}
\usepackage{nicematrix}
\DeclareUnicodeCharacter{202F}{\,}
\emergencystretch=\maxdimen
\title{A Novel Mathematical Framework for Objective Characterization of Ideas}

\author{B. Sankar}
\affiliation{Department of Mechanical Engineering, Indian Institute of Science (IISc), Bangalore, India - 560012}
\email[B. Sankar]{sankarb@iisc.ac.in}

\author{Dibakar Sen}
\affiliation{Department of Design and Manufacturing (erstwhile CPDM), Indian Institute of Science (IISc), Bangalore, India - 560012}


\begin{document}

\begin{abstract}
The demand for innovation in product design necessitates a prolific ideation phase. Conversational AI (CAI) systems that use Large Language Models (LLMs) such as GPT (Generative Pre-trained Transformer) have been shown to be fruitful in augmenting human creativity, providing numerous novel and diverse ideas. Despite the success in ideation quantity, the qualitative assessment of these ideas remains challenging and traditionally reliant on expert human evaluation. This method suffers from limitations such as human judgment errors, bias, and oversight. Addressing this gap, our study introduces a comprehensive mathematical framework for automated analysis to objectively evaluate the plethora of ideas generated by CAI systems and/or humans. This framework is particularly advantageous for novice designers who lack experience in selecting promising ideas. By converting the ideas into higher dimensional vectors and quantitatively measuring the diversity between them using tools such as UMAP, DBSCAN and PCA, the proposed method provides a reliable and objective way of selecting the most promising ideas, thereby enhancing the efficiency of the ideation phase.
\end{abstract}

\section*{Introduction}
The advent of Conversational AI (CAI) technologies, particularly large language models (LLMs) like GPT, Llama, Gemini, etc., has revolutionized the field of natural language processing~(\cite{yuhan-etal-2023-unleashing}). These tools offer unprecedented capabilities in generating coherent and contextually relevant text, facilitating new forms of co-creation and collaboration in both individual and group settings~(\cite{meyer2023a}). In recent times, since its major outbreak, these models have found their way into several applications, such as Content Generation, Customer Service, Education, Healthcare, Research and Development, Entertainment, Translation and Localization, Personal Assistants, etc.~(\cite{Desmond_Ashktorab_Pan_Dugan_Johnson_2024}). Among these applications, the ability to generate new human-like content holds significant value in various domains. One such field that stands to benefit greatly is product design~(\cite{Gonzalez_Moran_Houde_He_Ross_Muller_Kunde_Weisz}). Particularly during the conceptual design phase, designers are tasked with generating ideas—a process that involves the creation of new content~(\cite{Liu_Iter_Xu_Wang_Xu_Zhu_2023}). Recently, the application of Conversational Artificial Intelligence (CAI) models in this context has garnered increasing attention~(\cite{Shaer_Cooper_Kun_Mokryn}). The authors of this paper have also previously published their research on utilizing CAI as a tool for idea generation in product design.

Conversational AI models have significant implications for the creative content generation process, encompassing both the divergence stage of idea generation and the convergence stage of evaluation and selection of ideas\cite{shaer2024aiaugmented}. During the divergence stage, LLMs can generate numerous ideas rapidly, providing a rich pool of options for further refinement and selection. This capability is particularly advantageous in brainstorming sessions, where the goal is to produce a wide array of ideas without immediate judgment~(\cite{Shaer_Cooper_Kun_Mokryn}). The increasing availability of LLMs, such as GPT, has enabled widespread adoption in various domains, including creative writing, design, and problem-solving~(\cite{Shaer_Cooper_Mokryn_Kun_Shoshan_2024}). These models can generate text that is often indistinguishable from human-written content, making them valuable tools for augmenting human creativity \cite{shaer2024aiaugmented} during the ideation process. However, the integration of LLMs into the creative process also necessitates robust evaluation mechanisms to ensure the quality and relevance of the generated ideas.

\CUPTWOCOL 

A crucial aspect of using LLMs in idea generation is distinguishing between AI-generated and human-generated content. Participants in various studies have emphasized the importance of clearly identifying AI contributions, often using visual cues like icons and outlines \cite{gonzalez-a}. This distinction not only aids in assigning credit but also helps maintain accountability and ethical considerations in collaborative settings. For instance, capturing the prompt that resulted in the AI output and acknowledging the user who wrote the prompt can provide transparency and accountability in the ideation process.

The ability to generate a diverse range of ideas is vital for effective ideation. Participants have recognized the need for generating unexpected or provocative ideas to stimulate creative thinking \cite{gonzalez-a}. Novelty and diversity are essential dimensions of idea quality, as they contribute to the originality and uniqueness of the generated ideas~(\cite{Fiorineschi_Rotini_2023}). While traditional tools may lack controls for low-level generative parameters, modern LLMs can be prompted to evaluate their own ideas for relevance, thereby filtering out less pertinent suggestions and enhancing the overall quality of the ideation process. For example, adjusting the temperature parameter in LLMs can control the randomness and creativity of the generated ideas. Higher temperatures result in more diverse and creative outputs, while lower temperatures produce more focused and predictable ideas.

Group brainstorming sessions often face barriers such as peer judgment, free riding, and production blocking, which can limit the effectiveness of the ideation process \cite{shaer2024aiaugmented}. Peer judgment refers to the influence of group members' opinions on individual contributions, which can discourage the sharing of unconventional ideas. Free riding occurs when some group members rely on others to contribute, leading to reduced individual participation. Production blocking happens when group members wait for their turn to share ideas, resulting in lost opportunities for spontaneous ideation. Online visual workspaces and the integration of LLMs into these platforms offer new avenues for enhancing group creativity by providing diverse perspectives and reducing evaluation apprehension \cite{Desmond2024}. While holistic measures can be efficient, they often conflate multiple constructs, leading to inconsistencies in ratings. Therefore, a more granular approach is necessary to capture the specific aspects of idea quality systematically. For example, using a well-defined set of criteria can provide a structured and consistent evaluation framework.

Evaluating the quality of ideas is a critical aspect of the creative process. Traditional methods for idea evaluation can be broadly categorized into subjective and objective evaluations.

Subjective evaluations have traditionally been the cornerstone of idea assessment. These evaluations often involve human judges who rate the quality of ideas based on personal judgment, which can be influenced by individual biases and inconsistencies~(\cite{Ben_2010}). While subjective evaluations can capture nuanced insights, they are inherently limited by their lack of scalability and potential for variability \cite{Dean2006}. For example, one rater may intuitively include novelty or workability in their evaluation, while another may not, leading to different ratings. Moreover, a single rater may be inconsistent across ideas because different constructs may seem more important to some ideas than others~(\cite{Boudier_2023}). Thus, despite their efficiency, holistic measures do not address specific evaluation components in a predictable way.

Several researchers~(\cite{Fiorineschi_Rotini_2023, Kim_Maher_2023, Christensen_Ball_2016, Nelson_Wilson_Rosen_Yen_2009, Linsey_Clauss_Kurtoglu_Murphy_Wood_Markman_2011, Bryant_2005, PUCCIO2012189, Kurtoglu_Campbell_Linsey_2009, Karimi_Maher_Davis_Grace_2019, Han_Shi_Chen_Childs_2018}) have reported multiple dimensions for the subjective evaluation of ideas, each capturing a specific aspect of idea quality. Different metrics have been reported in the literature for assessing ideas, which can be categorized into \textit{individual quality}, \textit{population quality}, and \textit{relevance} depending upon the specific aspects they aim to measure. 
\begin{description}
    \item Individual Quality: Assessing the quality of a given idea with respect to a known set of ideas
    \begin{enumerate}
        \item Novelty~(\cite{Shah2003, Fiorineschi_Rotini_2023}): The degree to which the idea is new and not derivative.
        \item Originality~(\cite{Christensen_Ball_2016, PUCCIO2012189}): The novelty of the idea.
        \item Creativity~(\cite{Altshuller1984, Karimi_Maher_Davis_Grace_2019}): The degree of imagination and inventiveness in the idea.
        \item Innovativeness~(\cite{Mirabito2021, Bryant_2005}): The degree of breakthrough thinking in the idea.
        \item Insightfulness~(\cite{Plucker2010}): The depth of understanding and insight reflected in the idea.
        \item Rarity~(\cite{Milan2016, Kurtoglu_Campbell_Linsey_2009}): The uniqueness of the idea compared to existing ideas.
    \end{enumerate}

    \item Population Quality: Assessing the quality of a given set of ideas as a whole
    \begin{enumerate}
        \item Diversity~(\cite{Nelson_Wilson_Rosen_Yen_2009, Fiorineschi_Rotini_2023}): The variety of ideas explored.
        \item Quantity (Fluency)~(\cite{Kim_Maher_2023, Plucker2010}): The number of ideas generated within a given time.
    \end{enumerate}

    \item Relevance: Assessing the quality of a given idea for solving the problem at hand
    \begin{enumerate}
        \item Usefulness~(\cite{Christensen_Ball_2016}): The practical applicability of the idea.
        \item Workability~(\cite{PUCCIO2012189}): The implementability of the idea.
        \item Thoroughness (Specificity)~(\cite{PUCCIO2012189}): The level of detail in the idea.
        \item Feasibility~(\cite{Cheeley2018, Srivathsavai2010}): The degree to which the idea can be implemented.
        \item Effectiveness~(\cite{Cheeley2018}): The potential impact and success of the idea.
        \item Impact~(\cite{Milan2016}): The potential effect and influence of the idea on solving the problem.
        \item Utility~(\cite{Han_Shi_Chen_Childs_2018}): The usefulness of the idea in solving the problem.
        \item Practicality~(\cite{Linsey_Clauss_Kurtoglu_Murphy_Wood_Markman_2011}): The ease of implementation of the idea with existing technologies.
    \end{enumerate}    
\end{description}

Objective evaluations aim to mitigate the limitations of subjective assessments by employing systematic and quantifiable metrics. Traditional automatic metrics like BLEU, ROUGE, and METEOR have been used to evaluate natural language generation systems~(\cite{blagec2022globalanalysismetricsused}), but these metrics often show a low correlation with human judgments~(\cite{liu2023geval}). To the best of the author's knowledge, there are no objective evaluation metrics similar to subjective evaluation metrics that align with human judgment. Hence, one of the novel contributions of this research work involves developing certain dimensional metrics that correlate with human judgement.

The use of AI for evaluating ideas promises increased speed and objectivity. Various types of AI evaluation systems have been explored by several researchers, each offering unique advantages and challenges. Table~\ref{tab:ai_text_analysis_systems} summarizes their application, evaluation criteria and their input and output format.
\begin{table*}[t!]
    \centering
    \resizebox{\textwidth}{!}{
    \begin{tabular}{|m{0.5cm}|m{2cm}|m{4cm}|m{3cm}|m{4cm}|}
        \hline
        \textbf{S. No.} & \centering \textbf{Name of the System} & \centering \textbf{Description} & \centering \textbf{Evaluation Criteria} & \textbf{Input and Output} \\
        \hline       
        1 & Task-Specific Evaluators & Measure the quality of a given NLP task such as summarization, translation, report generation, etc. & Accuracy, Relevance, Fluency, Adequacy, Clarity, Coherence & Input: Reference Text and Query Text.  \newline   Output: BLEU, METEOR, ROUGE, F1 and Exact Match Scores\\
        \hline
        2 & Unified Evaluators & Measure the quality of multiple NLP tasks in the same system. & Coherence, Fluency, Relevance, Clarity, Consistency & Input: Reference Text and Query Text. \newline  Output: BERTScore\\
        \hline
        3 & LLM-based Evaluators & Leverage LLMs' inherent capability to assess the quality of text. & Arbitrary as per the interpretation of LLM, e.g. Coherence, Relevance, etc. & Input: Set of Query texts.  \newline   Output: Aribtrary as per user query\\
        \hline
        4 & Embedding-based Metrics & Use vector representations of the text based on similarity & Word Mover Distance & Input: Set of texts.  \newline  Output: Cosine similarity, Euclidean distance\\
        \hline
    \end{tabular}
    } 
    \caption{Representative List of AI-based Text Analysis Systems}
    \label{tab:ai_text_analysis_systems}
\end{table*}

Despite the promise of AI-based evaluation systems, challenges like prompt sensitivity, bias, and hallucinations remain. Ensuring robust and transparent evaluation processes, integrating human oversight, and validating evaluation criteria are essential for enhancing the reliability and validity of AI-based evaluations.

The rapid advancements in artificial intelligence (AI) and natural language processing (NLP) have ushered in a new era of innovation, particularly in the realm of idea generation. Conversational AI (CAI) systems, powered by large language models (LLMs) such as GPT, have been shown to demonstrate unprecedented capabilities in generating numerous ideas within a short period. This transformative potential of CAI systems necessitates the development of automated, objective methods for evaluating the vast quantity of ideas they produce.

In traditional creative processes, human designers generate ideas through brainstorming sessions, individual reflection, and collaborative discussions. The volume of ideas produced in these settings is typically manageable, allowing human experts to evaluate and rate each idea subjectively. This manual evaluation process, although time-consuming, benefits from the nuanced judgment and contextual understanding of human evaluators. However, the advent of CAI systems has fundamentally altered the landscape of idea generation. These systems can produce an overwhelming number of ideas in a fraction of the time it would take human designers. While this capability significantly enhances the creative process by providing a rich pool of potential solutions, it also introduces the challenge of idea abundance. The sheer volume of ideas generated by CAI systems renders manual evaluation impractical and inefficient.

Given the impracticality of manual evaluation, there is a pressing need for automated methods to assess and evaluate the ideas generated by CAI systems. Such methods must be objective to ensure consistency, scalability, and reliability in the evaluation process. Objective evaluation is particularly crucial in scenarios where the ideas generated are used to inform critical decision-making processes, such as product design, strategic planning, and innovation management. However, in the absence of any objective framework to assess the quality and relevance of machine-generated ideas, both the usefulness and trustworthiness of this system are still to be adjudged through the scrutiny of human experts.

The primary motivation for this paper is to assess the quality of ideas and characteristics of idea exploration during the ideation phase of product design. Therefore, it is important to explore the nature of the distribution of a proposed set of ideas in the idea space. Human experts or a computational tool mentioned above can be used to assess the quality of individual ideas for novelty, usefulness, etc. However, there is no work available in the literature that enables an objective measure of the distribution of ideas in the idea space for characterizing the quality of an ideation exercise. This paper proposes a systematic mathematical framework using vector embeddings for the objective evaluation of the ideas and the ideation landscape. This paper proposes a systematic mathematical framework for the objective evaluation of the ideation process. We can apply quantitative metrics to assess their quality, relevance, and diversity by representing ideas as mathematical entities, such as vectors in high-dimensional space. This approach enables the systematic evaluation of ideas based on clearly defined measures, eliminating the subjectivity and variability inherent in manual evaluations.

Embedding-based methods are inherently statistical, necessitating a large set of sample ideas to ensure robust and meaningful analysis. However, obtaining such extensive datasets from human participants often results in a loss of uniformity and a limited number of ideas, complicating systematic analysis. To address this challenge, we leveraged a CAI-based ideation tool, as previously published in our work, which facilitates the generation of a large set of ideas within a specific structured framework. This tool enabled us to generate the necessary sample ideas for the development and validation of our evaluation method. For the sake of completeness, the following section briefly describes the CAI-based generation process, providing context for its role in our study.

\hfill \hrule


\section{Background}
\label{sec:background}

\subsection{Subjective Methods for Idea Assessment}
\label{subsec:subjective_methods}
Subjective methods for evaluating ideas primarily involve human judgment to rate or assess the quality of ideas. These methods rely heavily on the subjective perceptions of experts or non-expert judges to assess ideas based on aspects such as creativity, originality, novelty, and feasibility. One of the most prevalent subjective approaches is the Consensual Assessment Technique (CAT), initially proposed by Amabile (\cite{Amabile1982}). This method involves a panel of expert judges independently rating the creativity of each idea according to their implicit understanding and consensus. This approach has been considered the gold standard for creativity assessment in diverse domains, due to its wide acceptance among experts~(\cite{Beaty2021, Ceh2021, Cseh2019}).

Another prominent subjective method is the Subjective Top-Scoring method, including the Top-2 scoring approach, which involves designers themselves selecting their best or most creative ideas for subsequent evaluation by judges. Silvia et al.(\cite{Silvia2008}) introduced this participant-assisted scoring technique to minimize evaluation load on judges and improve inter-rater reliability~(\cite{Benedek2013}). Similarly, the Snapshot scoring approach has been employed to efficiently manage large idea pools by limiting raters' cognitive loads and focusing on a smaller subset of ideas~(\cite{Silvia2009, Benedek2013}).

Various alternative subjective methods have also been documented. For instance, the Multi-Voting Method has been popular among practitioners for screening large idea sets into more manageable subsets for deeper evaluation~(\cite{Kudrowitz2013}). Similarly, methods such as the Pugh Chart, Quality Function Deployment (QFD), Pahl and Beitz Utility Theory, and NUF (Novelty, Usefulness, Feasibility) tests have commonly been utilized in design practice. These methods typically rely on structured human judgments and are generally suitable for relatively smaller pools of ideas or concepts, often involving iterative rounds of evaluation and discussion~(\cite{Kudrowitz2013}).

Additionally, the creativity research literature also extensively uses psychometric instruments and subjective rating scales. Examples include the Creativity Achievement Questionnaire (CAQ) and various self-report inventories that measure creative achievements and creative personality attributes~(\cite{Carson2005}). Ratings by experts, non-experts, and self-report methods are frequently applied to provide subjective assessments of creative performance or potential~(\cite{Carson2005}). For example, the Creative Thinking-Drawing Production (TCT-DP) task assesses creativity based on human judgments of drawings, typically relying on human-coded scores assigned according to predefined criteria~(\cite{Cropley11042024}).

Despite variations, a common thread across subjective methods is their reliance on human perception, judgment, and often implicit criteria, which, while rich in interpretative depth, inherently introduce subjectivity and variability into the outcomes~(\cite{Ceh2021, Forthmann2017}).

\subsection{Limitations of Existing Subjective Methods}
While subjective methods have been foundational and widely used in creativity and idea assessment, several limitations have consistently been identified in the literature as follows:

Firstly, the fundamental limitation pertains to \textit{labour intensity and practical infeasibility when assessing large idea pools}. Methods such as the Consensual Assessment Technique (CAT) typically require expert judges to individually rate each idea, leading to significant human cognitive effort and resource expenditure~(\cite{Beaty2021, Ceh2021, Cseh2019}). This manual process quickly becomes infeasible, costly, and time-consuming as the number of ideas increases, significantly restricting the scalability of subjective evaluation approaches~(\cite{Beaty2021}).

Secondly, \textit{subjectivity and inconsistency} among raters are another key challenge. Since subjective methods rely heavily on individual judgment, inter-rater variability and disagreements frequently occur, even among expert judges. Variability in raters’ backgrounds, experience levels, and interpretation of creativity metrics can significantly affect evaluation consistency~(\cite{Beaty2021, Ceh2021, Cseh2019}). For instance, different raters may perceive the originality or feasibility of an idea differently, thus threatening the reliability and validity of subjective ratings~(\cite{Beaty2021}). Such variability highlights the potential for evaluation biases, reducing confidence in the assessments unless extensive training or consensus-building processes (like Delphi techniques) are employed~(\cite{Ceh2021}).

Third, the subjective evaluation methods often \textit{confuse ideas and concepts}, particularly when raters are not explicitly trained in distinguishing between early-stage ideas and developed concepts. Literature highlights that the terms "ideas" and "concepts" are frequently used interchangeably in practice~(\cite{Sankar2024}), causing evaluators to prematurely judge feasibility and utility, which are more appropriate for concepts rather than ideas~(\cite{Kudrowitz2013}).

Lastly, subjective methods are known to be \textit{vulnerable to psychological biases} stemming from the evaluator's subjective experiences and personal cognitive schemas. Evaluation methods that heavily rely on human judgment inherently reflect not only the creativity of the idea but also the evaluator's personality traits, preferences, and biases~(\cite{Carson2005, Forthmann2017}). For instance, the evaluator’s familiarity and/or preference for certain idea types might inadvertently bias judgments, particularly affecting assessments of other ideas in terms of novelty (originality), or usefulness~(\cite{Carson2005}).

Moreover, several \textit{cognitive factors} related to human judgment processes can negatively influence idea assessment outcomes. Three specific cognitive barriers have been identified in group-based evaluation processes: \textit{production blocking, evaluation apprehension, and free-riding}~(\cite{Kudrowitz2013}). Production blocking occurs when groups limit the effective expression or evaluation of ideas; evaluation apprehension happens when raters become overly cautious or conservative due to fear of criticism; and free-riding refers to participants withholding effort, assuming others will perform the evaluation tasks~(\cite{Kudrowitz2013}). These cognitive biases inherently reduce the effectiveness and reliability of human-based evaluation processes.

Considering these limitations, particularly scalability, subjectivity-induced inconsistencies, and cognitive biases, there is a need for complementary approaches that can objectively assess ideas. Such approaches, if properly designed, could substantially reduce the cognitive load on human evaluators and offer systematic, reproducible methods for efficiently handling large idea pools.

\subsection{Objective Methods for Idea Assessment}
Objective methods for idea assessment have gained prominence recently, particularly due to their ability to overcome some of the inherent limitations of subjective methods as stated above. Most objective methods employ computational techniques and algorithms to analyze, score, and classify ideas. These methods have their roots in statistical methods, natural language processing (NLP), machine learning (ML), semantic analysis, and data-driven assessment strategies.

A common approach among these methods is the use of distance-based measures. These typically quantify how distinct or novel an idea is by computing its distance relative to a predefined baseline, reference set, or corpus of existing ideas. This distance measure can be computed through various methods. For instance, Latent Semantic Analysis (LSA), which represents textual ideas as high-dimensional numerical vectors, has been extensively employed. LSA computes semantic similarities or dissimilarities among ideas based on their underlying meaning derived from large textual corpora~(\cite{Acar2023, Bossomaier2009, Dumas2014, Dumas2021}). On the other hand, Semantic distance-based methods that use embedding models to create embedding vectors have been validated in divergent thinking tasks such as the Alternate Uses Task (AUT)~(\cite{stevenson2020a, Acar2023, Dumas2014, Dumas2021}). For example, Beaty et al.(\cite{Beaty2021, Beaty2022}) have introduced the SemDis platform, which utilizes static embedding models (e.g., GloVe, Word2Vec) to measure semantic distances between ideas to score originality within AUT settings~(\cite{Beaty2021, Beaty2022, Dumas2021}). These methods have demonstrated strong predictive validity and reliability in open-ended, divergent thinking assessments, correlating highly with human judgments~(\cite{Beaty2022, Dumas2021, Gunther2019, Johnson2023}).

Another prominent approach includes text mining and NLP techniques aimed at systematically evaluating the creativity of ideas. These methods encompass unweighted word counts, frequency of occurrence, stop-word inclusion, part-of-speech analysis, and inverse frequency weighting~(\cite{Dumas2021f}). Additionally, NLP-based systems have been used to automatically analyse creativity within narrative tasks by examining the diversity and associative richness of textual content. Such automated systems provide rapid, consistent, and scalable assessments of creativity, originality, and elaboration~(\cite{Jackson2022, Johnson2023, Dumas2021f}).

Recently, Large language models (LLMs) have also emerged as prominent tools for automated creativity assessment. Originality and Creativity Scoring using these Artificial Intelligence (AI) models, for example, employs foundational and fine-tuned models to directly provide scores for originality and creativity~(\cite{Organisciak2023, stevenson2022p}). For example, Organisciak et al. introduced Originality and Creativity Scoring (OCSAI), which leverages fine-tuned GPT models to provide automated originality and creativity scores for ideas generated during AUT tasks~(\cite{Organisciak2023}). Similarly, LLM-driven evaluation techniques like the Divergent Semantic Integration (DSI) approach apply BERT embeddings to evaluate narratives in storytelling by quantifying semantic connections~(\cite{Johnson2023}). These advanced computational methods can quickly handle large ideas in the form of texts, providing objective scores in specific contexts~(\cite{Johnson2023, Organisciak2023}).

Beyond textual idea evaluation, objective methods also include visual creativity assessments. An example is AuDrA, an automated drawing assessment platform that uses machine learning models to predict creativity scores for visual artwork and sketches~(\cite{Patterson2023}). These visual approaches indicate the potential to expand automated creativity assessments beyond textual boundaries, enabling the evaluation of multi-modal creativity.

A common thread unifying these objective methods lies in their reliance on mathematical frameworks. This foundation enables automated, rapid, reproducible, and scalable evaluations, effectively reducing subjective bias. However, this reliance also carries the potential for oversimplification, which could inadvertently lead to overlooking valuable ideas.

\subsection{Limitations of Existing Objective Methods}
Despite their effectiveness in specific contexts, existing objective methods exhibit several limitations that significantly restrict their applicability to evaluating ideas for solving complex, real-world product design problems. These limitations include semantic ambiguity, narrow task-specific validation, and insufficient alignment with product design contexts.

Firstly, existing methods that uses semantic models such as LSA, Word2Vec, and GloVe, are predominantly validated and applied within divergent thinking tasks, such as the Alternate Uses Task (AUT), Remote Associates Test (RAT), and narrative creativity assessments for storytelling~(\cite{stevenson2020a, Acar2023, Beaty2022, Dumas2021, Johnson2023}). These tasks typically present open-ended, context-free problems with minimal constraints. However, product design integrates divergent and convergent thinking, domain-specific knowledge integration, technical and practical constraints, and iterative refinement, addressing complex real-world problems. These factors are not adequately captured or addressed by existing semantic-based approaches~(\cite{Kudrowitz2013, Forthmann2017}).

Secondly, semantic embedding models, particularly LSA, have been shown to encounter significant semantic limitations, notably polysemy and homonymy~(\cite{Bossomaier2009, Gunther2019}). Polysemy refers to words that carry multiple meanings depending on the context, while homonymy involves words spelt identically but possessing completely different meanings. These linguistic ambiguities present considerable challenges for computational systems using static embeddings, reducing accuracy and reliability in similarity measurements~(\cite{Bossomaier2009, Gunther2019}). Especially in ideas within the context of product design, where terminologies can be highly nuanced, technical, and context-specific, such semantic ambiguities become even more problematic, potentially undermining the validity of automated idea assessments.

Moreover, the existing methods emphasized in creativity literature have primarily been to understand psychological aspects of idea generation rather than directly evaluate the practical utility or design value of the ideas themselves~(\cite{Forthmann2017, Jackson2022}). Most methods developed thus far aim primarily to uncover psychological insights, such as cognitive styles, semantic memory structures, or creative potential of idea generators~(\cite{Beaty2023, Forthmann2017, Jackson2022, Kenett2019}). These approaches analyze ideas as proxies for psychological processes, often overlooking how these ideas may practically contribute to product development, innovation outcomes, design effectiveness and aid novice designers during the process~(\cite{Forthmann2017, Jackson2022}).

Additionally, advanced large language models (LLMs) like GPT, despite their capacity for rich semantic representation and language generation, also exhibit significant limitations when directly employed for automated idea evaluation. LLM-based evaluations typically depend on carefully crafting input prompts and are sensitive to the phrasing and elaboration level of idea descriptions, thus affecting their reliability and generalizability~(\cite{Organisciak2023, stevenson2022p}). Such models also lack transparency in decision-making, often limiting interpretability, reproducibility and justification for why a specific score was assigned to an idea. This black-box nature of LLM-based scoring can diminish their acceptance and trustworthiness in professional design environments, where designers require transparent, understandable, and actionable evaluations to effectively advance ideas through the product development pipeline~(\cite{Organisciak2023, stevenson2022p}).

Although objective methods represent significant advancements over subjective methods, their current limitations—specifically narrow task validation, semantic ambiguities, psychological orientation rather than design orientation, limited transparency, and inadequate consideration of domain-specific constraints—underscore the requirement for developing a new computational framework explicitly aimed at supporting designers in systematically selecting a few promising ideas from abundant idea pools within constrained product design contexts. This framework should include the benefits of both subjective and objective methods, but also address their respective shortcomings.

\subsection{Metrics Used in Idea Assessment}
Several metrics and measures have been introduced in the literature thus far to assess ideas, each aiming to capture specific facets of idea quality~(\cite{ijaz2023}). These metrics can be grouped into three categories: \textit{individual idea quality, population quality, and relevance-related metrics}.

Metrics of individual idea quality predominantly include originality, novelty, creativity, and rarity. Originality and novelty, while often used interchangeably, capture slightly different nuances: novelty typically refers to the degree to which an idea is new relative to existing solutions, whereas originality emphasizes uniqueness or uncommonness within the specific context or population of generated ideas~(\cite{Kudrowitz2013, Forthmann2017}). Creativity encompasses broader cognitive aspects such as imagination, inventiveness, and conceptual richness. Additionally, elaboration—representing the level of detail, clarity, or sophistication of an idea description—has also been used as an individual measure, typically quantified by unweighted word count, part-of-speech inclusion, and inverse frequency weighting in textual responses~(\cite{Dumas2021f}).

Metrics of population quality focus on collective attributes of an entire idea set. Diversity (also termed variety) is a critical metric within this group, capturing the breadth or range of ideas within the generated set. Diversity is particularly important in scenarios broadly exploring the idea space~(\cite{Kudrowitz2013}). Another related measure, fluency, refers simply to the quantity of ideas generated and is commonly used in divergent thinking tests such as the Torrance Tests of Creative Thinking (TTCT) and Alternate Uses Tasks (AUT)~(\cite{Acar2014, Beaty2022, Dumas2021}).

Lastly, metrics related to relevance include feasibility, usefulness, utility, practicality, specificity, thoroughness, effectiveness, and potential impact. Feasibility assesses whether an idea can realistically be implemented given constraints, whereas usefulness or utility evaluates whether an idea effectively solves the intended problem or fulfills the goal~(\cite{Kudrowitz2013, Forthmann2017}). Relevance metrics, thus, directly bridge the gap between creative ideation and real-world implementation and are crucial in applied design contexts.

An inherent challenge noted in the literature is that different assessment methods emphasize different metrics, sometimes inconsistently. For instance, methods primarily used in divergent thinking tests (e.g., TTCT, AUT) often heavily emphasize originality, novelty, or creativity~(\cite{Acar2014, Beaty2022, Dumas2021}). Conversely, Pugh Chart, QFD, and NUF tests focus excessively on feasibility and usefulness~(\cite{Kudrowitz2013}).

\subsection{Ambiguity between Ideas and Concepts}
A critical and often overlooked issue in idea evaluation processes is the persistent confusion between ideas and concepts. Although these terms are widely used interchangeably in practice, their conceptual distinction carries significant implications for assessment criteria. An idea represents the preliminary proposal generated in response to a design challenge. At this stage, ideas are typically less detailed, not necessarily thoroughly elaborated, and aim to solve the challenge in a way that provokes new thinking or novel directions. A concept, however, embodies a more developed, structured, and refined form of an idea. Concepts are usually assessed based on feasibility, practicality, and alignment with predetermined constraints and criteria relevant to the design problem~(\cite{Sankar2024, Kudrowitz2013}).

The literature highlights that evaluators often prematurely apply criteria such as feasibility, usefulness, or practicality during the early stages of idea assessment. For instance, using assessment methods like the Pugh Chart, Quality Function Deployment (QFD) Matrix, or the Novelty, Usefulness, and Feasibility (NUF) test can lead evaluators to discard potentially valuable creative ideas simply because they appear initially impractical or less useful without sufficient elaboration or development~(\cite{Kudrowitz2013}). Thus, applying these criteria too early restricts genuinely novel ideas from moving forward, potentially compromising the overall innovation potential of the ideation process.

\subsection{Need for Consistency in Idea Description}
Idea assessment methods rely fundamentally on the textual descriptions of ideas, albeit subjective or objective. Therefore, the quality, clarity, and consistency of these descriptions—referred to as elaboration—influence the accuracy and reliability of assessments. In literature, elaboration generally refers to the degree of detail, precision, and completeness in idea descriptions~(\cite{Acar2023, Dumas2021f}). More importantly, the consistency of elaboration directly impacts semantic measurement techniques, including latent semantic analysis (LSA) and embedding-based approaches (e.g., GloVe, Word2Vec, BERT). Specifically, the semantic representation of ideas strongly depends on the textual detail provided, as insufficient or inconsistent elaboration can significantly alter semantic distances, resulting in inaccurate judgments about idea originality or novelty~(\cite{Acar2023, Dumas2014, Dumas2021}).

Several researchers have highlighted the necessity for ensuring uniformity in the detail level of idea descriptions when employing automated computational methods. For example, Forthmann et al. demonstrated that LSA-based methods are systematically biased by elaboration, meaning that more elaborated responses (longer, detailed descriptions) tend to receive higher originality or novelty scores compared to shorter responses, regardless of their actual creative merit~(\cite{Dumas2014, Dumas2021}). This elaboration bias threatens validity and fairness in automated creativity assessments. As a result, techniques such as inverse frequency weighting, removal of stop words, and other preprocessing steps have been suggested and employed to mitigate this bias~(\cite{Dumas2021}). Nonetheless, ensuring consistency in the initial idea elaboration remains the most straightforward and effective method to address such biases proactively. Consistent elaboration practices~(\cite{Sankar2024}) not only enhance the accuracy and fairness of computational evaluations but also ensure comparability and reliability of evaluation outcomes, particularly when assessing large pools of ideas systematically.

\subsection{Need for an Automated Method}
Given the significant limitations associated with subjective methods—including labour intensity, cognitive biases, inconsistency among human raters, and scalability—there is a need for automated, objective methods for idea assessment, particularly when confronted with a large number of ideas. Recent advancements in Conversational AI (CAI) systems have dramatically increased the capability to generate extensive pools of ideas in the order of thousands within a brief period. While this abundance represents tremendous potential for innovation, it simultaneously poses evaluation challenges. Human evaluators, even when employing efficient methods, would be unable to practically review and adequately assess such extensive idea pools. Objective methods can somewhat alleviate these challenges by offering rapid, consistent, reproducible, and scalable assessments. Computational methods leveraging text mining, natural language processing (NLP), and semantic distance approaches have been increasingly explored within creativity research to systematically measure aspects such as novelty, originality, and diversity. These computational methods, therefore, significantly reduce cognitive load and biases inherent in human evaluation. However, despite their promising capabilities, existing objective methods still exhibit notable limitations, as detailed earlier. Moreover, current literature predominantly focuses on understanding the creators rather than the created ideas themselves. Most assessments often analyze linguistic or semantic attributes of idea generation primarily to uncover psychological insights about the designers generating the ideas, rather than supporting practical design decisions. Such approaches, while valuable for psychological research, offer limited utility for novice designers seeking direct guidance in identifying and shortlisting innovative ideas. In response to these challenges, there is a requirement for developing automated computational assessment frameworks specifically tailored to product design scenarios to aid novice designers in efficiently navigating extensive idea pools. This research seeks to bridge this identified gap by introducing a novel computational method suited to address these challenges in product design ideation.


\hfill \hrule

\section{Categorization of Idea Assessment Methods}
\begin{figure*}[t!]
    \centering
    \includegraphics[width=0.95\textwidth]{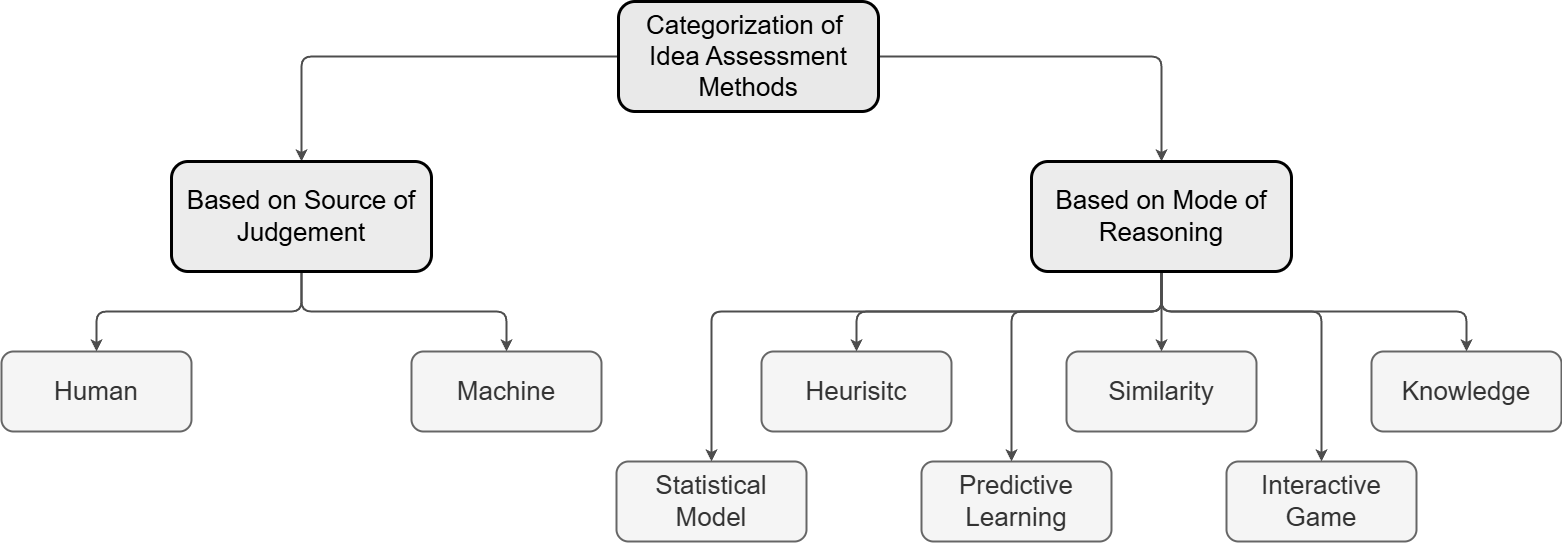}
    \caption{Categorization of Idea Assessment Methods}
    \label{fig:class_idea_ass}
\end{figure*}

The assessment of ideas in the product design process determines the trajectory from ideation to implementation. We have made a systematic categorization of idea assessment methods to provide clarity to researchers and design practitioners, as shown in Figure~\ref{fig:class_idea_ass}. This categorization is broadly structured along two dimensions: (a) Source of Judgment and (b) Mode of Reasoning.

\subsection{Categorization Based on Source of Judgment}

\subsubsection{Human-Based Methods}
Human-based assessment methods rely on direct judgments by individuals, typically experts, stakeholders, peer groups, or the idea generators (designers) themselves. The primary input to these methods comprises verbal, written, or visual descriptions of ideas or concepts. Outcomes of these methods are often numerical or qualitative ratings, rankings, or detailed expert commentaries reflecting subjective assessments of dimensions such as novelty, diversity, feasibility, usefulness, etc. The mode of assessment is fundamentally interpretative, grounded in personal or group-based heuristics developed through experience and expertise. While human-based methods offer significant advantages, such as capturing nuanced insights, context sensitivity, interpretive flexibility, and incorporation of tacit knowledge, they can also introduce biases, inconsistencies, and challenges in scalability, especially when evaluating large idea pools. Notable human-based methods include the Consensual Assessment Technique (CAT), Expert Panel Reviews or Focus Groups, Top-N Scoring methods, Self and Peer Evaluations (Snapshot scoring), Creative Achievement Questionnaires (CAQ), Scattered CAT, AUT Scoring, Pugh Chart, (Quality Function Deployment) QFD, (Novelty, Usefulness and Feasibility) NUF Test, Forward Flow, Creativity Quotient (CQ), etc.

\subsubsection{Machine-Based Methods}
Machine-based assessment methods use computational tools to objectively assess ideas. Inputs to these methods are typically structured or semi-structured textual descriptions, visual data, or numerical representations of ideas. The outcome usually consists of quantitative scores or classifications across various creativity metrics such as novelty, diversity, originality, feasibility, etc. The mode of assessment leverages computational algorithms, semantic embeddings, predictive models, or rule-based systems, which systematically compare ideas to established datasets, ontologies, or domain standards. These approaches provide scalability, objectivity, reproducibility, and consistency, which significantly help designers navigate large volumes of generated ideas quickly. However, their accuracy and validity depend critically on algorithmic sophistication, quality of input data, and alignment with human judgment criteria. Examples of machine-based methods include Distance-based similarity assessment platforms (e.g., SemDis, Latent Semantic Analysis (LSA), Divergent Semantic Integration (DSI), Maximum Associative Distance (MAD), etc.), Large Language Model (LLM)-based evaluation tools (e.g., OCSAI, CLAUS, AIDE, CAP, SCTT-AI, GPT-based evaluators, Text mining methods, etc.), Rule-based systems (Design Heuristic Checklists, patent matching tools, etc.), AI based Predictive Models (e.g., classifier-based feasibility prediction systems), and visual evaluation systems (e.g., AuDrA, TCT-DP, automated drawing assessment).

\subsection{Categorization Based on Mode of Reasoning}

\subsubsection{Heuristic-Based Methods}
Heuristic-based reasoning methods use predefined criteria, checklists, or intuitive judgments derived from experience or expert consensus. Inputs are structured or semi-structured idea descriptions evaluated through expert-defined criteria. Outcomes include numerical scores, categorical decisions, or descriptive evaluations. Assessments are typically performed by human experts or through automated checklist-based algorithms. These methods support designers by simplifying complex decision processes, rapidly filtering ideas, and integrating domain expertise. However, heuristic reasoning may lack objectivity, scalability, and consistency across evaluators and contexts. Representative heuristic-based techniques include CAT, Expert Panels, Pugh Charts, Quality Function Deployment (QFD), and Novelty-Usefulness-Feasibility (NUF) testing.

\subsubsection{Statistical Model-Based Methods}
Statistical model-based reasoning methods quantify idea quality using statistical formulae, frequency counts, or distribution analyses. Inputs are numerical or textual data representing ideas, analyzed statistically to generate outcomes such as originality scores, fluency counts, or elaboration metrics. These evaluations are typically automated, leveraging algorithms that compare ideas against large datasets or statistical distributions. This method benefits designers by offering objective and quantitative comparisons across idea pools. However, the oversimplification of complex dimensions such as creativity may limit its effectiveness in nuanced assessments. Examples include Response Frequency methods, Elaboration and Fluency Scoring (e.g., Torrance Tests of Creative Thinking), AUT Scoring Algorithm, NUF Testing, Utility Theory, IF, etc.

\subsubsection{Similarity-Based Methods}
Similarity-based methods most often employ semantic distances to assess ideas through semantic analysis, measuring the conceptual distance between ideas. Textual inputs are converted into numerical vectors within embedding spaces, resulting in quantitative outcomes of similarity scores. Evaluations are automated using NLP techniques such as Latent Semantic Analysis (LSA), Word2Vec, GloVe embeddings, and transformer-based models (BERT, GPT). These methods greatly assist designers by providing objective and rapid measures of conceptual divergence of ideas in conceptual spaces. However, they can be limited by linguistic ambiguities (polysemy and homonymy) and may struggle with context-specific issues in technical or domain-specific texts. Key examples include SemDis, DSI, LSA-based methods, Text Mining Methods, Associative Distance models, etc.

\subsubsection{Predictive Learning-Based Methods}
Predictive learning-based methods employ machine learning or deep learning techniques from AI and use classifiers or regression models trained on historical datasets to predict creativity or success likelihood. Inputs are typically labelled datasets of previously evaluated ideas, resulting in predictive outcomes such as probability scores or classifications regarding idea feasibility, originality, or market potential. The mode of assessment is computationally intensive, relying on algorithmic patterns and correlations learned from extensive historical data. While these methods are highly scalable and objective, they require substantial initial investment in dataset preparation, validation, and algorithm training. Examples include OCSAI (Originality Classification System using AI), CLAUS, AIDE, SCTT-AI, LLM-based predictive systems (GPT, Gemini, etc), classifier-based feasibility or success-likelihood prediction models, etc.

\subsubsection{Knowledge-Driven Methods}
Knowledge-driven methods evaluate ideas through structured knowledge bases, domain ontologies, or using simulation scenarios. Inputs to these systems are usually technical descriptions of ideas or concepts compared against existing knowledge structures or patent databases. Outcomes are assessments of novelty, feasibility, or practicality expressed as similarity scores, feasibility ratings, or conceptual maps. The assessment relies heavily on domain-specific knowledge structures, providing designers valuable insights into practical viability. However, their effectiveness is limited by the comprehensiveness, accuracy, and precision of underlying knowledge bases. Prominent examples include Patent Similarity Matching tools, Technology Landscape Mapping, Forward Flow metrics, Creativity Quotient, etc.

\subsubsection{Interactive Game-Based Methods}
Game-based interactive methods implicitly assess creativity through user interactions within gamified environments based on physical or digital scenarios. Inputs are typically behavioural data collected from user interactions within the game, and outcomes are creativity assessments inferred from these interactions, presented as implicit behavioural scores or creativity indices. Assessments rely on real-time behavioural analytics and stealth tracking, providing engaging and unobtrusive evaluation experiences. However, the complexity of designing valid interactive scenarios and accurately interpreting user behaviour presents significant methodological challenges. Notable methods include Stealth Assessment within games, Digital Creativity Games, and interactive role-playing-based games designed to capture the creative potential of ideas.

\hfill \hrule

\section*{Idea Generation using CAI}
\label{sec:idea_gen_cai}
Idea generation is a cognitively expensive process that requires experience to master~(\cite{Sankar2023}). Novice designers often face the struggle to generate a large variety of novel ideas. Conversational AI (CAI) systems powered by advanced natural language processing (NLP) models have been reported in our earlier work~(\cite{Sankar2024})  for generating a required number of ideas quickly. The conversational AI system leverages large language models to generate text based on given prompts. These models are trained on extensive datasets, enabling them to understand and generate human-like text across various domains. In the context of idea generation,~(\cite{Sankar2024}) established that CAI-based ideation facilitates the creative process by overcoming human cognitive limitations such as fatigue and design fixation. 

\subsection{Idea Representation}
\begin{figure}
    \centering
    \includegraphics[width=0.98\textwidth]{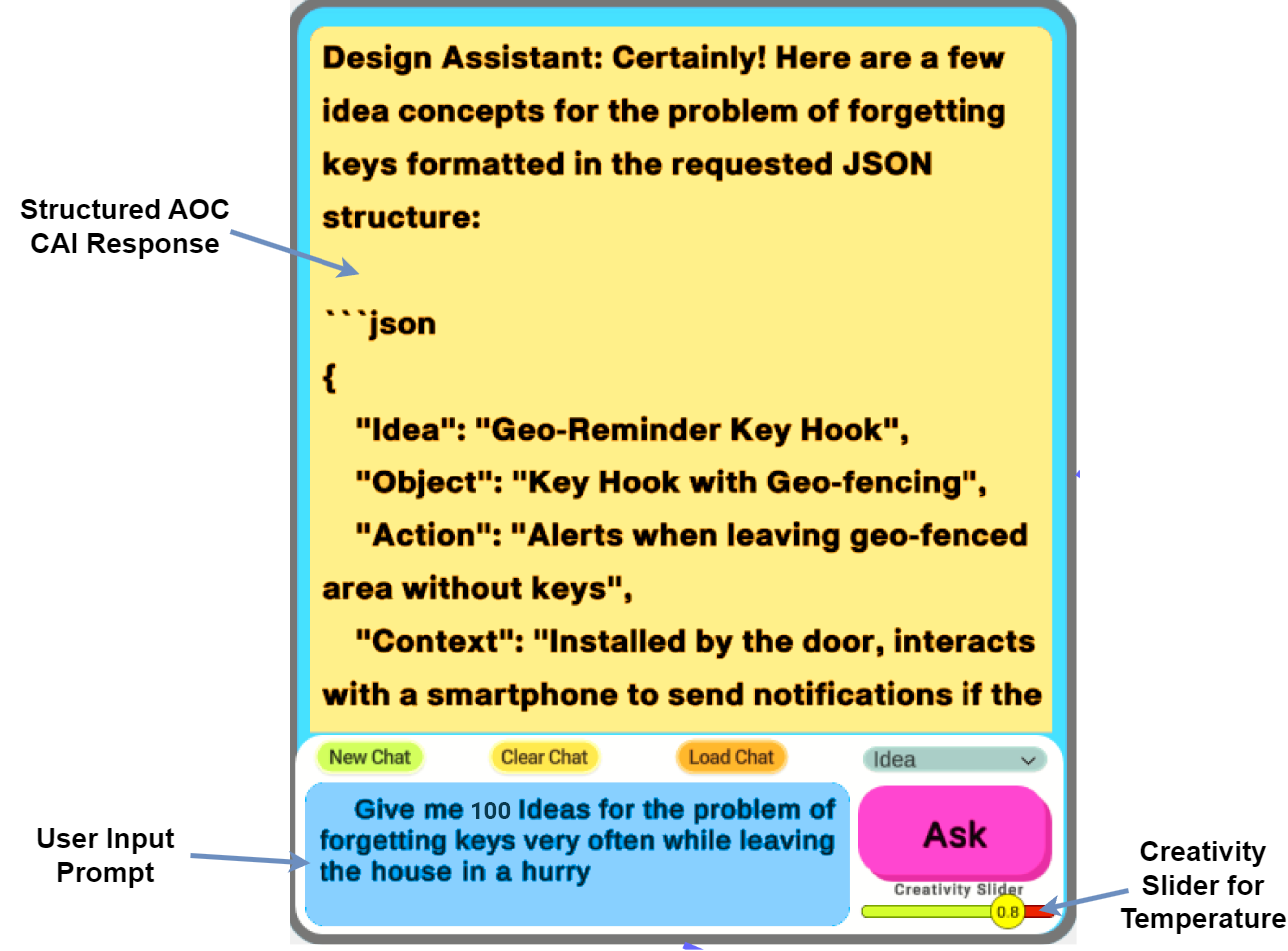}
    \caption{Interface of the Design Chatbot of the Custom-Built Conversational AI (CAI) Tool}
    \label{fig:cai_tool}
\end{figure}

A Custom-built interface was developed for generating ideas, as shown in Figure~\ref{fig:cai_tool} with labels for different fields. 
The response of the CAI system is managed with a field marked as 'creativity slider', which adjusts the parameter commonly called temperature; a higher temperature setting produces more diverse outputs, while a lower temperature setting produces focused and predictable ideas.

In order for CAI to present ideas that are similarly detailed, coherent, and easily interpretable, an Action-Object-Context (AOC) model was developed. Here, \emph{action} describes what the idea aims to achieve or perform; \emph{object} refers to the entity involved in the action, and \emph{context} provides the situation where the action happens. The ideas generated are stored as a JSON file for further investigation.

\subsection{Idea Generation}
The user interface of the tool was designed and developed by the authors as shown in Figure~\ref{fig:cai_tool} with labels for different fields. The DesignerAI is now integrated into a standalone software application that has not yet been publicly released. The preliminary testing from our earlier study~(\cite{Sankar2024}) has shown promising results in its ability to assist designers in generating a variety of ideas. In the same study, the authors developed a structure for the input prompts and output responses. The creativity of the ideas generated by our tool is controlled with a field marked as 'creativity slider', which adjusts the parameter called temperature; a higher temperature setting produces more diverse outputs, while a lower temperature setting produces focused and predictable ideas.

The problem statements serve as input prompts for the CAI system, guiding the generation of relevant ideas. It was established in~\cite{Sankar2024} that the statements of the machine-generated ideas were valid and human-like to the experts and faired significantly better vis-a-vis human-generated ideas in terms of novelty, fluency and variety. That study used the following six problem statements derived from common day-to-day experiences.

\begin{itemize}
    \item PS1: Product for segregation as a means for effective waste management
    \item PS2: Product for footwear disinfection and cleaning for improved hygiene and safety
    \item PS3: Product for enhancing household dish cleaning efficiency and sustainability
    \item PS4: Product for enhancing comfort and efficiency for prolonged standing in queues
    \item PS5: Product for bird-feeding for fostering mental well-being of elderly individuals at Home 
    \item PS6: Product for convenient umbrella drying and storage on travel
\end{itemize}

The CAI tool uses these prompts to generate 100 ideas for each problem statement, each structured according to the AOC model. A representative sample of CAI-generated ideas is shown in Table~\ref{tab:ideas}. \footnote{Any human-generated ideas can also be processed in subsequent sections if they are presented as per the AOC structure.}


\begin{table*}[htbp]
    \centering
    \begin{tabular}{|p{1cm}|p{3cm}|p{3cm}|p{2cm}|p{6cm}|}
        \hline
        \textbf{S. No.} & \textbf{Title of the Idea} & \textbf{Action} & \textbf{Object} & \textbf{Context} \\
        \hline
        \multicolumn{5}{|l|}{Problem Statement 1: Product for segregation as a means for effective waste management} \\
        \hline       
        1 & Smart Segregation Bins & Automatically sort waste & Bins & Use sensors and AI to identify and segregate waste into recyclables, organics, and general waste \\
        \hline
        2 & Colour-Coded Waste Bags & Visually indicate waste type & Waste Bags & Provide different coloured bags for different types of waste to encourage proper segregation at the source\\
        \hline
        3 & Segregation Education App & Educate and guide users & Mobile App & An app that teaches waste segregation and helps track the environmental impact of proper waste management\\
        \hline
        4 - 100 & \dots & \dots & \dots & \dots \\
        \hline
        \multicolumn{5}{|l|}{Problem Statement 2: Product for convenient umbrella drying and storage on travel} \\
        \hline
        1 & Magnetic Quick-Dry Umbrella Case &	Magnetically attaches and speeds up drying & Umbrella Case & Useful for attaching to metal doors or lockers, particularly in public areas or small apartments\\
        \hline
        2 & Solar-Powered Outdoor Umbrella Stand & Uses solar energy to dry umbrellas & Umbrella Stand & Ideal for eco-friendly outdoor storage solutions like patios or gardens\\
        \hline
        3 & Automated Umbrella Spin Dryer & Spins the umbrella to shed water & Spin Dryer & Perfect for entrance areas, offering a quick method to reduce dripping water from freshly used umbrellas\\
        \hline
        4 - 100 & \dots & \dots & \dots & \dots \\     
        \hline
        \multicolumn{5}{|l|}{Problem Statement 3: Product for footwear disinfection and cleaning for improved hygiene and safety} \\
        \hline
        1 & UV Sanitization Mat & Disinfects using UV light & Mat & Placed at the entrance of homes or offices to sanitize shoes upon entry\\
        \hline
        2 & Self-Cleaning Smart Shoe Cabinet & Automatically cleans and disinfects & Shoe Cabinet & For storage of shoes with built-in disinfection technology\\
        \hline
        3 & Antibacterial Shoe Spray Dispenser & Releases antibacterial spray & Spray Dispenser & A hands-free dispenser that applies a disinfectant to shoes\\
        \hline
        4 - 100 & \dots & \dots & \dots & \dots \\  
        \hline
        \multicolumn{5}{|l|}{Problem Statement 4: Product for enhancing household dish cleaning efficiency and sustainability} \\
        \hline
        1 & Eco-Friendly Dish Scrubber & Cleans using sustainable materials & Dish Scrubber & Made from biodegradable materials to scrub dishes effectively while being environmentally friendly\\
        \hline
        2 & Water-Saving Sink Attachment & Regulates water flow & Sink Attachment & It limits water usage during dish cleaning without compromising on cleanliness.\\
        \hline
        3 & Solar-Powered Dishwasher & Operates using solar energy & Dishwasher & Harnesses solar power to save electricity and reduce carbon footprint for household dish cleaning.\\
        \hline
        4 - 100 & \dots & \dots & \dots & \dots \\
        \hline
        \multicolumn{5}{|l|}{Problem Statement 5: Product for enhancing comfort and efficiency for prolonged standing in queues} \\
        \hline
        1 & Portable Foot Roller & Massage and relax feet & Foot Roller & Provides relief from standing fatigue\\
        \hline
        2 & Queue Companion Stool & Provide temporary seating & Stool & Allows rest without losing place in the queue\\
        \hline
        3 & Mobile Queue Barrier & Guide and organize queue formation & Barrier & Streamlines queues to prevent overcrowding\\
        \hline
        4 - 100 & \dots & \dots & \dots & \dots \\
        \hline
        \multicolumn{5}{|l|}{Problem Statement 6: Product for bird-feeding for fostering mental well-being of elderly individuals at Home} \\
        \hline
        1 & Smart-Connect Birdhouse & Facilitates remote interaction & Birdhouse & Allows elderly individuals to watch and interact with birds via a smartphone app, promoting a sense of connection with nature.\\
        \hline
        2 & Chirp-O-Meter Feeder & Measures bird activity & Bird Feeder & Logs the frequency of bird visits, providing elderly users with a fun way to track and anticipate bird interactions.\\
        \hline
        3 & Garden Flight Paths & Guides Birds &	Landscape Feature & Designs in the garden that attract specific species to feeding spots, offering a visually stimulating activity for the elderly.\\
        \hline
        4 - 100 & \dots & \dots & \dots & \dots \\
        \hline
    \end{tabular}
    \caption{A Sample List of 100 Ideas Generated by the CAI in the AOC Structure for Each of the Six Problem Statements}
    \label{tab:ideas}
\end{table*}

\subsection{Advantages of Using CAI for Idea Generation}
Using the CAI tool for idea generation offers the following advantages:

\textit{Speed and Efficiency}: The CAI tool can generate many ideas in a fraction of the time it would take human designers. This rapid ideation accelerates the creative process and allows for exploring a broader idea space.

\textit{Overcoming Cognitive Limitations}: The CAI tool is not subject to human cognitive limitations such as fatigue, design fixation, and mental blocks. This enables a more comprehensive and unbiased exploration of the idea space.

\textit{Standardization and Consistency}: The AOC model provides a standardized format for representing ideas, ensuring consistency and coherence in the generated ideas. This standardization facilitates easier assessment and comparison of ideas.

\hfill \hrule

\section*{Characteristics of Idea Exploration}
Idea exploration involves generating a wide range of ideas to solve a given problem, with the aim of identifying unique, novel, and effective solutions.

\subsection{Concept of Idea Space}
The concept of idea space is rooted in several theoretical frameworks in creativity research and cognitive psychology. One such framework is the Geneplore Model~(\cite{hunt-earl, Ward2004-mw}), which posits that creative cognition involves two main processes: generation and exploration. In the generation phase, individuals produce a variety of mental representations known as pre-inventive structures. In the exploration phase, these structures are elaborated, refined, and evaluated to produce novel and useful ideas.

Another relevant theory is the Conceptual Blending Theory~(\cite{Fauconnier2003}), which suggests that creative ideas arise from the combination of different distinct mental sub-spaces. By blending elements from different sub-spaces, individuals can generate new ideas that transcend the limitations of any single domain.

Thus, the idea space, also known as the solution space, represents the entire range of potential ideas that can be generated to address a specific problem. For the context of this paper, the authors define \emph{Idea Space as a multi-dimensional conceptual space where each point corresponds to a unique idea}. In an ideal scenario, this space should be uniformly explored to uncover diverse and innovative solutions.

Designers navigate this idea space using their cognitive abilities, drawing on their knowledge, experience, and creativity. However, the exploration of idea space is not always uniform. Designers may face cognitive bottlenecks that hinder their ability to generate a wide range of ideas, leading to clusters of similar ideas and gaps in other space areas.

\subsubsection{Cognitive Processes Involved in Exploration of Idea Space}
Exploring the idea space involves complex cognitive processes that include divergent thinking, convergent thinking, and analogical reasoning:

\emph{Divergent Thinking}: The ability to generate multiple, diverse ideas from a single starting point. Divergent thinking is crucial for expanding the breadth of the idea space.

\emph{Convergent Thinking}: The ability to evaluate and refine ideas to identify the most promising solutions. Convergent thinking is essential for exploring the depth of the idea space.

\emph{Analogical Reasoning}: The ability to draw parallels between different domains or contexts to generate novel ideas. Analogical reasoning facilitates the blending of different conceptual spaces.

\subsubsection{Challenges in the Exploration of Idea Space}
Despite the potential for generating innovative solutions, exploring the idea space is fraught with challenges~(\cite{Sankar2024}). Designers often face cognitive bottlenecks that hinder their ability to explore the idea space comprehensively. These bottlenecks are classified by the authors~(\cite{Sankar2024}). For sake of completion, the bottlenecks are given below:

\textit{Design Fixation}: A tendency to become fixated on a particular idea, limiting the exploration of the idea space.

\textit{Cognitive Biases}: Systematic patterns of deviation from rationality in judgment and decision-making, such as confirmation bias and anchoring.

\textit{Lack of Domain Knowledge and/or Experience}: Insufficient knowledge or experience in a particular domain can restrict the ability to generate diverse and relevant ideas from different corners of the idea space.

\textit{Mental Block}: Limited time, budget, and technological resources can constrain the exploration of the idea space.

\subsection{Measures for Quantifying Exploration of Idea Space}

To objectively assess the exploration of idea space, we introduce two key measures: dispersion and distribution.

\subsubsection{Dispersion}
Dispersion refers to the spread or variability of the idea points in the 2D space. It indicates how much the points are spread out from each other and from the centre of the distribution across the idea space. Dispersion can be thought of as the extent to which the ideas vary from one another. High dispersion indicates that the ideas are diverse and cover a wide range of the solution space, while low dispersion suggests that the ideas are clustered in a specific region.

\emph{Key Aspects of Dispersion}:
\begin{enumerate}
    \item Range: The distance between the minimum and maximum values in each dimension.
    \item Variance and Standard Deviation: Measures of how much the points deviate from the mean.
    \item Spread: The extent to which points within a cluster are spread out.
    \item Outliers: Points that are far away from the majority of other points, indicating extreme values or unique ideas.
\end{enumerate}

\subsubsection{Distribution}
Distribution refers to the overall arrangement or pattern of the idea points in the idea space. It indicates how the points are organized or grouped and where the points are more concentrated or sparse. It refers to the uniformity with which ideas are spread across the idea space. A uniform distribution indicates that all regions of the idea space are explored equally, while a non-uniform distribution suggests that certain areas are more densely populated with ideas than others.

\emph{Key Aspects of Distribution}:
\begin{enumerate}
    \item Shape: The overall form of the point cloud (e.g., normal, skewed, bimodal).
    \item Density: Areas where points are more concentrated versus areas where points are sparse.
    \item Clusters: Groups of points close to each other, indicating similar ideas.
    \item Patterns: Specific arrangements or structures in the data (e.g., linear, circular).
\end{enumerate}

\begin{figure*}[!htbp]
 \begin{subfigure}[b]{0.49\textwidth}
     \includegraphics[width=\textwidth]{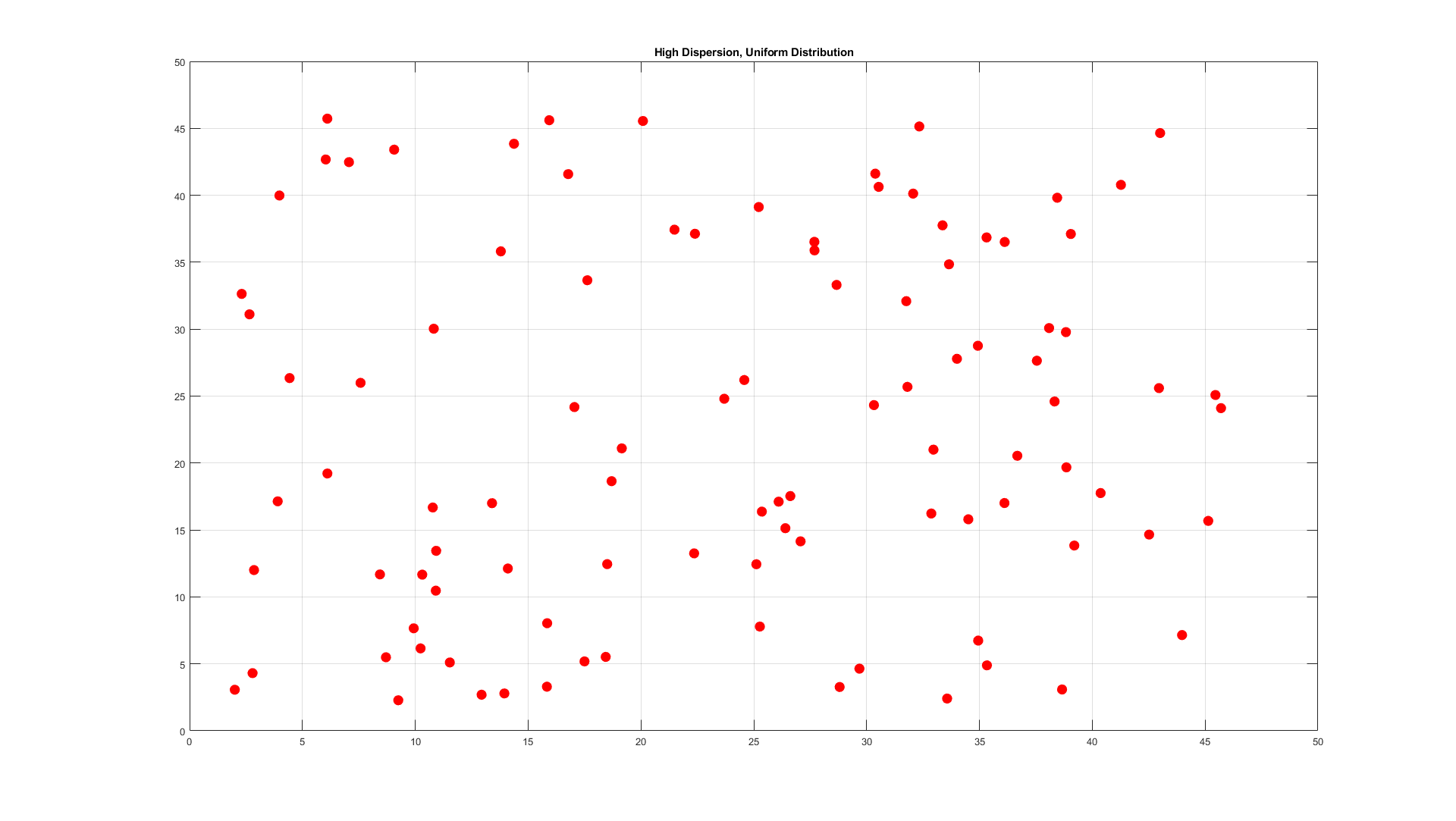}
     \caption{1.High Dispersion, Uniform Distribution}
     \label{fig:1_HDUD}
 \end{subfigure}
 \hfill
 \begin{subfigure}[b]{0.49\textwidth}
     \includegraphics[width=\textwidth]{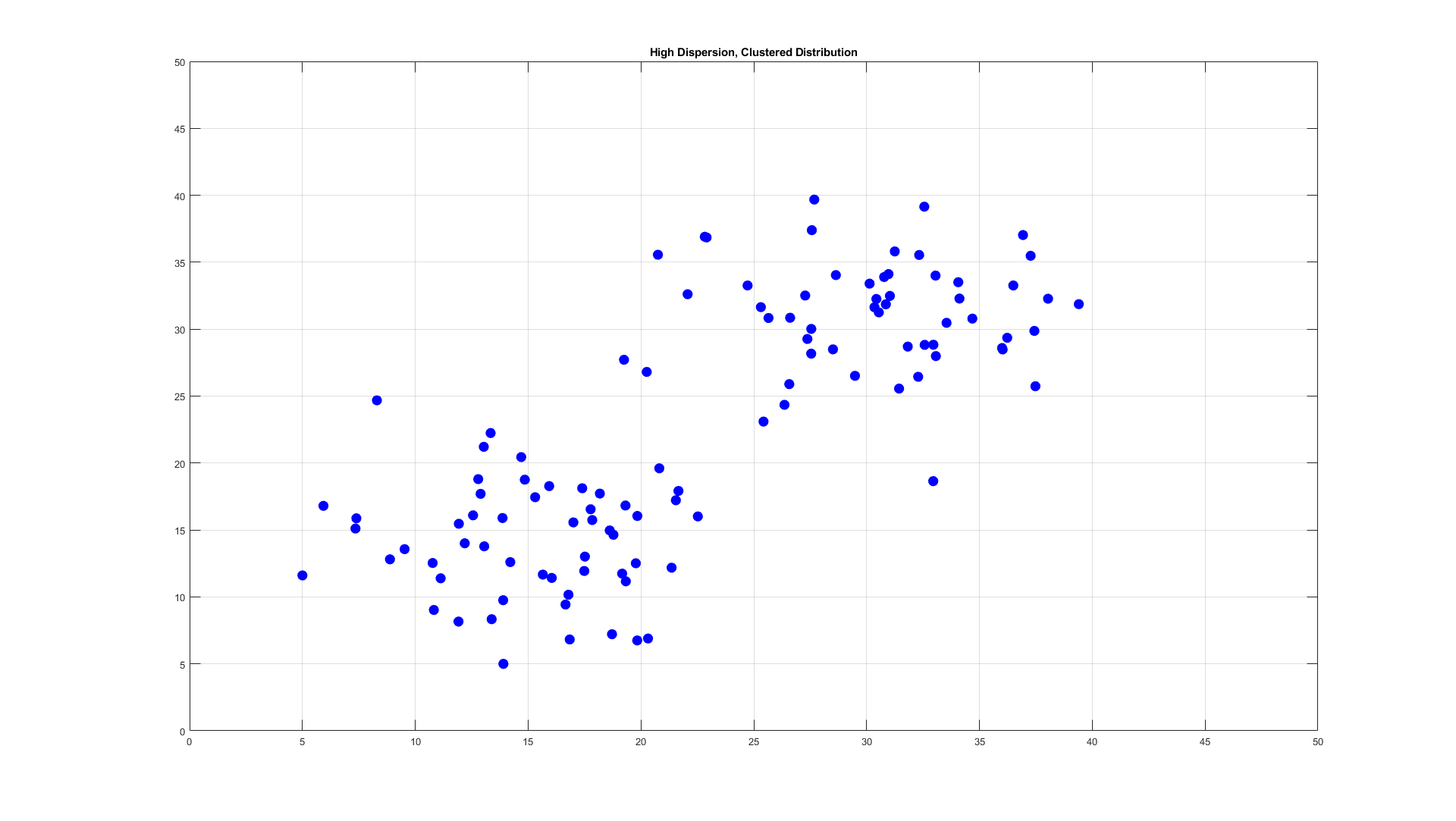}
     \caption{2.High Dispersion, Non-uniform Distribution}
     \label{fig:2_HDCD}
 \end{subfigure}
 
 \vfill
 \begin{subfigure}[b]{0.49\textwidth}
     \includegraphics[width=\textwidth]{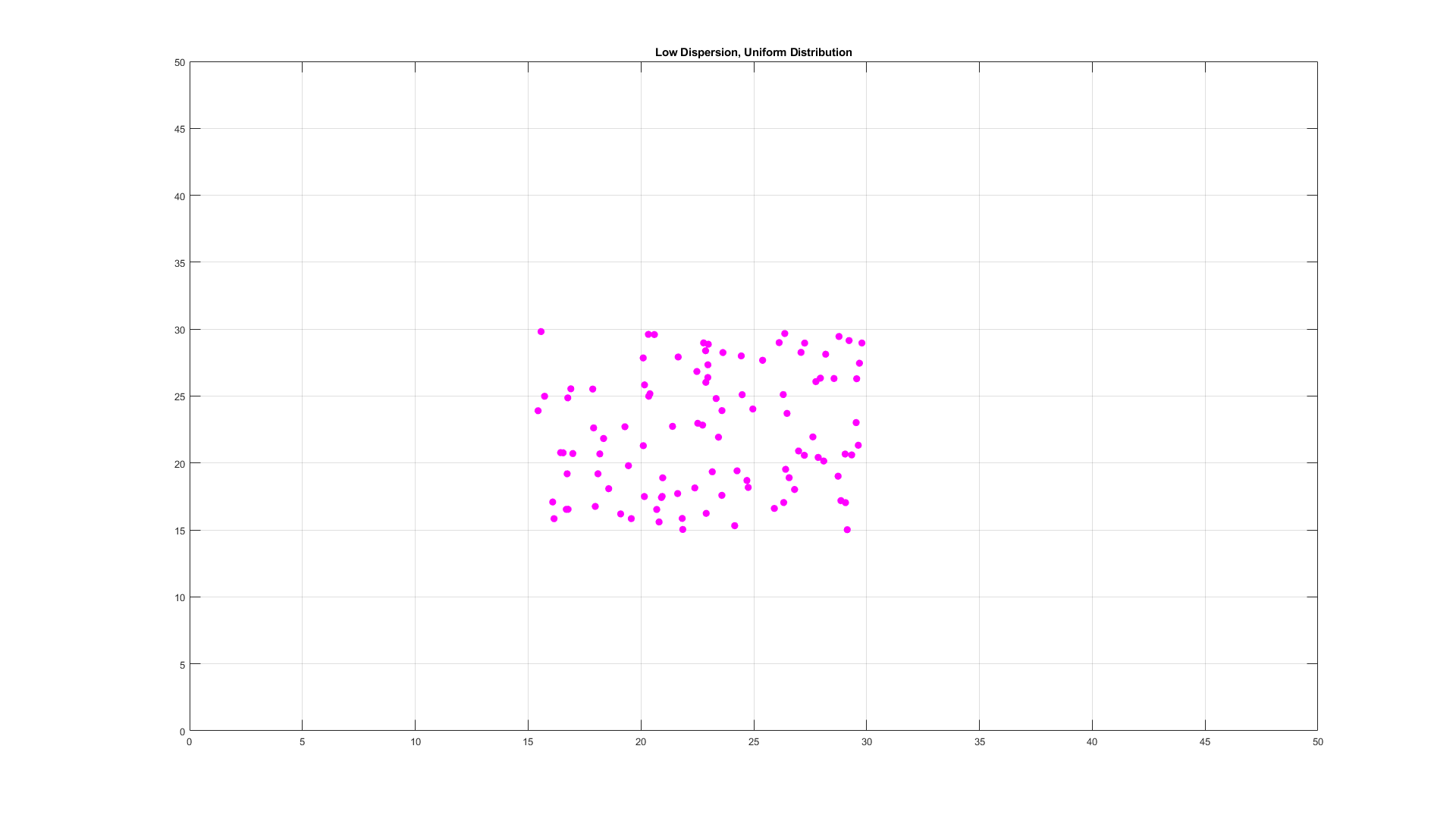}
     \caption{3.Low Dispersion, Uniform Distribution}
     \label{fig:3_LDUD}
 \end{subfigure}
 \hfill
 \begin{subfigure}[b]{0.49\textwidth}
     \includegraphics[width=\textwidth]{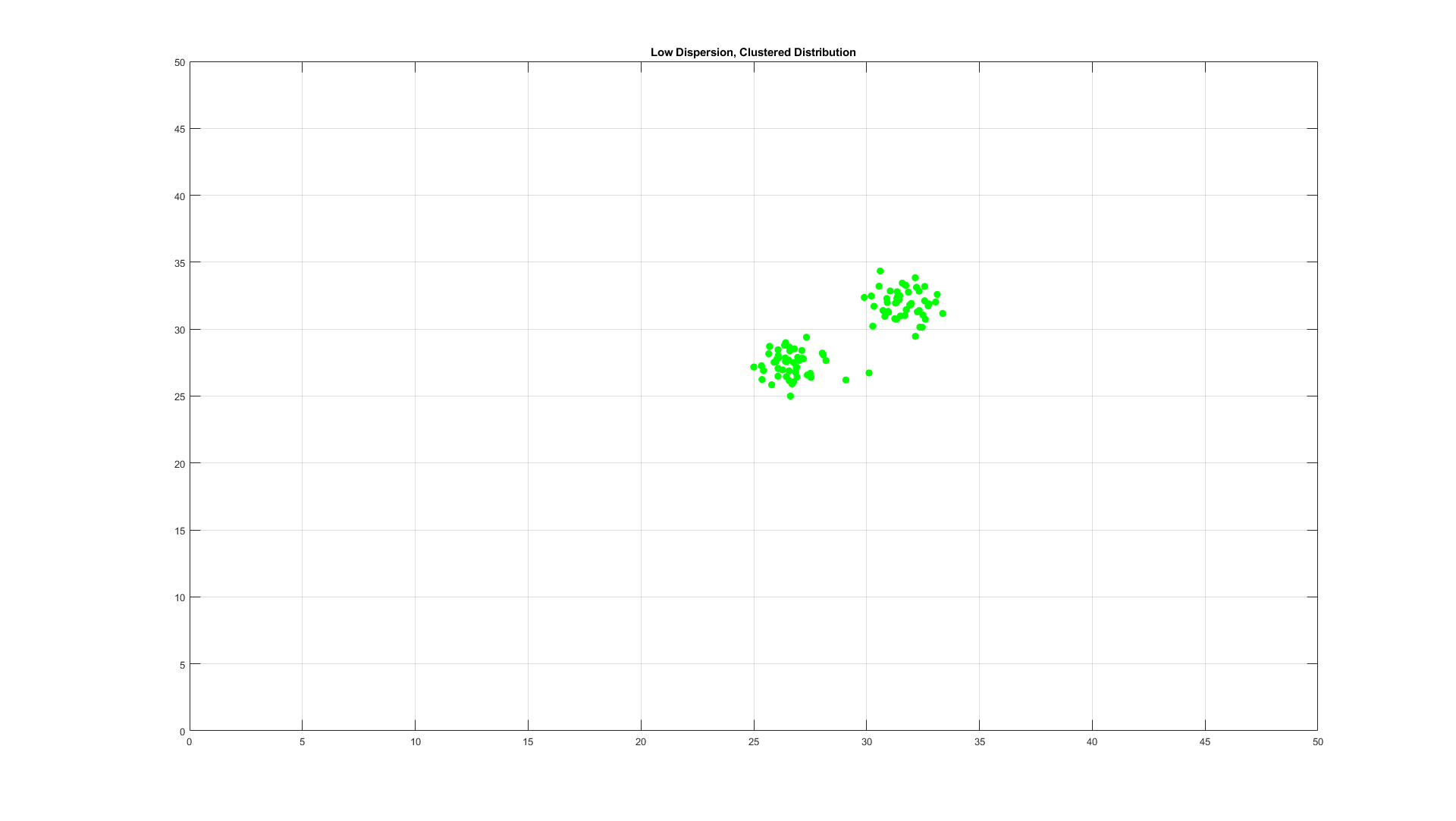}
     \caption{4.Low Dispersion, Non-uniform Distribution}
     \label{fig:4_LDCD}
 \end{subfigure}

 \caption{Classification of Idea Space}
 \label{fig:ideaspace_classification}

\end{figure*}

\subsection{Classification of Idea Space}

To illustrate the concept of idea exploration, we consider four possible scenarios based on the measures of dispersion and distribution:
\begin{enumerate}
    \item \textit{High Dispersion and Uniform Distribution} (Figure~\ref{fig:1_HDUD}): This scenario represents the ideal exploration of idea space, where ideas are diverse and evenly spread across the entire solution space. It indicates a comprehensive and balanced approach to idea generation.
    
    \item \textit{High Dispersion and Non-Uniform Distribution} (Figure~\ref{fig:2_HDCD}): In this scenario, ideas are diverse but clustered in certain regions of the idea space. While there is a wide range of ideas, some areas are over-explored while others are neglected.
    
    \item \textit{Low Dispersion and Uniform Distribution} (Figure~\ref{fig:3_LDUD}): This scenario represents a limited range of ideas that are evenly spread across the idea space. While the ideas are uniformly distributed, they lack diversity and may not provide innovative solutions.
    
    \item \textit{Low Dispersion and Non-Uniform Distribution} (Figure~\ref{fig:4_LDCD}): This scenario is the least desirable, where ideas are both limited in range and clustered in specific regions. It indicates a narrow and imbalanced approach to idea generation.
\end{enumerate}

\hfill \hrule

\section*{Embeddings for Idea Representation}
To objectively assess idea exploration, it is essential to represent ideas as points in a mathematical sense. This allows for the application of quantitative measures to evaluate the dispersion and distribution of ideas. By converting ideas into high-dimensional vectors, we can use techniques such as embeddings to capture the semantic and contextual similarities between ideas. Embedding techniques transform natural language descriptions of ideas into dense vector representations in a high-dimensional space. These vectors can then be analyzed to objectively assess the spread and uniformity of ideas within the idea space. This provides a foundation for developing automated methods to evaluate the ideas generated by Conversational AI systems, ensuring that the exploration of idea space is both comprehensive and balanced.

Embeddings is a collective name for a set of language modelling and feature learning techniques used in Natural Language Processing (NLP) and are pivotal for the functioning of Conversational AI (CAI) systems that use Large Language Models (LLMs) like GPT (Generative Pre-trained Transformers). Embeddings are dense vector representations of words, sentences, or even entire documents. These vectors capture semantic meaning and contextual relationships between different pieces of text, enabling machines to understand and generate human language more effectively. Embeddings transform textual data into numerical form, making it possible for deep learning algorithms to process and analyze text. They are particularly useful for LLMs because they provide a way to represent complex linguistic information in a compact, numerical and computationally efficient manner.

\subsection{Nature of Embeddings}
Embeddings work by mapping tokens, which are subwords, words or phrases, to vectors in a high-dimensional space (typically in the order of 1000s). This mapping is typically learned during the training phase of the embedding model. The goal is to position semantically and contextually similar words close to each other in this space while dissimilar words are positioned farther apart.

\subsubsection{Semantic Encoding}
The key to the effectiveness of embeddings lies in their ability to encode semantic meaning. This is achieved through training on large corpora of text, where the model learns to predict words based on their meaning. For example, in the sentences "The cat sat on the mat" and "The cat sat on the rug", embeddings capture the relationship between "mat" and "rug" and position them nearby in the vector space due to the frequent analogous usage.

\subsubsection{Positional Encoding}
In addition to semantic meaning, modern language models like Transformers also incorporate positional encoding to understand the context in which the words are being used. This is crucial because the meaning of a word can depend on its position in a sentence. Positional encoding adds information about the position of each word in the sequence, enabling the model to understand the structure, order and context of the text better. For example, in the sentences "The design in the dress was detailed." and "She has a natural talent for fashion design.", the usage of the word design refers to two different contexts, and this relationship is captured by the embeddings to understand the context.

\subsection{Mechanism of Embeddings}
Tokenization is the process of splitting text into smaller units called tokens, which can be words, subwords, or characters. For example, the sentence "Natural Language Processing" could be tokenized into ["Natural", "Language", "Processing"] or even into subword units like ["Nat", "ural", "Lang", "uage", "Processing"] depending on the tokenization strategy. The subword strategy is employed to help the model understand spelling mistakes in words while maintaining their meaning.

The dimension of an embedding vector is a critical parameter that influences the model's performance. Common dimensions range from 50 to 4096 or even higher for language models with larger parameters. The choice of dimension depends on the complexity of the task and the computational resources available. Higher dimensions can capture more nuanced relationships but require more computational power. While individual values do not have explicit meanings, collectively, they capture various aspects of semantic and syntactic information. For instance, certain dimensions might capture gender, tense, plurality and other linguistic attributes.

\subsection{Properties and Types of Embeddings}
In mathematical terms, embeddings are vectors in a high-dimensional space. They exhibit properties such as linearity, where vector arithmetic can capture analogies (for example, the vector arithmetic between the vectors of "king" - "man" + "woman" $\approx$ gives the vector for "queen"). This linearity is a powerful feature that enables embeddings to generalize across different contexts.

Embeddings can be created at various levels of granularity as follows:
\begin{enumerate}
    \item Word Embeddings: Represent individual words (e.g., Word2Vec, GloVe, Text-Embedding-3).
    \item Sentence Embeddings: Represent entire sentences (e.g., Sentence-BERT).
    \item Paragraph Embeddings: Represent longer text segments (e.g., Doc2Vec).
\end{enumerate}

For static embedding models like Word2Vec, the embeddings for a given word are constant. However, in contextual models like BERT or Text-Embedding-3, embeddings can vary depending on the word's context within a sentence. This allows for a more nuanced understanding of language.

\subsection{Types of Embedding Models}
There are two types of embedding models based on their ability to capture the similarities in the text. They are as follows:

\subsubsection{Static Embedding Models}
Static embedding models capture the semantic similarity in the text. Semantic similarity measures how much the meanings of two words, phrases, or sentences are alike, regardless of their specific contexts. This concept focuses on the intrinsic meaning of the words themselves.  Traditional word embedding models like Word2Vec, GloVe (Global Vectors for Word Representation), and FastText produce a single, static vector for each word, which represents its overall semantic meaning. These embeddings do not change based on context. Semantic similarity is about capturing the essence of meaning. For example, "device" and "instrument" would have high semantic similarity because they represent the same concept.

\subsubsection{Dynamic Embedding Models}
Dynamic embedding models capture contextual similarity in addition to semantic similarity in the text. Contextual similarity refers to the similarity between words or phrases based on the contexts in which they appear. This concept is often associated with contextual embeddings, which take into account the surrounding words to generate a representation that varies depending on the context. Contextual embeddings, like those produced by models such as BERT (Bidirectional Encoder Representations from Transformers), ELMo (Embeddings from Language Models), and Text-Embedding-3 model (OpenAI), generate different embeddings for the same word depending on its context. For example, the word "bank" will have different embeddings in "bank of a river" and "savings bank". These embeddings are generated by considering surrounding words, capturing the nuances and specific meanings in different contexts.

\subsection{Process of Conversion of Text to Embedding Vectors}
The process of converting text into vectors using any of the aforementioned embedding models involves several steps, such as Tokenization (Splitting the text into tokens), Embedding Generation (Mapping each token to its corresponding vector), and Aggregation (Combining the vectors to form a representation of the entire text segment). In practice, an embedding layer in a neural network architecture performs these steps. During training, the model adjusts the embedding vectors to minimize a loss function, which measures the difference between the model's predictions and the actual outcomes. The result is a set of embeddings that capture the underlying structure and meaning of the text.

\subsection{Contribution to CAI Behavior}
Embeddings are crucial for the behaviour of Conversational AI (CAI) models like GPT. They enable the model to understand and generate coherent and contextually appropriate responses. By capturing the semantic and contextual relationships between words, embeddings allow the model to generate text that is not only grammatically correct but also meaningful. The use of embeddings enhances the model's ability to understand context, making it possible to generate more relevant and accurate responses. This is particularly important in conversational AI, where understanding the context of a conversation is key to providing useful and meaningful interactions.

\subsection{Demonstrating the Validity of Embeddings}
In this paper, we use the \emph{Text-Embedding-3} (TE3), a dynamic embedding model from OpenAI for implementation. The following examples are shown to demonstrate the power of embeddings to translate text to vectors meaningfully.

\subsubsection{4.7.1. Word Embedding}
To illustrate the concept of semantic similarity between words, we compared the word \emph{chair} against three sets of words using their corresponding embeddings. These sets were selected based on human judgment to represent varying degrees of similarity:
\begin{enumerate}
    \item Set 1 (High Similarity): seat, sofa, bench
    \item Set 2 (Moderate Similarity): desk, table, cushion
    \item Set 3 (Low Similarity): light, monitor, fan
\end{enumerate}

\begin{figure}[htbp]
\centerline{\includegraphics[width=1.0\textwidth]{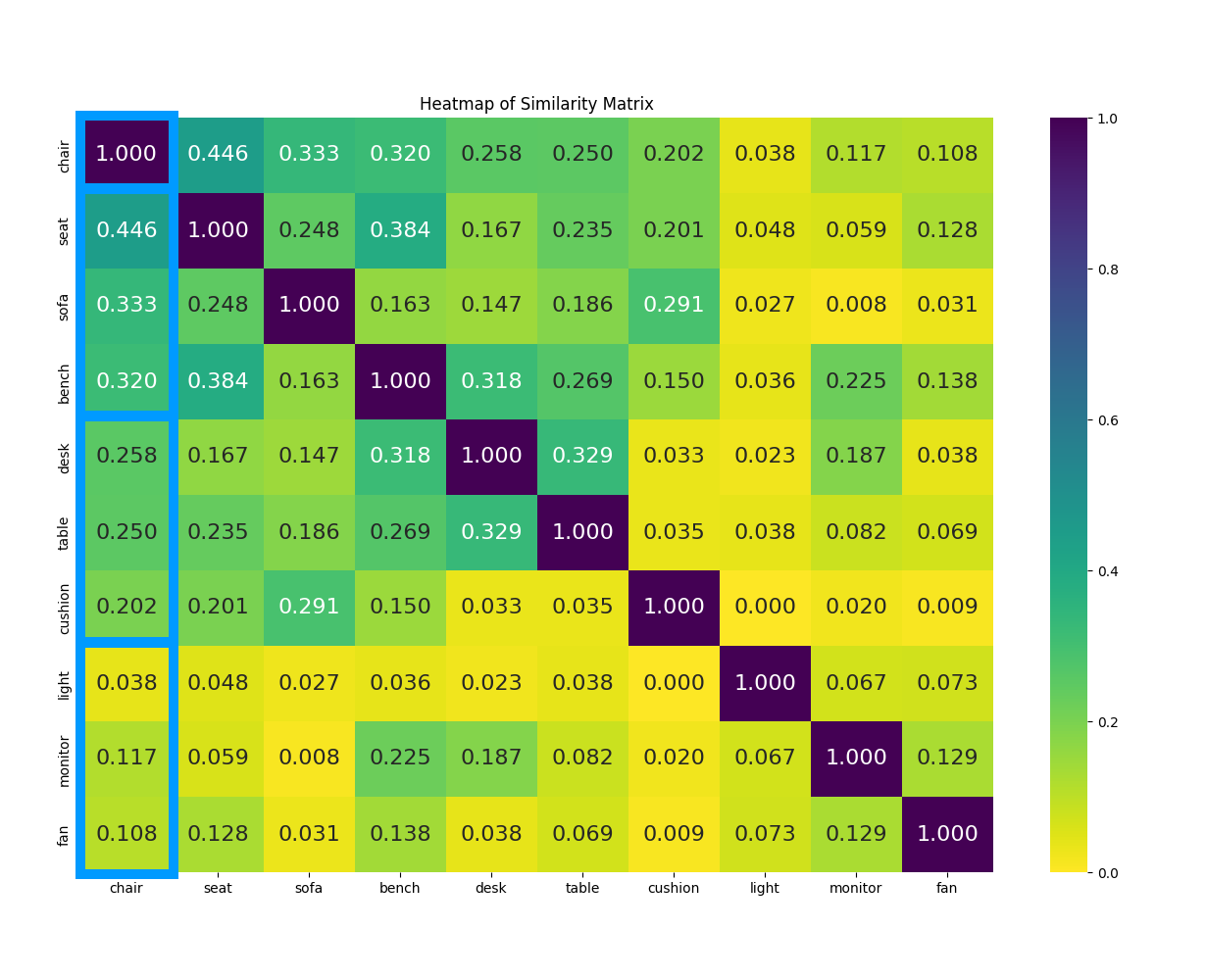}}
\caption{Heat Map of Similarity Matrix for Word Embedding}
\label{fig:heatmap_similarity_word}
\end{figure}

\begin{figure}[htbp]
\centerline{\includegraphics[width=1.0\textwidth]{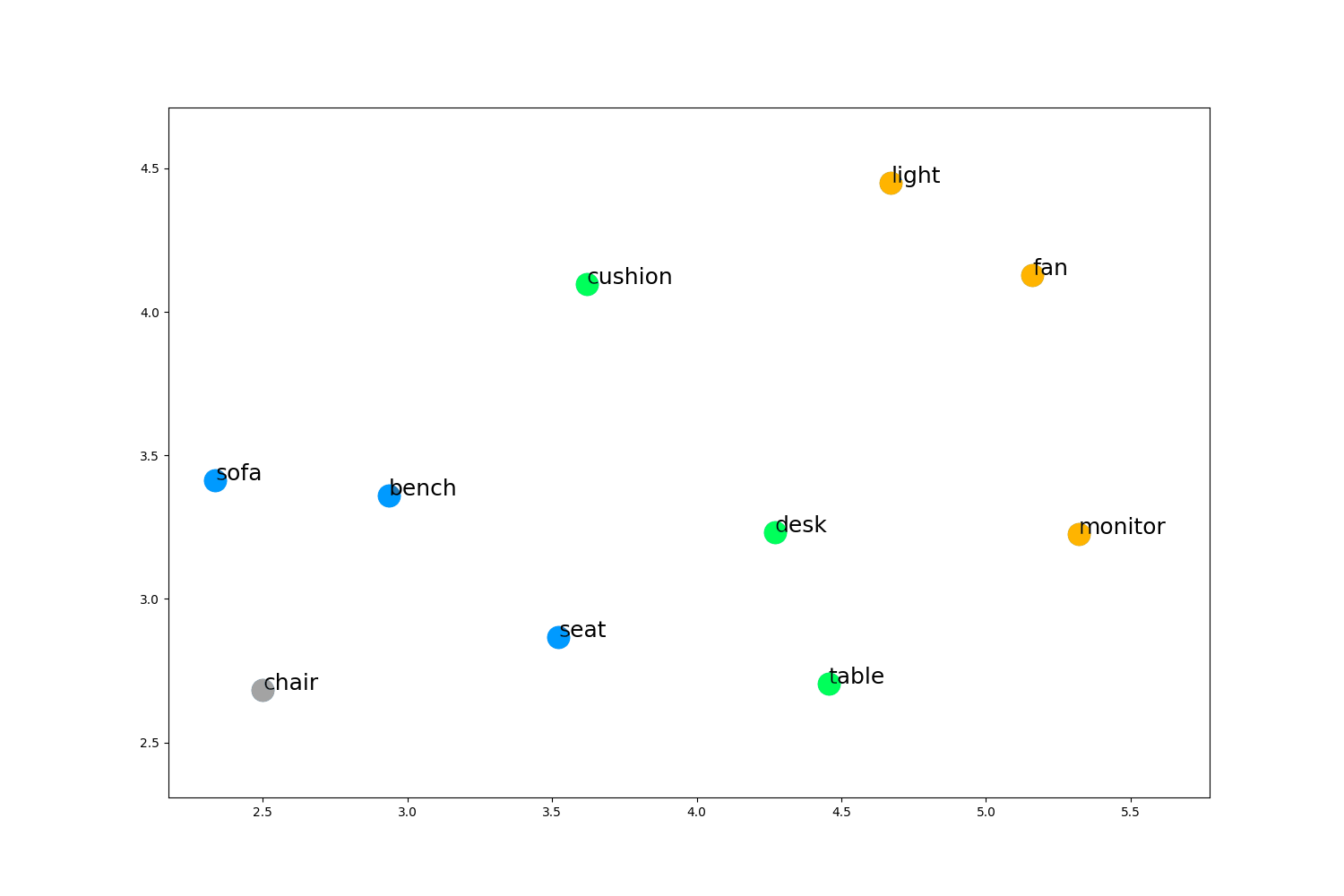}}
\caption{UMAP of Word Embedding}
\label{fig:umap_word}
\end{figure}

Embeddings for all these words were generated as $3072 x 1$ dimensional vectors using the TE3 embedding model, as shown in Table~\ref{tab:word_embeddings}.

\begin{table*}[]
    \setlength\extrarowheight{5pt}
    \centering
    \begin{tabular}{p{1cm}|p{16cm}}
        Words & Generated Vector Embeddings of 3072 x 1 dimension \\
        \hline
        chair & [-0.014324468, -0.019203374, -0.013582874, -0.027462386, \dots \dots \dots \dots 0.005507309, -0.011475187, 0.004418337, 0.017657736] \\
        \hline
        seat & [-0.018222768, -0.011229559, -0.002593147, -0.017478982, \dots \dots \dots \dots 0.017478982, -0.003989768, -0.023138227, 0.007466161] \\ 
        sofa & [-0.009118021, 0.006891137, -0.003268764, 0.028170729, \dots \dots \dots \dots -0.001830129, -0.017545667, -0.022681395, -0.005337789] \\
        bench & [-0.026748225, -0.000545074, -0.02579003, -0.011465845, \dots \dots \dots \dots 0.002626913, -0.017003881, -0.017621022, -0.001066803] \\
        \hline
        desk & [-0.001566614, -0.00333886, -0.017942928, -0.021080397, \dots \dots \dots \dots 0.022861123, -0.003298582, 0.00262445, 0.00787335] \\
        table & [0.000270125, -0.030253995, -0.016318725, 0.045984801, \dots \dots \dots \dots 0.037722155, 0.014578803, 0.004198854, 0.016700078] \\
        cushion & [0.010662668, -0.040683772, -0.003556379, 0.016528862, \dots \dots \dots \dots 0.016934318, 0.002203057, -0.021308083, 0.006099115] \\
        \hline
        light & [-0.005160935, -0.041870821, 0.01972012, 0.010678356, \dots \dots \dots \dots 0.023139138, -0.000709932, -0.003030125, -0.013951539] \\
        monitor & [-0.007739906, -0.00146106, -0.015165323, 0.0000751, \dots \dots \dots \dots 0.004533872, 0.009784079, -0.011574916, -0.016606728] \\
        fan & [-0.028556051, 0.00263738, -0.019429563, 0.003650993, \dots \dots \dots \dots 0.001529363, 0.016662996, -0.016424499, -0.006236698] \\
        \hline
    \end{tabular}
    \caption{Vectors Generated for Words using the TE3 Embedding Model}
    \label{tab:word_embeddings}
\end{table*}

The cosine similarity  between these vectors was calculated using the formula given below.
\[
\text{cosine similarity} = \cos(\theta) = \frac{\mathbf{A} \cdot \mathbf{B}}{|\mathbf{A}| |\mathbf{B}|} = \frac{\sum_{i=1}^{n} A_i B_i}{\sqrt{\sum_{i=1}^{n} A_i^2} \sqrt{\sum_{i=1}^{n} B_i^2}}
\]
and subsequently normalized to form a similarity matrix. The resulting heat map of this matrix is depicted in Figure~\ref{fig:heatmap_similarity_word}. The heat map clearly indicates that the word "chair" exhibits similarity scores in the range of 0.3 to 0.45 with the words in Set 1 compared to 0. to 0.25 for Set 2 and 0 to 0.1 for Set 3. As seen, the similarity scores are highest for Set 1, moderate for Set 2, and lowest for Set 3. This aligns with human judgment and demonstrates the effectiveness of embeddings in capturing semantic similarity.

To further analyze the spatial relationships between these word embeddings, a Uniform Manifold Approximation and Projection (UMAP) was calculated for all the embeddings to reduce the dimensionality from 3072 to 2. The resulting scatter plot is shown in Figure~\ref{fig:umap_word}. The scatter plot reveals that the embeddings of the words in Set 1 are positioned closer to the word "chair" than those in Set 2 and Set 3. This proximity indicates that the vectors for Set 1 words lie closer to the "chair" vector, followed by Set 2 and then Set 3. 

These results conclusively demonstrate that embeddings effectively capture semantic similarity between linguistic terms. The analysis shows that words with higher semantic similarity to "chair" are represented by vectors that are closer in the high-dimensional space, validating the robustness of embeddings in representing semantic relationships in the natural language (English language in the current context).

\begin{figure}[]
\centerline{\includegraphics[width=1.0\textwidth]{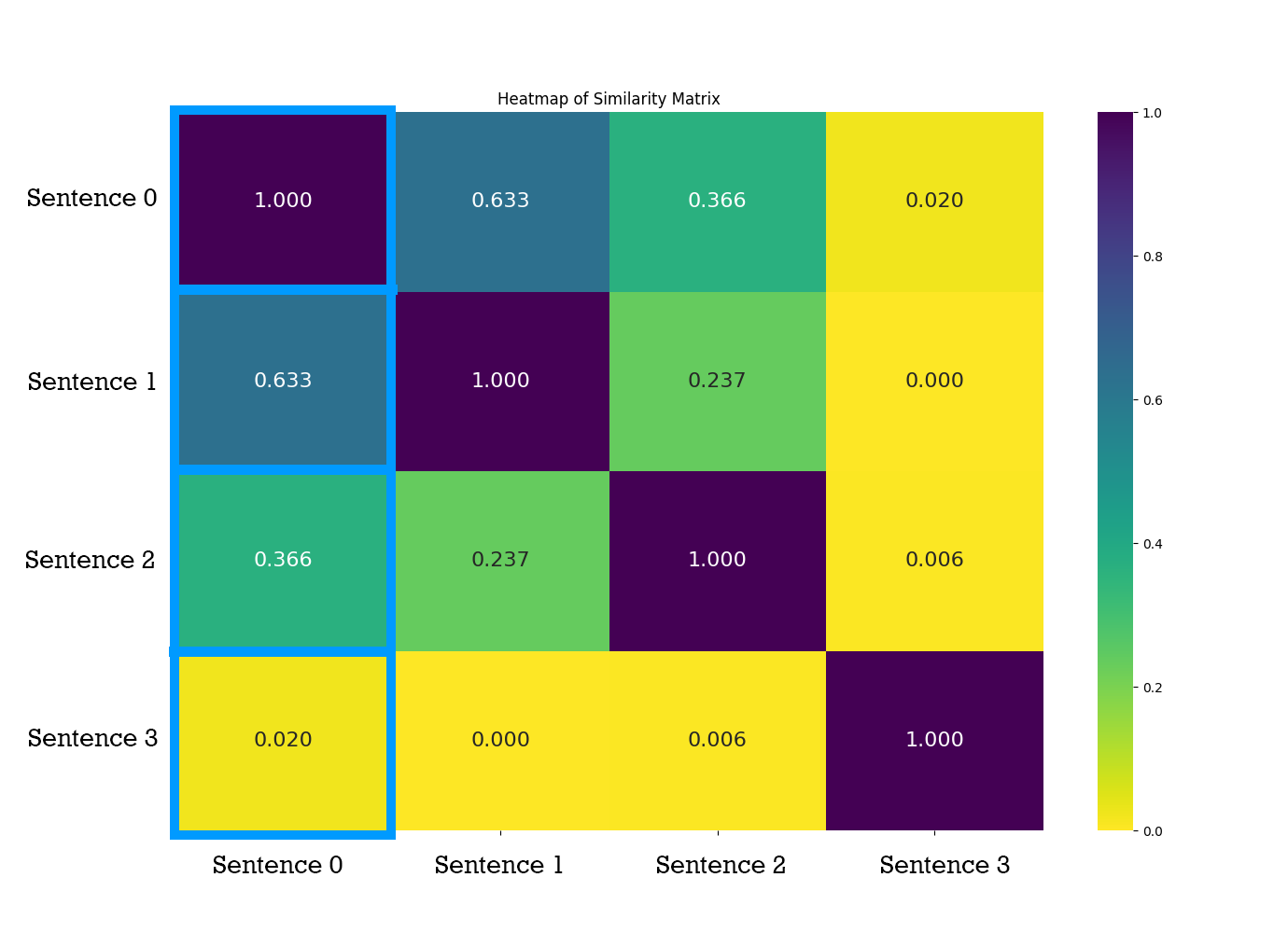}}
\caption{Heat Map of Similarity Matrix for Sentence Embedding}
\label{fig:heatmap_similarity_sentence}
\end{figure}

\begin{figure}[]
\centerline{\includegraphics[width=1.0\textwidth]{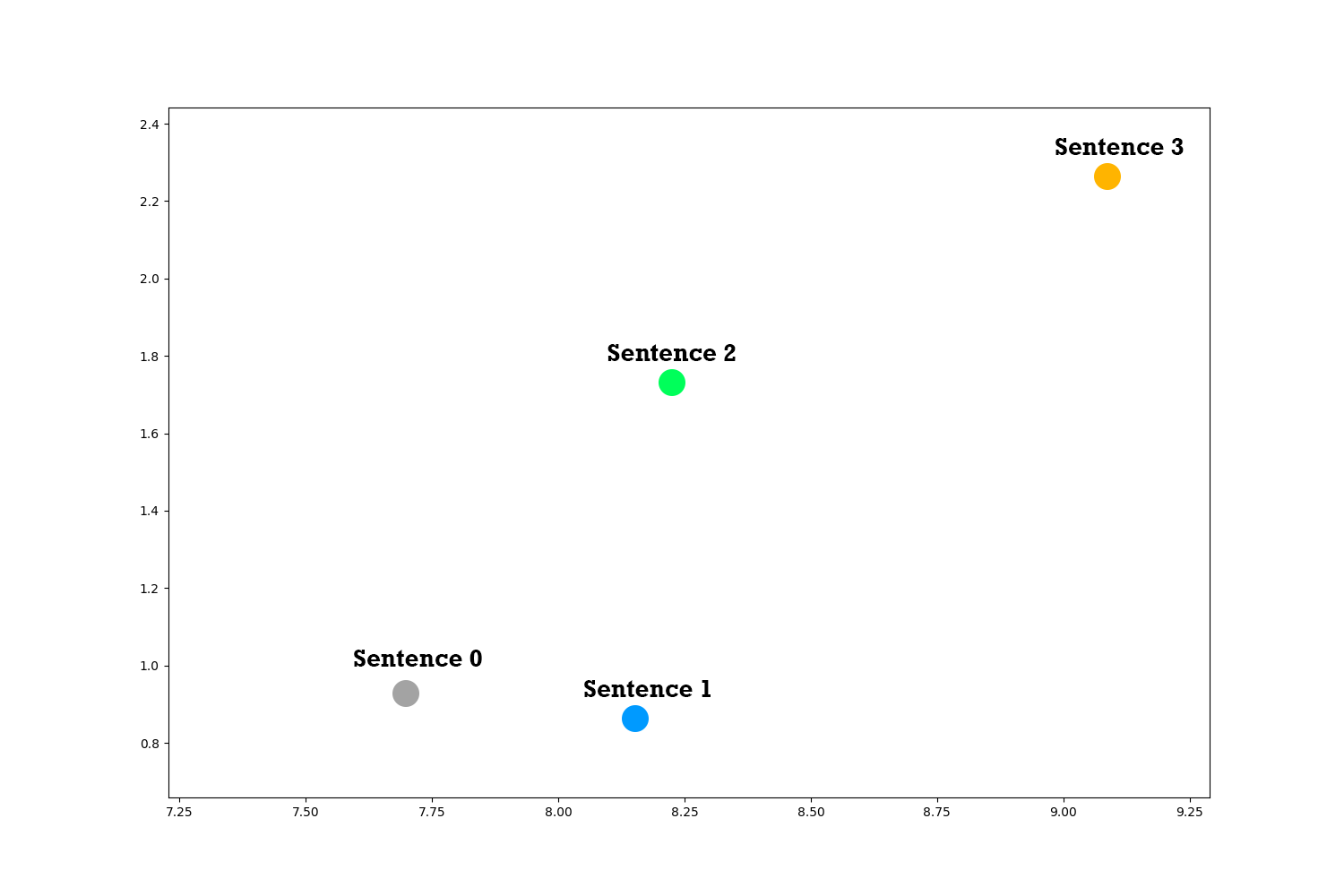}}
\caption{UMAP of Sentence Embedding}
\label{fig:umap_sentence}
\end{figure}

\subsubsection{Sentence Embedding}
To illustrate the concept of contextual similarity, we compared the sentence \emph{Sentence 0: "The chair was designed with ergonomic features to ensure user comfort."} against three other sentences using their corresponding embeddings. These sentences were selected to represent varying degrees of contextual similarity:
\begin{enumerate}
    \item Sentence 1 (High Similarity): "The seat incorporates ergonomic principles to maximize comfort."
    \item Sentence 2 (Moderate Similarity): "The desk was built to provide ample workspace and adjustable height."
    \item Sentence 3 (Low Similarity): "The light uses LED technology to provide energy-efficient lighting."
\end{enumerate}

Embeddings for these sentences were generated as $3072 x 1$ dimensional vectors using the TE3 embedding model, as shown in Table~\ref{tab:sentence_embeddings}.

\begin{table*}[]
    \setlength\extrarowheight{5pt}
    \centering
    \begin{tabular}{p{4cm}|p{13cm}}
        Sentences & Generated Vector Embeddings of 3072 x 1 dimension \\
        \hline
        The chair was designed with ergonomic features to ensure user comfort. & [-0.019350942, -0.00016339, -0.021010781, -0.025575334, \dots \dots \dots \dots 0.010235666, -0.00643533, -0.000910317, 0.013569174] \\
        \hline
        The seat incorporates ergonomic principles to maximize comfort. & [0.005157992, -0.021131463, -0.018314859, -0.032051004, \dots \dots \dots \dots 0.017260367, -0.02217208, 0.002526965, 0.02342082] \\ 
        \hline
        The desk was built to provide ample workspace and adjustable height. & [-0.018921912, 0.0000983, -0.010809345, -0.006264086, \dots \dots \dots \dots 0.006312243, -0.01546944, 0.005260202, 0.000995549] \\
        \hline
        The light uses LED technology to provide energy-efficient lighting. & [-0.011437991, -0.004313813, -0.013822543, -0.017071823, \dots \dots \dots \dots -0.013626014, 0.014582455, -0.023845525, 0.023360753] \\
        \hline
    \end{tabular}
    \caption{Vectors Generated for Sentences using the TE3 Embedding Model}
    \label{tab:sentence_embeddings}
\end{table*}

The normalized cosine similarity between these vectors was calculated as mentioned above. The resulting heat map of this matrix is depicted in Figure~\ref{fig:heatmap_similarity_sentence}.

The heat map clearly indicates that the sentence "The chair was designed with ergonomic features to ensure user comfort." exhibits a similarity score of 0.6 with the sentence in the high similarity category compared to 0.3 in the moderate and 0.02 in the low similarity categories.

The scatter plot of UMAP, as shown in Figure~\ref{fig:umap_sentence}, reveals that the embeddings of the sentences with high contextual similarity are positioned closer to the reference sentence. The sentences with moderate and low contextual similarity are progressively farther away.

These results conclusively demonstrate that sentence embeddings effectively capture contextual similarity between sentences. The analysis shows that sentences with higher contextual similarity to the original sentence are represented by vectors that are closer in the high-dimensional space, validating the robustness of embeddings in representing contextual relationships in the natural language (English language in the present context).

\hfill \hrule

\section*{Requirements of Idea Embedding for Designers}
It is demonstrated above that embeddings reasonably capture the semantic and contextual aspects of a statement. However, it is required to establish their utility for aiding designers and the design process. It can be appreciated that the task of sifting through a large volume of ideas generated during the ideation phase is challenging even for experts to manually identify and select a few novel and diverse ideas for further development. Also, assessing the effectiveness of the idea-generation phase helps in producing fruitful and innovative solutions. Therefore, a robust and objective framework to assess the result of ideation activity would be useful. To achieve this, we aim to address the following research questions.

\textbf{Semantic Validity of Idea Embedding}

The \emph{Idea Embeddings} are nothing but vector embeddings generated by a CAI system for the idea statements. However, the \emph{validity} of these embeddings needs to be established in the design community. The following two research questions are framed to address this.

\emph{RQ1.1: Meaningfulness: is idea embedding semantically acceptable to the designers?}

It is important to determine if these embeddings align with the designers' understanding and interpretation of the ideas. By confirming the semantic validity to be in line with the designer's expert opinion, it can be ensured that the quantitative metrics applied to these vectors reflect meaningful dimensions of the idea space, thus facilitating a more accurate and effective evaluation process.

\emph{RQ1.2: Usefulness: do idea embeddings help designers in selecting diverse ideas from a large pool of available ideas?}

Although it is desirable to have a large pool of ideas to improve the chance of a good/out-of-the-box solution, it is not possible to develop all the ideas towards a practical design. It is, therefore, important to efficiently shortlist diverse and innovative ideas from the given pool, preferably without imposing a significant cognitive burden on the designers. We believe that idea embeddings can be used to objectively assess and select ideas based on their disposition in the idea space.

\textbf{Characterization of Idea Space}

The goal of the ideation process is to generate innovative and diverse solutions. It is necessary to evaluate how well the idea space has been explored. A \emph{uniform and comprehensive} exploration ensures that no potential solutions are overlooked. The following two research questions are framed to address the need to assess the thoroughness of the idea-generation process.

\emph{RQ2.1: Distribution: Can the effectiveness of idea exploration be assessed objectively?}

This question assesses whether the ideas generated are uniformly distributed over the idea space. Uniform exploration is desirable as it would ensure that no region of the idea space is over-represented or under-represented. An over-representation can be a consequence of cognitive bottlenecks experienced by designers leading to similar ideas and would manifest as \emph{clusters} of ideas in the idea space.

\emph{RQ2.2: Dispersion: Can the comprehensiveness of the idea exploration be assessed objectively?}

This question investigates the \emph{extent} of the idea space that was covered during ideation. A high dispersion indicates high \emph{diversity} and \emph{comprehensiveness} of the exploration, which is desirable as it would imply that the generated ideas spanned the various dimensions of the idea space effectively.

\hfill \hrule

\section{Empirical Exploration of Idea Embedding}
To answer the above questions, a CAI-based method for the generation, representation, and analysis of ideas is adopted. The following paragraphs present the steps involved in this approach.

\subsection{Methodology}
The structured method of ideation using CAI has been described in Section~\nameref{sec:idea_gen_cai}. In the present study, we used six problem statements, each representing a distinct design challenge. For each problem statement, the CAI system generated 100 ideas using the structured Action-Object-Context (AOC) model as shown in Table~\ref{tab:ideas}. Then, we convert them into high-dimensional vectors using \emph{Text Embedding Model 3 (TE3)}, a dynamic embedding model; this generated 3072-dimensional vectors for each idea statement. Each value in the vector ranges from -1 to +1. Thus, all ideas are now mapped as points in the idea space represented as a high-dimensional hypercube of size 2, with the centre at the origin. The study of the distribution of the points in this cube constitutes the analysis of the disposition of ideas in the idea space.

We propose to use techniques such as \emph{UMAP (Uniform Manifold Approximation and Projection)} and \emph{DBSCAN (Density-Based Spatial Clustering of Applications with Noise)} for natural clustering of points in the idea space. UMAP is chosen for its ability to preserve both local and global proximity of the data, ensuring that the intrinsic structure of the idea space is maintained even after dimensionality reduction to 2D for ease of visualization. This ensures an accurate representation of their semantic and contextual similarities in a lower-dimensional space as well.

DBSCAN clusters points (ideas in the idea space) based on their proximity and density. This technique was chosen as it effectively constructs clusters of varying shapes and sizes; it has the additional benefit of being robust to noise and outliers. Since the proximity of points or ideas can be interpreted as semantic similarity, clusters here are interpreted as groups of similar ideas.

\begin{figure*}
\centering
 \begin{subfigure}[b]{0.49\textwidth}
    \centering
     \includegraphics[width=0.68\textwidth]{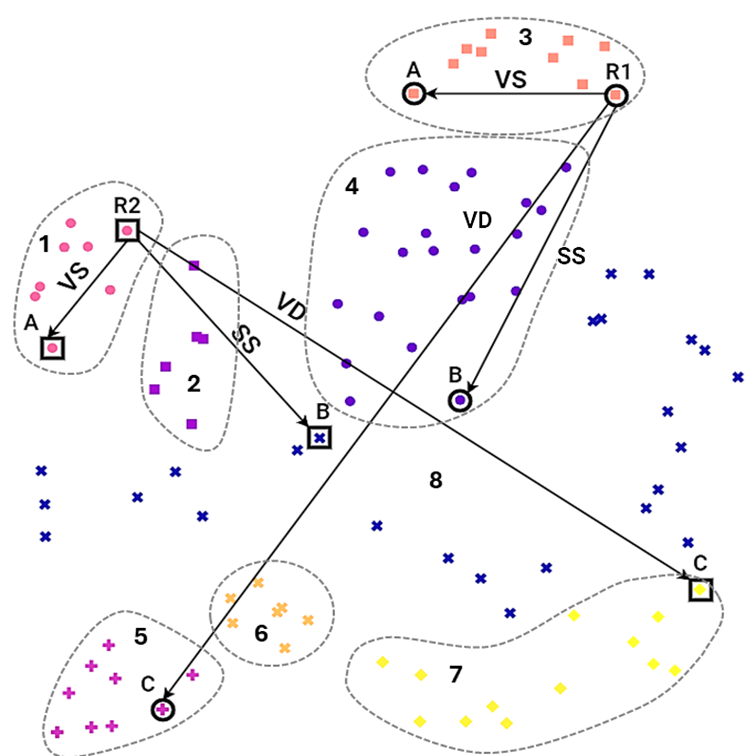}
     \caption{Idea Set 1}
     \label{fig:1_CTP}
 \end{subfigure}
 \hfill
 \begin{subfigure}[b]{0.49\textwidth}
     \centering
     \includegraphics[width=0.63\textwidth]{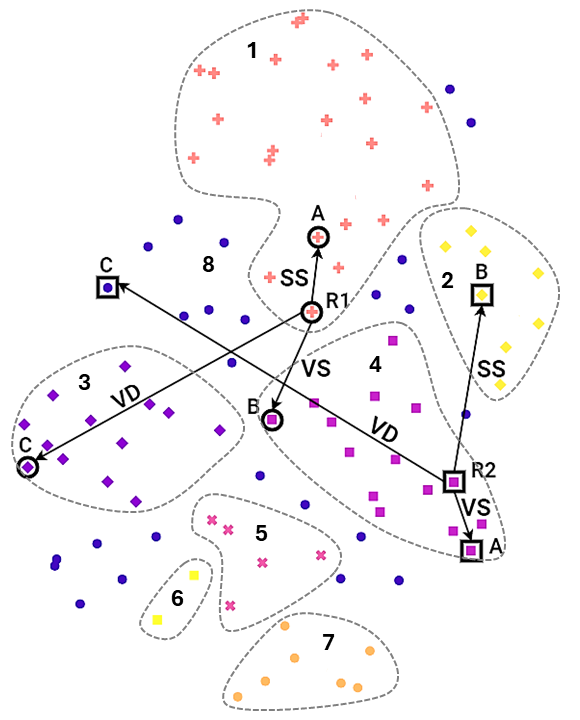}
     \caption{Idea Set 2}
     \label{fig:2_CTP}
 \end{subfigure}
 
 \vfill
 \begin{subfigure}[b]{0.49\textwidth}
     \centering
     \includegraphics[width=0.68\textwidth]{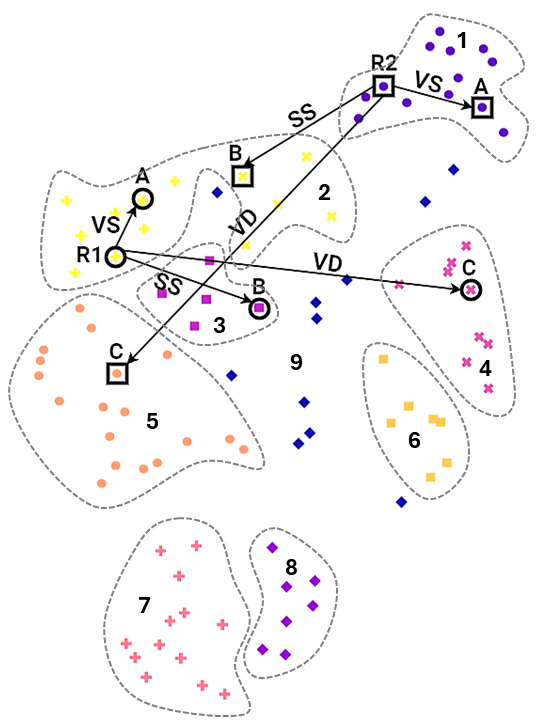}
     \caption{Idea Set 3}
     \label{fig:3_CTP}
 \end{subfigure}
 \hfill
 \begin{subfigure}[b]{0.49\textwidth}
     \centering
     \includegraphics[width=0.68\textwidth]{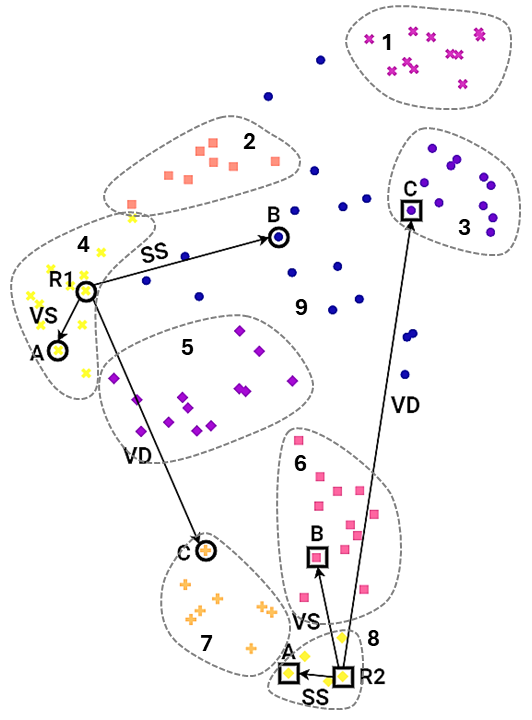}
     \caption{Idea Set 4}
     \label{fig:4_CTP}
 \end{subfigure}

  \vfill
 \begin{subfigure}[b]{0.49\textwidth}
     \centering
     \includegraphics[width=0.68\textwidth]{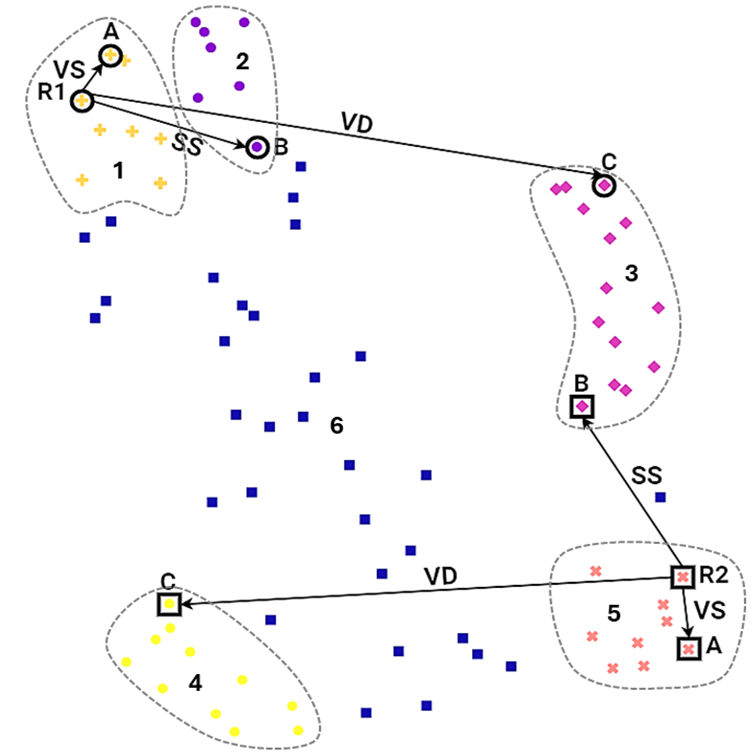}
     \caption{Idea Set 5}
     \label{fig:5_CTP}
 \end{subfigure}
 \hfill
 \begin{subfigure}[b]{0.49\textwidth}
     \centering
     \includegraphics[width=0.68\textwidth]{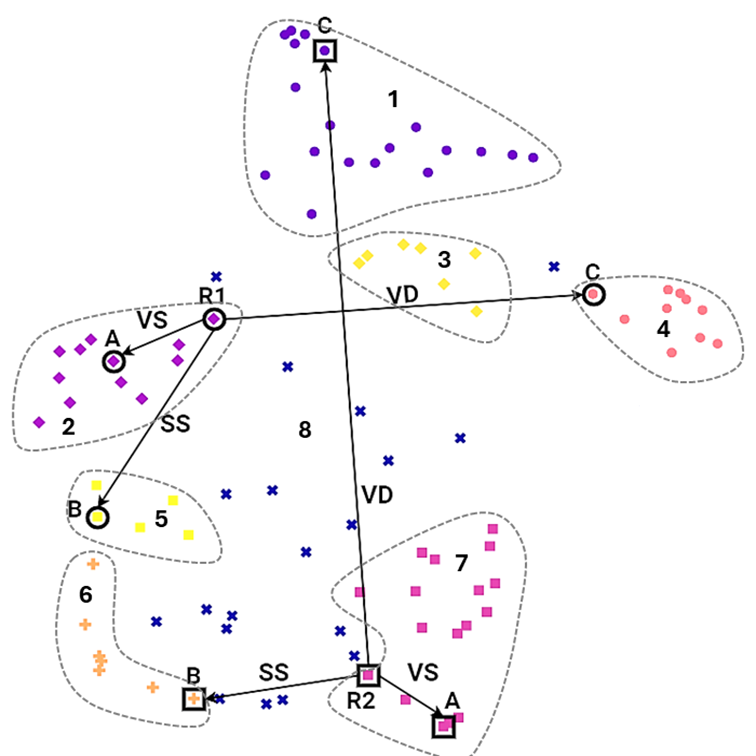}
     \caption{Idea Set 6}
     \label{fig:6_CTP}
 \end{subfigure}

 \caption{Cluster and Triad Plot of Vector Embeddings for Ideas Generated for the Six Problem Statements using DBSCAN and UMAP Algorithm}
 \label{fig:cluster_triad_plot}
 \footnote{\textbf{R1, R2} - Reference Idea Statement 1 \& 2;  \textbf{A} - Idea from the same cluster; \textbf{B} - Idea from the neighbouring cluster; \textbf{C} - Idea from far away cluster}
\footnote{\textbf{VS} - Very Similar; \textbf{SS} - Somewhat Similar; \textbf{VD} - Very Different}
\footnote{\textbf{\# }{1 to 9} - Cluster Index}

\end{figure*}

The clusters output by DBSCAN are demarcated using dotted lines (Figure~\ref{fig:cluster_triad_plot}). It is interesting to note that the result significantly matches the perceptual organization of the points.

\subsection{Meaningfulness of Idea Embeddings (RQ1.1)}
\label{subsec:meaningfulness}
To evaluate the meaningfulness of idea embeddings, a structured experiment was conducted. From each of the six idea sets, as shown in Figure~\ref{fig:cluster_triad_plot}, two ideas were randomly selected to serve as reference ideas (R1, R2); three additional ideas were selected for each of R1 and R2 as follows. Idea-A from the cluster of R1/R2, idea-B from a neighbouring cluster and idea-C from a distant cluster. This resulted in 12 groups, each comprising one reference idea and three additional ideas, as shown in Table~\ref{tab:googleform_ideas}. These groups were presented in a Google Form questionnaire wherein their cluster-based similarity levels were not disclosed to the respondents.

The questionnaire was then distributed to 30 expert designers, each with more than five years of experience in the design field. The experts were asked to assess idea-A, -B and -C in relation to the reference idea and indicate them as “Very Similar (VS)”, “Somewhat Similar (SS)”, or “Very Different (VD)”. The purpose of this evaluation was to determine whether the semantic similarities and differences captured by embeddings (clusters) aligned with the judgment of human experts.

\subsubsection{Results and Inference}
\begin{figure*}[h!]
\centering
 \begin{subfigure}[b]{0.49\textwidth}
    \centering
     \includegraphics[width=1.0\textwidth]{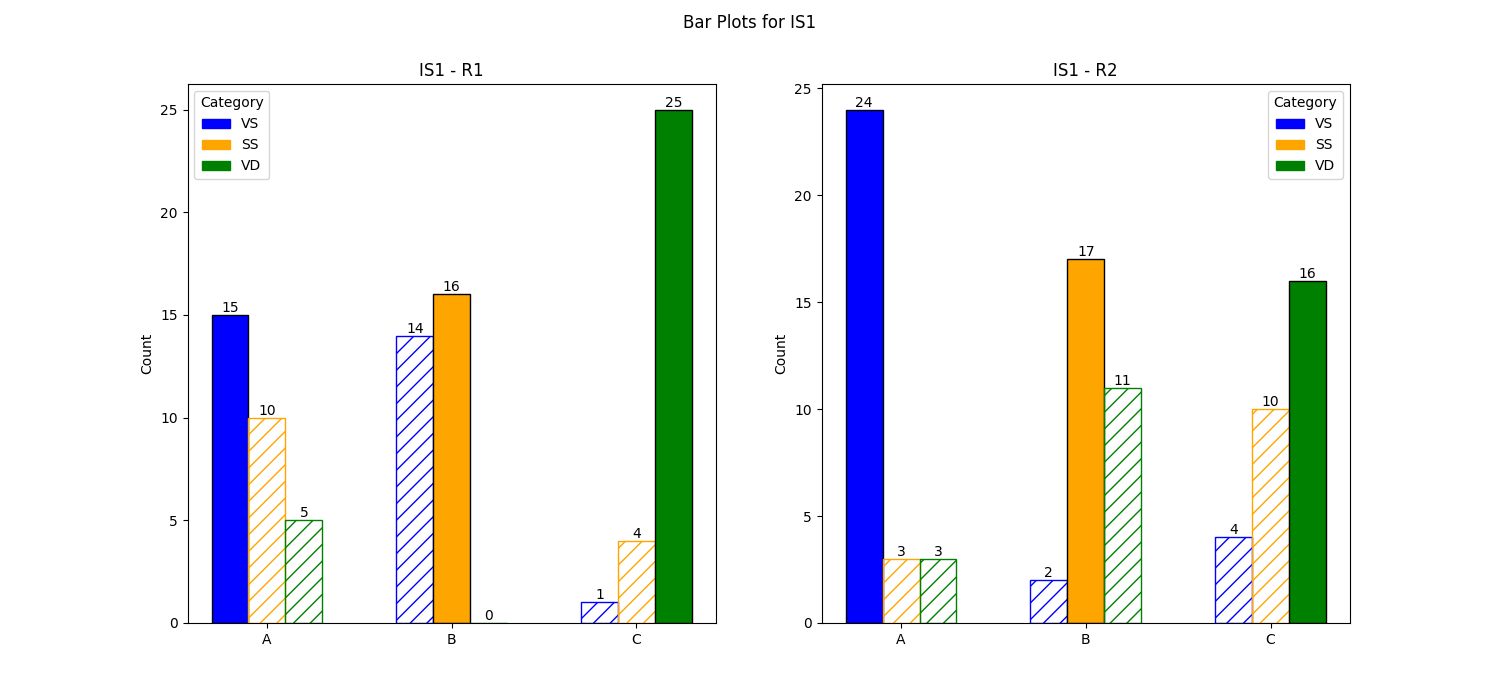}
     \caption{Idea Set 1}
     \label{fig:1_BPIS1}
 \end{subfigure}
 \hfill
 \begin{subfigure}[b]{0.49\textwidth}
     \centering
     \includegraphics[width=1.0\textwidth]{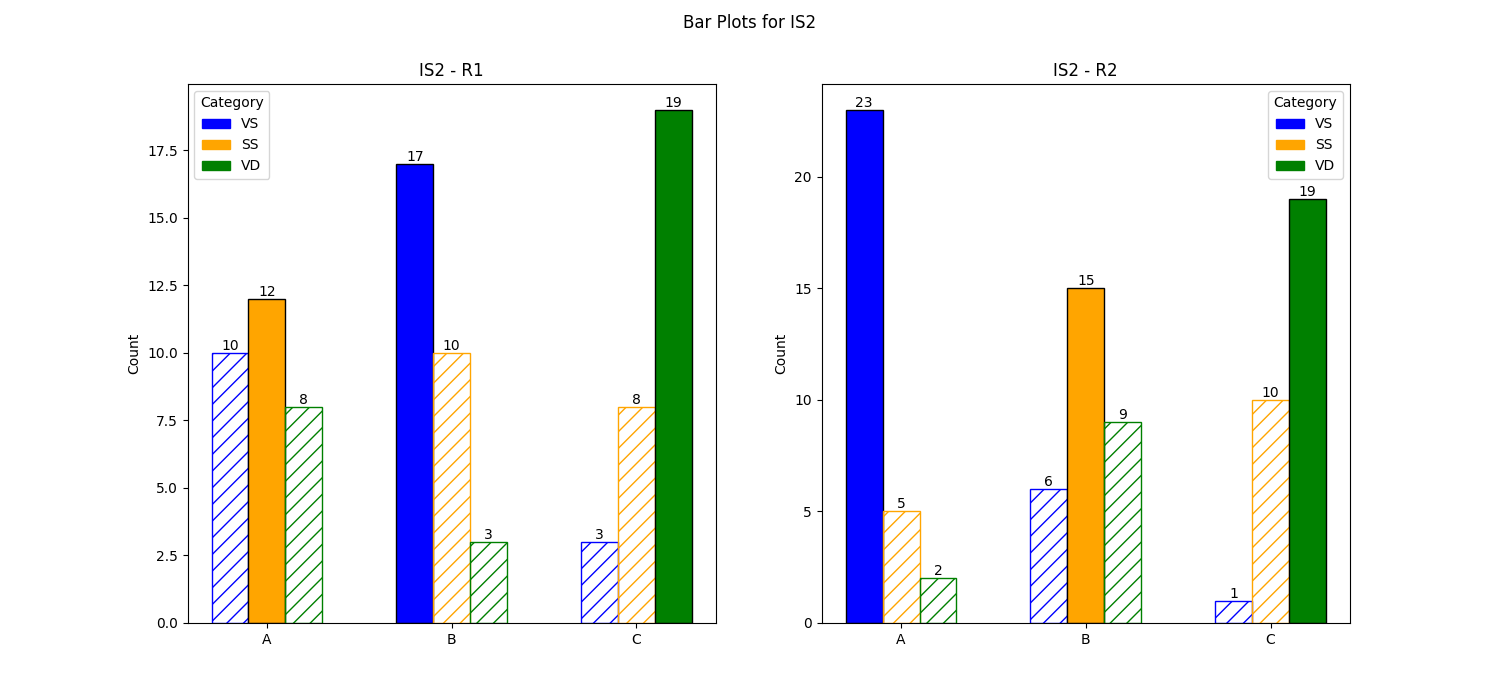}
     \caption{Idea Set 2}
     \label{fig:2_BPIS2}
 \end{subfigure}
 
 \vfill
 \begin{subfigure}[b]{0.49\textwidth}
     \centering
     \includegraphics[width=1.0\textwidth]{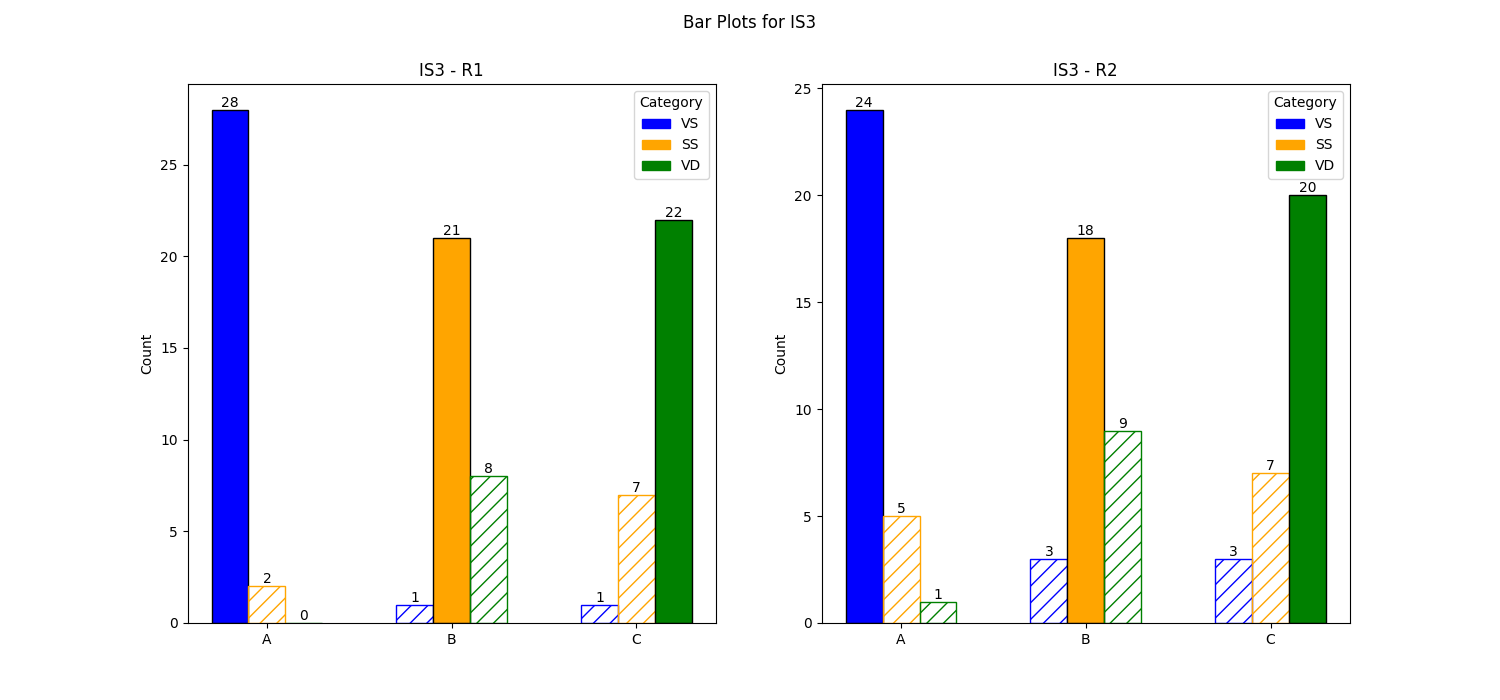}
     \caption{Idea Set 3}
     \label{fig:3_BPIS3}
 \end{subfigure}
 \hfill
 \begin{subfigure}[b]{0.49\textwidth}
     \centering
     \includegraphics[width=1.0\textwidth]{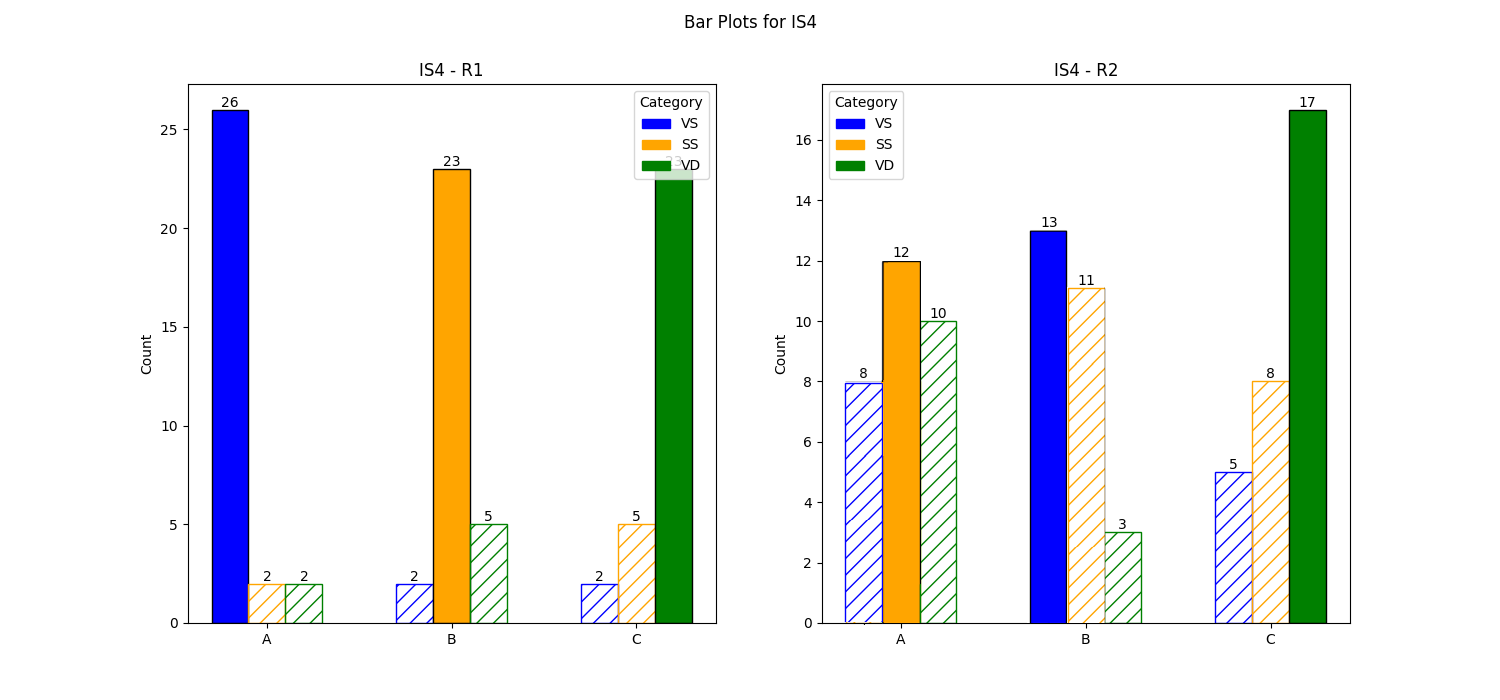}
     \caption{Idea Set 4}
     \label{fig:4_BPIS4}
 \end{subfigure}

  \vfill
 \begin{subfigure}[b]{0.49\textwidth}
     \centering
     \includegraphics[width=1.0\textwidth]{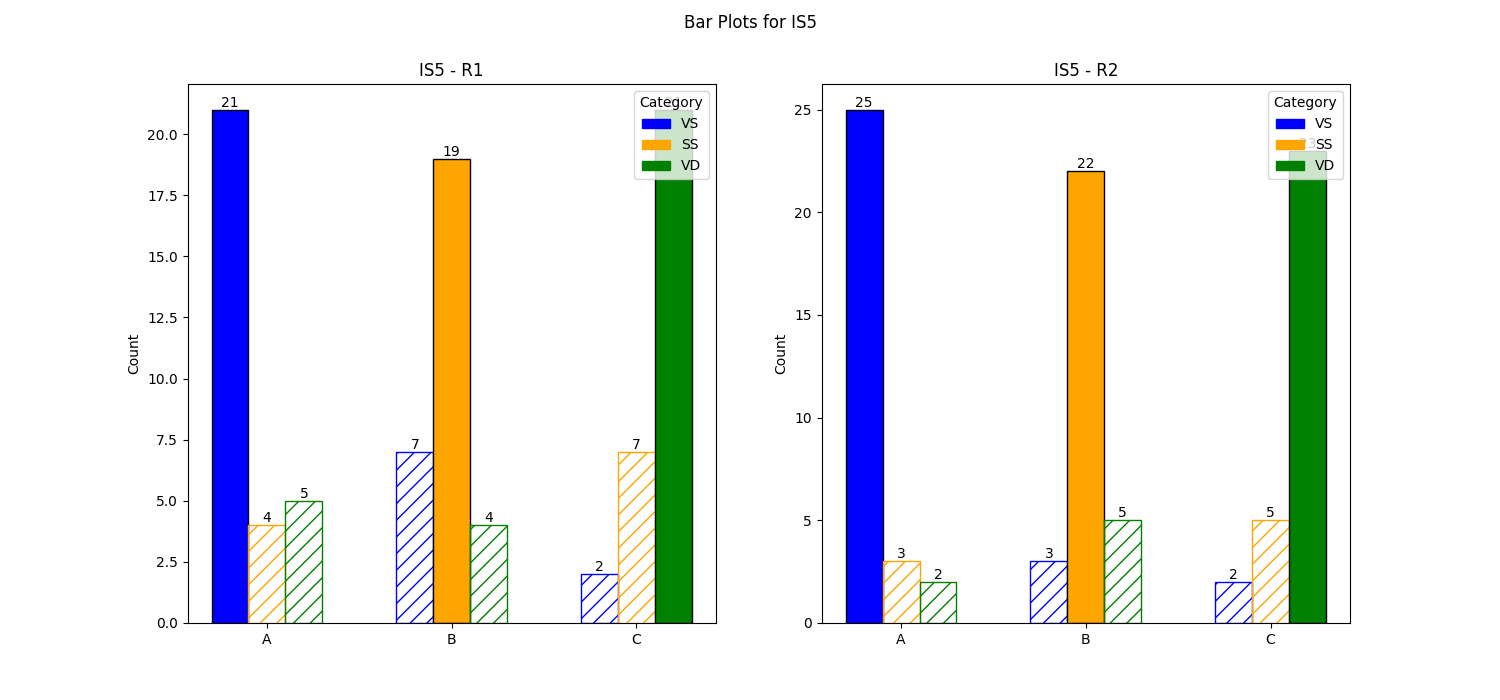}
     \caption{Idea Set 5}
     \label{fig:5_BPIS5}
 \end{subfigure}
 \hfill
 \begin{subfigure}[b]{0.49\textwidth}
     \centering
     \includegraphics[width=1.0\textwidth]{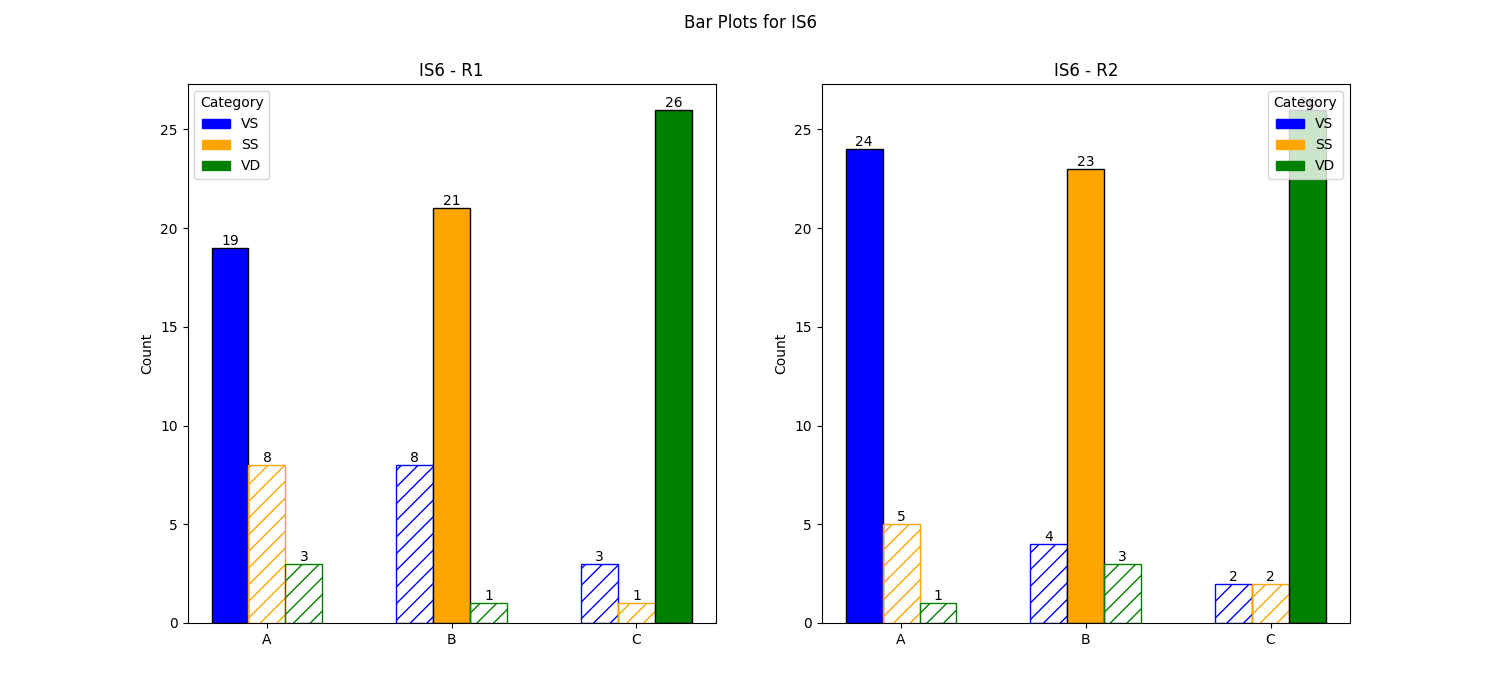}
     \caption{Idea Set 6}
     \label{fig:6_BPIS6}
 \end{subfigure}

 \caption{Bar Plot of Expert Responses from Questionnaire for Similarity of Ideas Generated for the Six Problem Statements}
 \label{fig:bar_plot_google_form}
 \footnote{\textbf{IS\#} - Idea Set [1/2/3/4/5/6]}
 \footnote{\textbf{R1, R2} - Reference Idea Statement 1 \& 2;  \textbf{A} - Idea very similar to R1/R2; \textbf{B} - Idea somewhat similar to R1/R2; \textbf{C} - Idea very different to R1/R2}
\footnote{\textbf{VS} - Very Similar; \textbf{SS} - Somewhat Similar; \textbf{VD} - Very Different}

\end{figure*}

The responses of the 30 experts are shown using bar plots in Figure~\ref{fig:bar_plot_google_form}. For each of the A, B and C, the dominant response (one with the highest vote) is highlighted with shading. The order of bars for each comparison idea is maintained as VS, SS, and VD from left to right.

To further elucidate the findings, the dominant responses identified by expert designers were marked on Figure~\ref{fig:cluster_triad_plot} using arrows. These arrows originate from the reference ideas (R1 and R2), point towards each idea (A, B, and C) and are labelled as VS, SS and VD as per the dominant response. It can be observed that the dominant categorization by experts aligned well with the cluster proximity except for the cases—R1 in Idea Set 2 and R2 in Idea Set 4—where the categories VS and SS were interchanged, the VD category still aligned with the expert judgment. This consistency indicates that the \emph{idea embeddings not only capture the semantic meaning of the ideas but also preserve the similarity relationships among them, falling within the human judgment}. Thus, the answer for RQ1.1 is \emph{Yes, the idea embedding is semantically acceptable to the designers}.


\begin{table*}[h!]
\centering
\begin{NiceTabular}{|p{0.5cm}|p{4.5cm}|p{11cm}|}
    \hline
    \rowcolor{lightgray}
    \textbf{Idea Set No.} & \textbf{Reference Idea (R)} & \textbf{Comparison Idea (A/B/C)}\\ 
    \hline
    \multirow{6}{*}{1} 
    & 
    \vspace{0.05em}
    \multirow{3}{=}{\textbf{R1}: Waste Streamline Funnel System: A funnel system that attaches to bins, guiding waste into the correct compartment based on its size and shape.} 
    & 
    \textbf{A}: Smart Trash Compactors:	Trash compactors that can detect the type of waste and compact it accordingly, reducing volume and facilitating recycling.
 \\ \cline{3-3}
    & & 
    \textbf{B}: Multi-Layer Segregation Shelves: Stackable shelves with labels for different types of waste, encouraging organized segregation in homes and offices.
 \\ \cline{3-3}
    & & 
    \textbf{C}: Zero Waste Starter Packs: Kits that include reusable items and information on how to start and maintain a zero-waste lifestyle.
 \\ \cline{2-3}
    & 
    \vspace{0.05em}
    \multirow{3}{=}{\textbf{R2}: Gamified Recycling Bins: Bins designed with game-like features that reward users with points or digital tokens for correctly segregating their waste.} 
    & 
    \textbf{A}: Segregation-Encouraging Trash Receptacles:	Design trash receptacles with engaging visuals and sounds that reward users for proper segregation.
 \\ \cline{3-3}
    & & 
    \textbf{B}: Waste Segregation Reminder System: A system that reminds users to segregate their waste at the disposal point through visual cues or audio messages.
 \\ \cline{3-3}
    & & 
    \textbf{C}: Waste Type Projection System: A system that projects images onto the ground or bins to indicate where different types of waste should be thrown.
 \\ \hline

    \multirow{6}{*}{2} 
    & 
    \vspace{0.05em}
    \multirow{3}{=}{\textbf{R1}: Multi-Tiered Carousel Umbrella Dryer:	Suitable for commercial spaces or public areas with heavy foot traffic needing efficient umbrella turnover.} 
    & 
    \textbf{A}: Hydraulic Umbrella Compression Dryer:	Designed for luxury hotels or high-end residential buildings looking for an efficient and delicate drying solution.
 \\ \cline{3-3}
    & & 
    \textbf{B}: Expandable Umbrella Drying Tunnel:	Designed for offices or residential buildings with many users and limited space.
 \\ \cline{3-3}
    & & 
    \textbf{C}: Solar-Powered Outdoor Umbrella Stand:	Ideal for eco-friendly outdoor storage solutions like patios or gardens.
 \\ \cline{2-3}
    & 
    \vspace{0.05em}
    \multirow{3}{=}{\textbf{R2}: Detachable Umbrella Drying Liner: Designed for convenience and rapid drying of personal umbrellas after arriving indoors.} 
    & 
    \textbf{A}: Umbrella Quick-Dry Sleeve:	Convenient for on-the-go individuals who need to store their umbrellas quickly after use.
 \\ \cline{3-3}
    & & 
    \textbf{B}: Modular Umbrella Rack with Drainage:	Targeted at public places or businesses with a high volume of umbrella usage.
 \\ \cline{3-3}
    & & 
    \textbf{C}: Umbrella Drying Lamp Post:	Ideal for outdoor venues, stylishly combining lighting and drying in one solution.
 \\ \hline

    \multirow{6}{*}{3} 
    & 
    \vspace{0.05em}
    \multirow{3}{=}{\textbf{R1}: Shoe Disinfection Powder:	A powder formulated to absorb excess moisture inside the shoes, which also contains antimicrobial agents to prevent bacterial and fungal growth.} 
    & 
    \textbf{A}: Shoe Disinfection Quick-Dry Spray:	A spray that not only disinfects the shoes but also contains agents that speed up the drying process.
 \\ \cline{3-3}
    & & 
    \textbf{B}: Shoe Disinfection Bubble Bath:	A playful bubble bath for shoes that cleans and disinfects with antibacterial soap bubbles.
 \\ \cline{3-3}
    & & 
    \textbf{C}: Shoe Sanitization and Polishing Robot:	A robotic device that cleans and disinfects shoes while polishing them.
 \\ \cline{2-3}
    & 
    \vspace{0.05em}
    \multirow{3}{=}{\textbf{R2}: Footwear Disinfection Walkway:	A walkway with built-in disinfection mechanisms for public spaces like malls or airports.} 
    & 
    \textbf{A}: Shoe Disinfection Conveyor Belt: A conveyor belt system that cleans shoes as people walk through a checkpoint or entrance.
 \\ \cline{3-3}
    & & 
    \textbf{B}: Shoe Sole Disinfection Wrap:	A wrap-around device that encases the shoe sole and applies a disinfectant solution for quick cleaning.
 \\ \cline{3-3}
    & & 
    \textbf{C}: Natural Herbal Shoe Disinfectant:	A disinfectant that uses herbal extracts with natural antibacterial and antifungal properties to cleanse shoes safely and organically.
 \\ \hline

    \multirow{6}{*}{4} 
    & 
    \vspace{0.05em}
    \multirow{3}{=}{\textbf{R1}: Smart Dishwashing Assistant:	An AI-powered device that provides real-time guidance on water usage, detergent dosage, and load arrangement to optimize dishwashing efficiency.} 
    & 
    \textbf{A}: Dishwasher Efficiency Monitor:	A device that monitors dishwasher usage and provides recommendations on improving efficiency, such as load size and detergent amount.
 \\ \cline{3-3}
    & & 
    \textbf{B}: Automated Pre-soak Dispenser:	Dispenses the right amount of pre-soak solution into the sink before dishwashing to loosen debris and reduce scrubbing effort.
 \\ \cline{3-3}
    & & 
    \textbf{C}: Hydrophobic Dish Coating:	A nanotechnology-based coating that can be applied to dishes, making them resistant to water and food residues, thus reducing the need for scrubbing and detergent use.
 \\ \cline{2-3}
    & 
    \vspace{0.05em}
    \multirow{3}{=}{\textbf{R2}: Ultrasonic Dish Cleaner:	A device that utilizes ultrasonic waves to agitate water and remove food particles from dishes without the need for scrubbing, saving time and reducing water consumption.} 
    & 
    \textbf{A}: Compact Dish Sterilizer:	A small, energy-efficient device that uses UV light to sterilize dishes after washing, ensuring they are bacteria-free and ready for safe use.
 \\ \cline{3-3}
    & & 
    \textbf{B}: Integrated Dish Pre-Rinser:	A sink attachment that pre-soaks and sprays dishes to remove food particles, easing the cleaning process and lowering water use.
 \\ \cline{3-3}
    & & 
    \textbf{C}: Dishwashing Gloves with Integrated Scrubbers:	A pair of dishwashing gloves with built-in scrubbing bristles, allowing for efficient cleaning without additional sponges or brushes.
 \\ \hline

    \multirow{3}{*}{5} 
    & 
    \vspace{0.05em}
    \multirow{3}{=}{\textbf{R1}: Pressure-Relief Insoles:	Provides comfort to feet during long periods of standing.} 
    & 
    \textbf{A}: Personalized Comfort Footpad:	A footpad with memory foam or smart materials that molds to the user's feet for maximum comfort.
 \\ \cline{3-3}
    & & 
    \textbf{B}: Portable Foot Roller:	Relieves standing fatigue.
 \\ \cline{3-3}
    & & 
    \textbf{C}: Retractable Seating Brace:	A wearable device that extends to form a temporary seat.
 \\ \hline

    \multicolumn{3}{r}{Continued in Next Page}
\\ \hline

\end{NiceTabular}
\caption{Shortlisted Ideas for Questionnaire on Evaluation of Similarity}
\label{tab:googleform_ideas}
\end{table*}

\begin{table*}[h!]
\centering
\begin{NiceTabular}{|p{0.5cm}|p{4.5cm}|p{11cm}|}
    \hline
    \multicolumn{3}{l}{Continued from Previous Page}
\\ \hline
    \rowcolor{lightgray}
    \textbf{Idea Set No.} & \textbf{Reference Idea (R)} & \textbf{Comparison Idea (A/B/C)}
\\ \hline

    \multirow{3}{*}{5} 
    & 
    \vspace{0.05em}
    \multirow{3}{=}{\textbf{R2}: Queue Wait Optimizer:	A system that uses data analytics to advise users on the best times to queue up, reducing wait times.} 
    & 
    \textbf{A}: Queue Space Optimizer:	A system that adjusts queue layouts for optimal space usage and flow.
\\ \cline{3-3}
    & & 
    \textbf{B}: Queue Space Marker:	A portable device that projects a boundary on the floor to indicate personal space in crowded lines.
 \\ \cline{3-3}
    & & 
    \textbf{C}: Ergonomic Queue Flooring:	Ergonomically designed flooring that reduces stress on legs and back.
 \\ \hline

     \multirow{6}{*}{6} 
    & 
    \vspace{0.05em}
    \multirow{3}{=}{\textbf{R1}: Tranquil Trek Trail:	A series of bird feeders placed along a garden path to encourage gentle exercise and bird watching, promoting physical and mental health.} 
    & 
    \textbf{A}: Garden Flight Paths:	Designs in the garden that attract specific species to feeding spots, offering a visually stimulating activity for the elderly.
 \\ \cline{3-3}
    & & 
    \textbf{B}: Sculpture Garden Sphere:	A bird feeder that doubles as a garden sculpture, with aesthetically pleasing designs that change with user interaction, promoting creativity and visual enjoyment.
 \\ \cline{3-3}
    & & 
    \textbf{C}: Nostalgic Nest Viewer:	A birdhouse that includes a digital screen to display old photos or videos, triggering happy memories as birds come and go.
 \\ \cline{2-3}
    & 
    \vspace{0.05em}
    \multirow{3}{=}{\textbf{R2}: Calm Canopy Companion:	A bird feeder that plays calming forest sounds to create a serene atmosphere, promoting relaxation and stress relief.} 
    & 
    \textbf{A}: Serenity Stream Sanctuary:	A bird feeder that includes a small, gentle water stream or fountain, providing a calming auditory experience and attracting birds with fresh water.
 \\ \cline{3-3}
    & & 
    \textbf{B}: Harmony Habitat Lantern:	A bird feeder with a built-in lantern that emits a soft, warm glow as the evening sets in, providing a sense of security and tranquillity.
 \\ \cline{3-3}
    & & 
    \textbf{C}: Pastime Puzzle Perch:	A bird feeder that features interchangeable puzzle pieces or brainteasers that the elderly can solve, stimulating cognitive function and providing a sense of achievement.
 \\ \hline

\end{NiceTabular}
\end{table*}

\subsection{Usefulness of Idea Embeddings (RQ1.2)}
To assess the practical utility of the perceptually clustered idea presentation in selecting a small subset of ideas from a large pool, a study was conducted involving 40 graduate design students (non-experts). The aim of this study was to observe both individual and group behaviours in the selection process, viz., how the visually clustered idea representation influences novice designers to select diverse ideas.

Multiple copies of each plot shown in Figure~\ref{fig:cluster_triad_plot} were printed on paper and distributed equitably to the participants. The clusters in each plot were marked with different colours to indicate the similarity of ideas within each cluster. Each plot had 100 marked points representing 100 ideas. The participants were required to select 10 arbitrary points/ideas from 100 in 10 minutes. Textual descriptions of the ideas or the associated problems were not disclosed. However, they were informed that the clusters represented similar ideas. The selected ideas were marked with a pen on the paper. 

\subsubsection{Results and Inference}
For each plot, the votes for each cluster were recorded; the division of this vote among the participants was also noted. For the purpose of analysis, the area of the convex hull of each cluster in each plot is also calculated. 

\begin{figure*}[h!]
\centering
 \begin{subfigure}[b]{0.49\textwidth}
    \centering
     \includegraphics[width=1\textwidth]{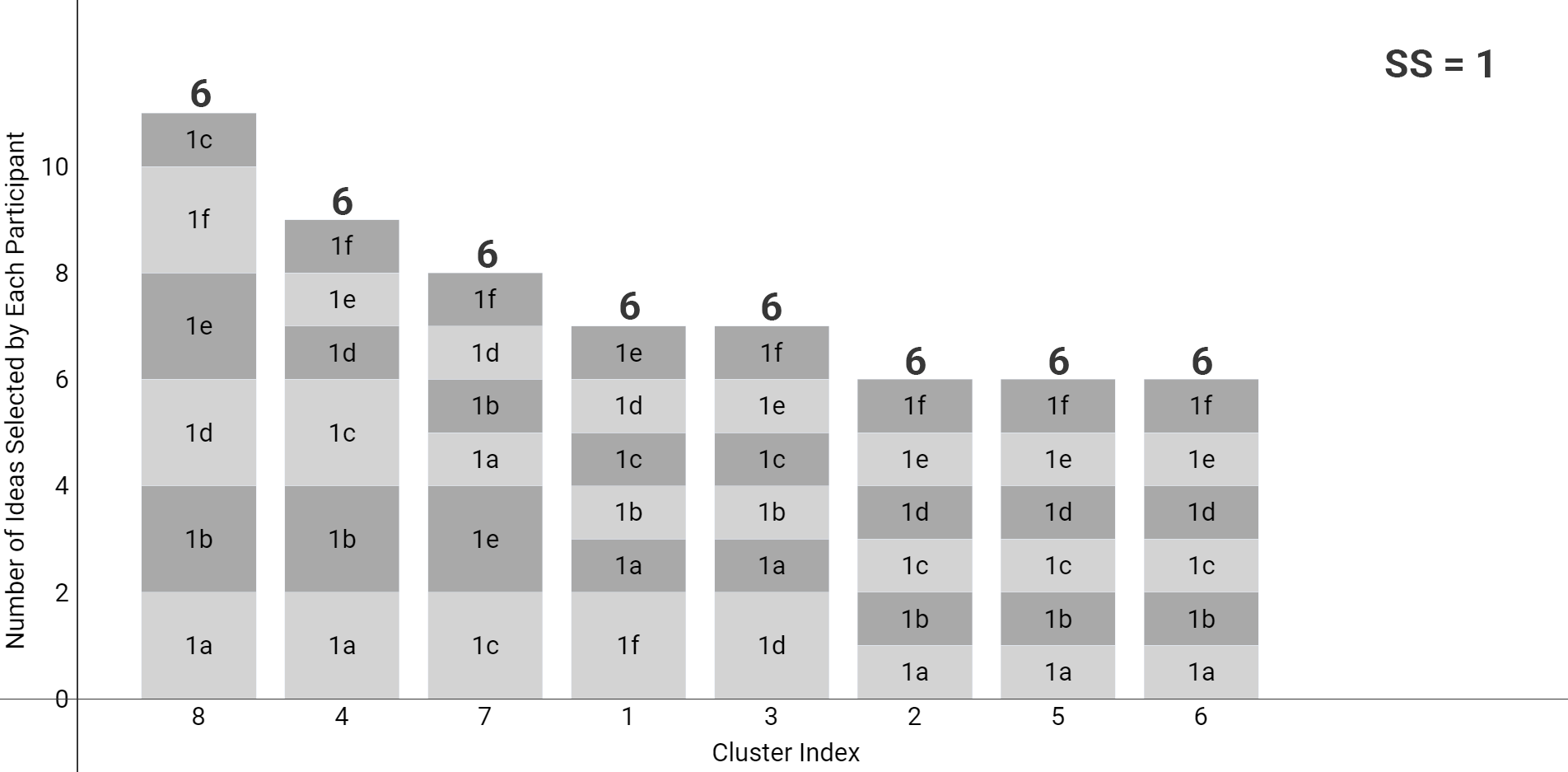}
     \caption{Idea Set 1}
     \label{fig:1_ISEL}
 \end{subfigure}
 \hfill
 \begin{subfigure}[b]{0.49\textwidth}
     \centering
     \includegraphics[width=1\textwidth]{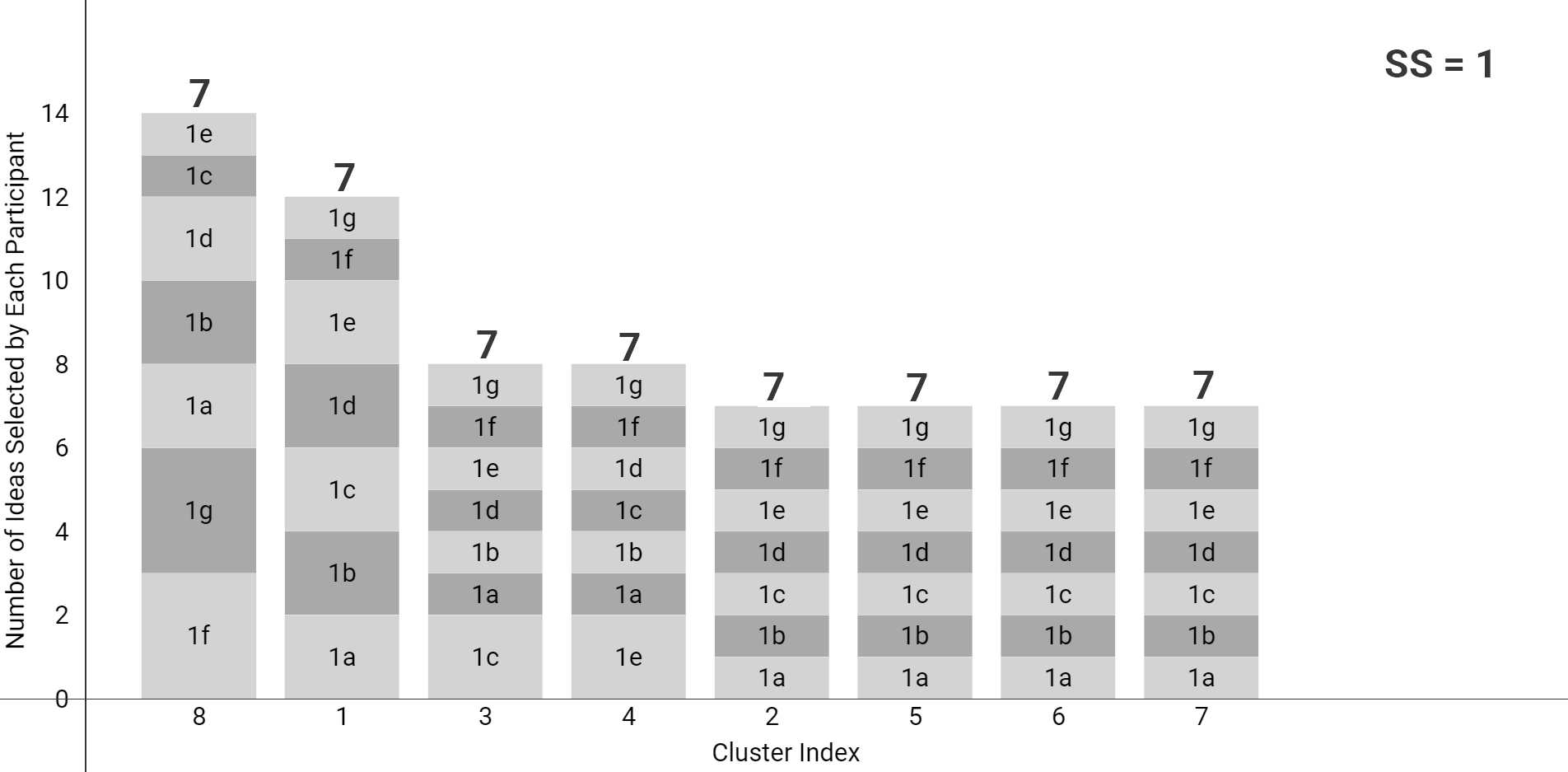}
     \caption{Idea Set 2}
     \label{fig:2_ISEL}
 \end{subfigure}
 
 \vfill
 \begin{subfigure}[b]{0.49\textwidth}
     \centering
     \includegraphics[width=1\textwidth]{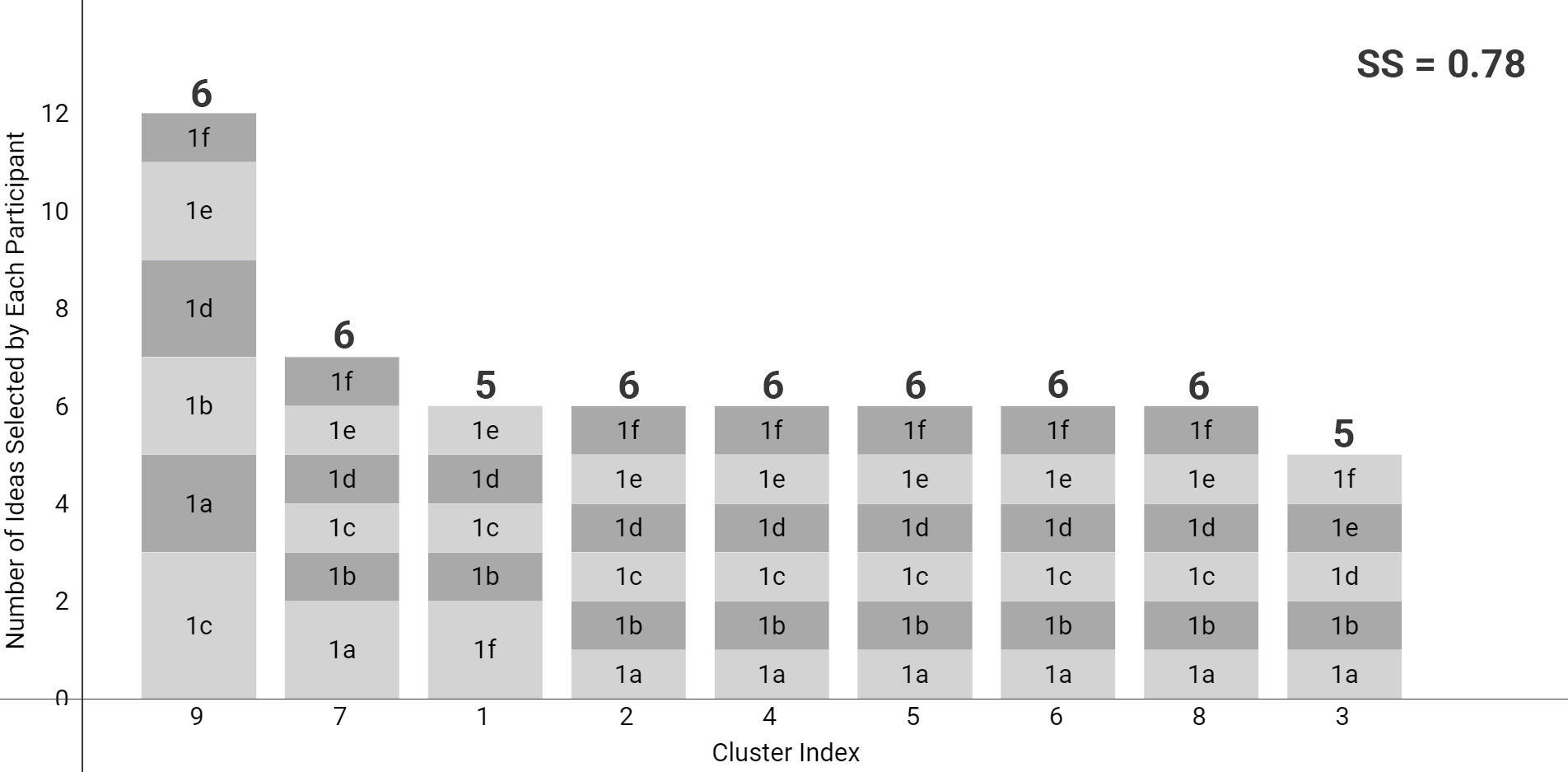}
     \caption{Idea Set 3}
     \label{fig:3_ISEL}
 \end{subfigure}
 \hfill
 \begin{subfigure}[b]{0.49\textwidth}
     \centering
     \includegraphics[width=1\textwidth]{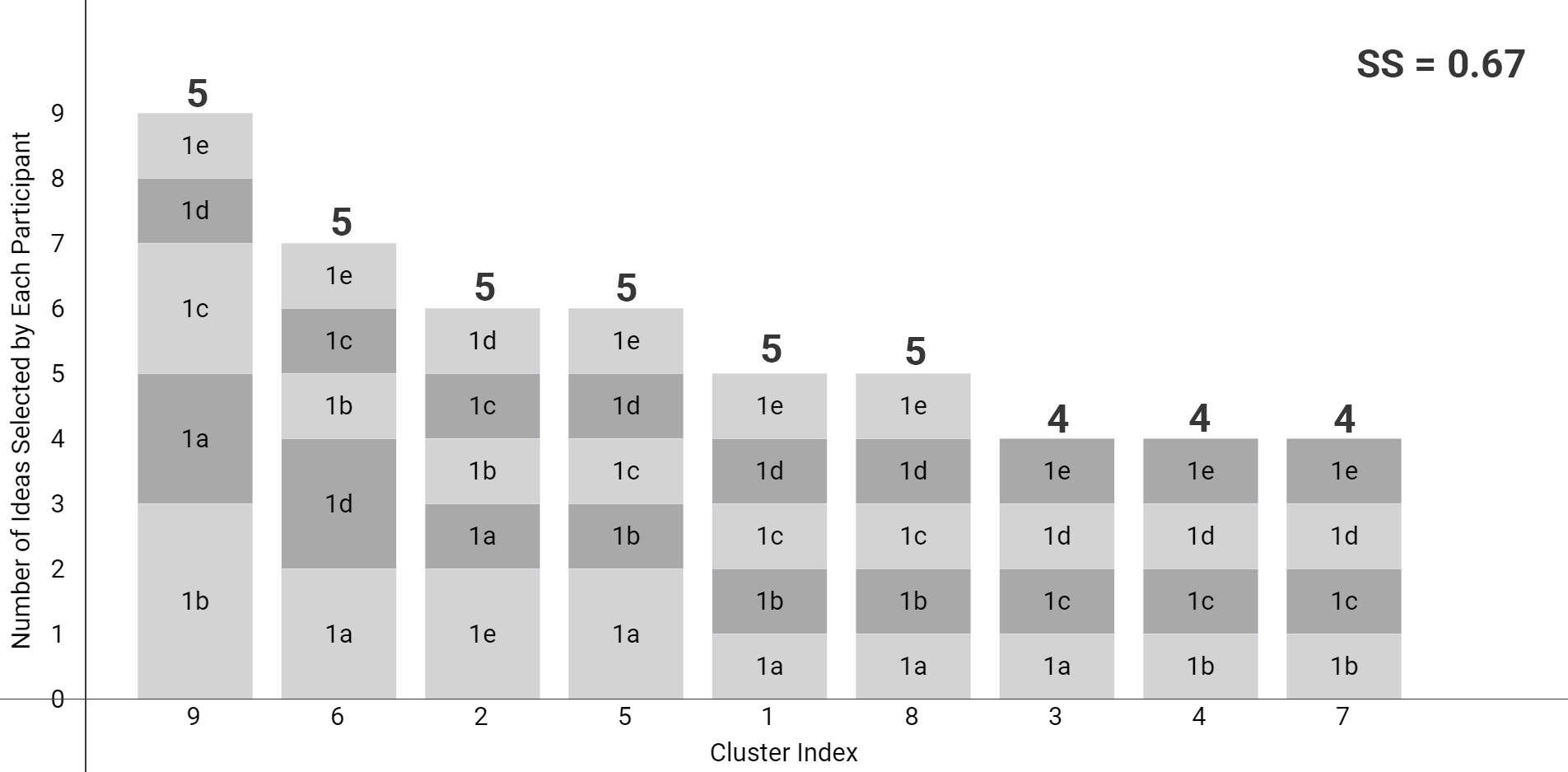}
     \caption{Idea Set 4}
     \label{fig:4_ISEL}
 \end{subfigure}

  \vfill
 \begin{subfigure}[b]{0.49\textwidth}
     \centering
     \includegraphics[width=1\textwidth]{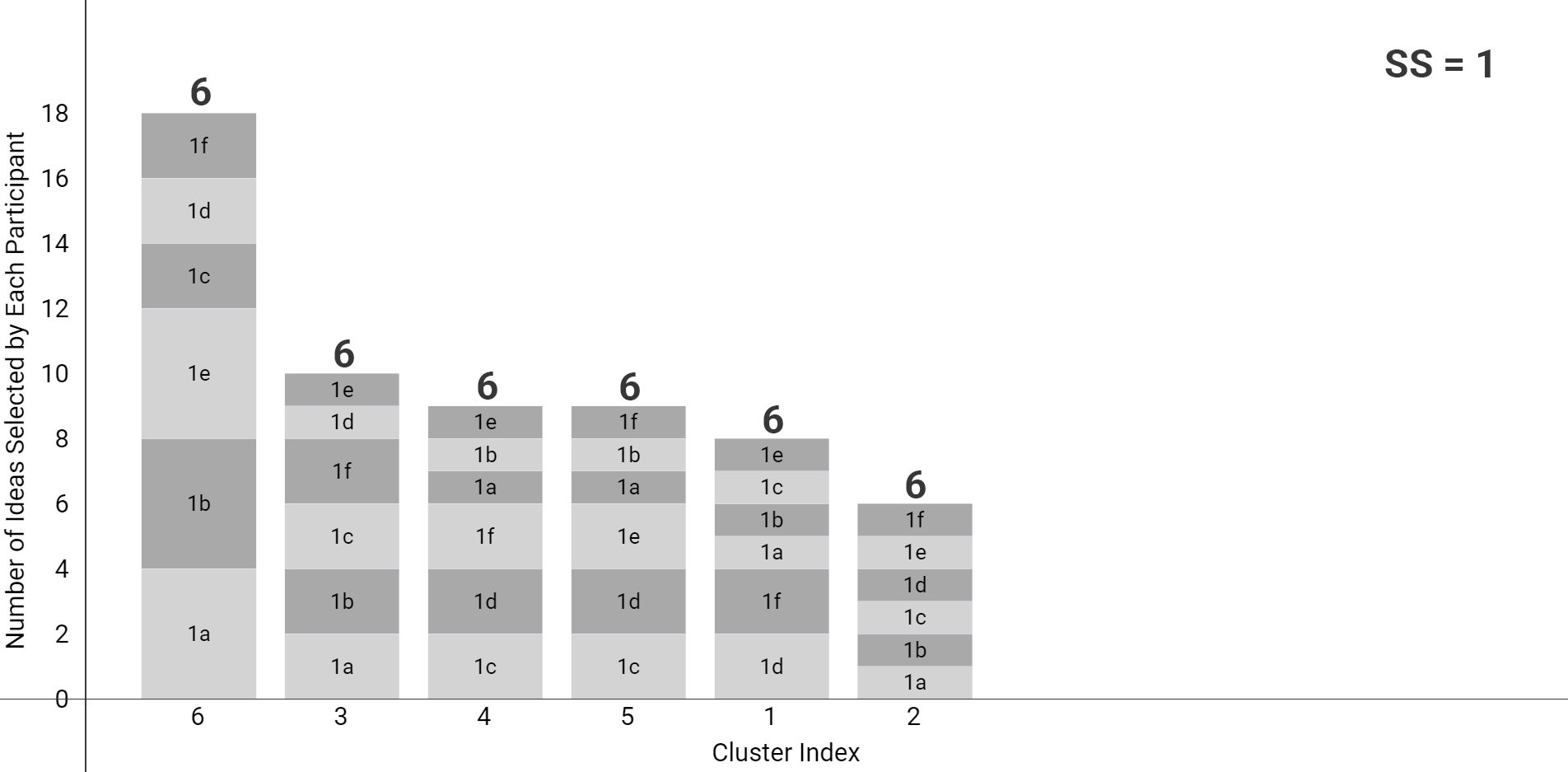}
     \caption{Idea Set 5}
     \label{fig:5_ISEL}
 \end{subfigure}
 \hfill
 \begin{subfigure}[b]{0.49\textwidth}
     \centering
     \includegraphics[width=1\textwidth]{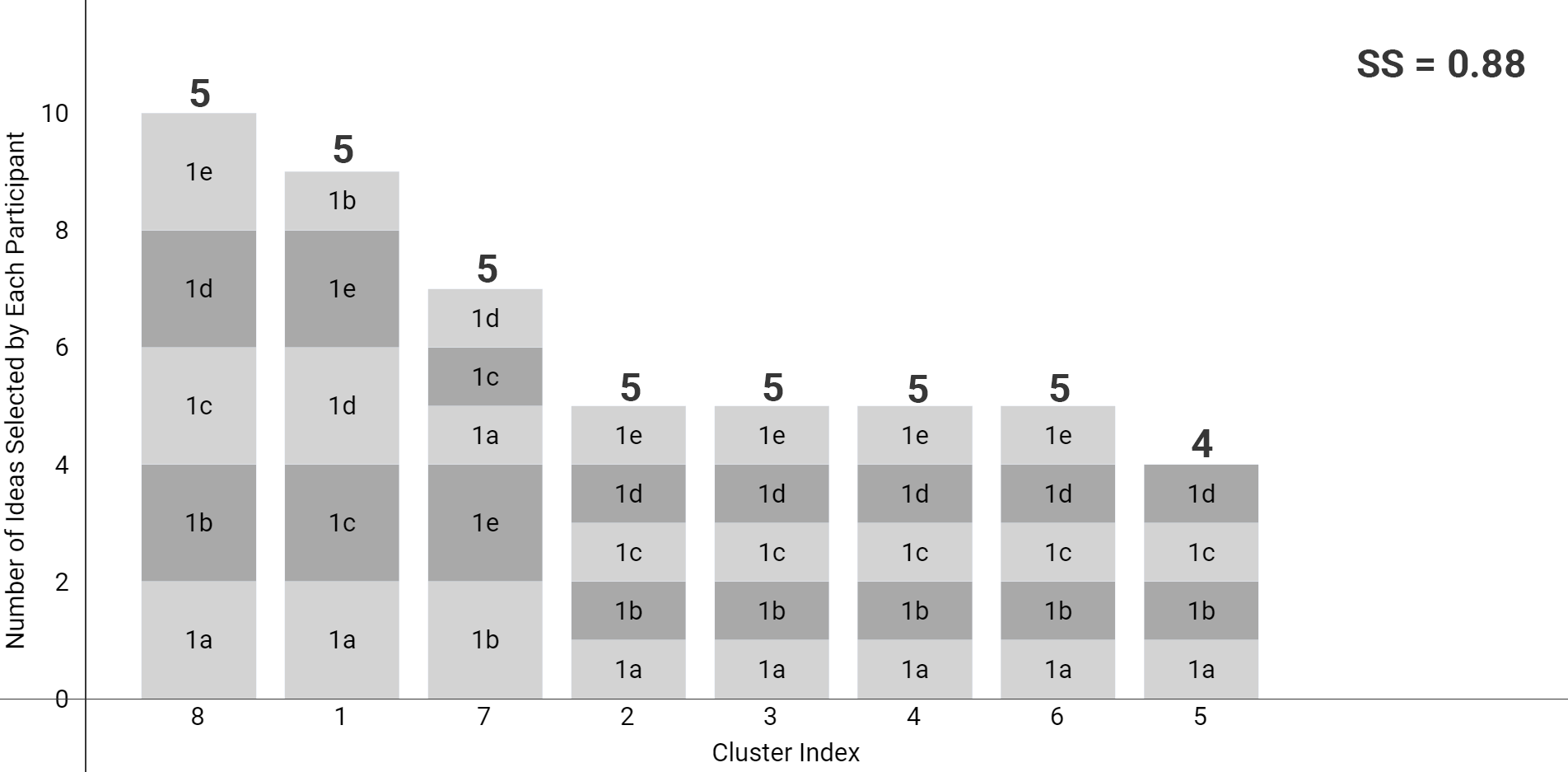}
     \caption{Idea Set 6}
     \label{fig:6_ISEL}
 \end{subfigure}

 \caption{Stacked Bar Plot of the Top 10 Selected Ideas from Each Idea Set}
 \label{fig:stacked_bar_plot}
\end{figure*}

Let us consider a plot that is used by a subset X of P participants. The number of participants selecting at least one idea from a cluster is its \emph{Selection Index (SI)}. If the number of clusters in a plot with SI=X is M where the total number of clusters is C, then we define \emph{Sampling Score (SS)} for the plot as M/C. These data are presented in Figure~\ref{fig:stacked_bar_plot}, wherein the bars are arranged in the \emph{decreasing order of the cluster areas}. The height of each bar represents the combined number of ideas selected by all participants from the respective clusters. The SI for each cluster is indicated at the top of each bar. The SS is indicated at the centre of each plot. From the plots the following observations can be made.

\begin{enumerate}
    \item The bars follow a monotonous trend. This implies that the number of ideas selected from a cluster is positively influenced by the size of the cluster. 
    \item In 3 of 6 plots, SS=1; in 2 of the remaining 3 plots, SS is significantly close to 1. This indicates the participants predominantly made use of the clusters for their sampling. This strategy ensured that the selections were diverse, as they originated from different clusters.
    \item In one of the cases (idea set 4), SS is low. It can be observed in Figure~\ref{fig:4_CTP} that this set has 3 clusters with a significantly smaller number of ideas compared to other clusters in the set. These are the clusters with SI<X. This shows that a cluster with less number of points has a lesser probability of selection. This observation holds true even in idea sets 3 and 6, which have one cluster each with a sparse population.
\end{enumerate}

An additional observation was that, although the participants were allotted 10 minutes for the task, the visual clusters enabled them to select 10 ideas within 2 minutes. This observation underscores the effectiveness of perceptual organization techniques, such as UMAP and DBSCAN, in facilitating quick and efficient idea selection while ensuring diversity. 

Thus, embedding and dimensionality reduction with clustering minimized the cognitive effort of the designers in selecting diverse and representative ideas and also provided valuable insight into the behaviour of the designers during the selection process. Thus, the answer for RQ1.2 is \emph{Yes, the idea embedding is useful to the designers}.


\subsection{Distribution of Ideas in the Idea Space (RQ2.1)}
Through the process of embedding, the idea space is geometrically a hypercube. An ideal idea exploration should cover this idea space with statistically uniform sampling. However, due to the cliche of the "Curse of Dimensionality", the assessment of uniformity in the dimension of the embedding is unreliable. Hence, we make use of the dimensionality-reduced cluster maps as shown in Figure~\ref{fig:cluster_triad_plot} for this assessment. If an idea space is uniformly explored, the ideas would be evenly distributed across the entire area. This would prohibit clustering. However, any finite sampling exercise generally exhibits clustering. DBSCAN inherently guarantees a certain level of uniformity of points within a cluster. However, this parameter for different clusters could be different because of their adaptive nature. With this understanding, we measure the distribution of points within a cluster and the distribution of clusters in an idea space as follows.

\begin{figure*}[h!]
\centerline{\includegraphics[width=0.95\textwidth]{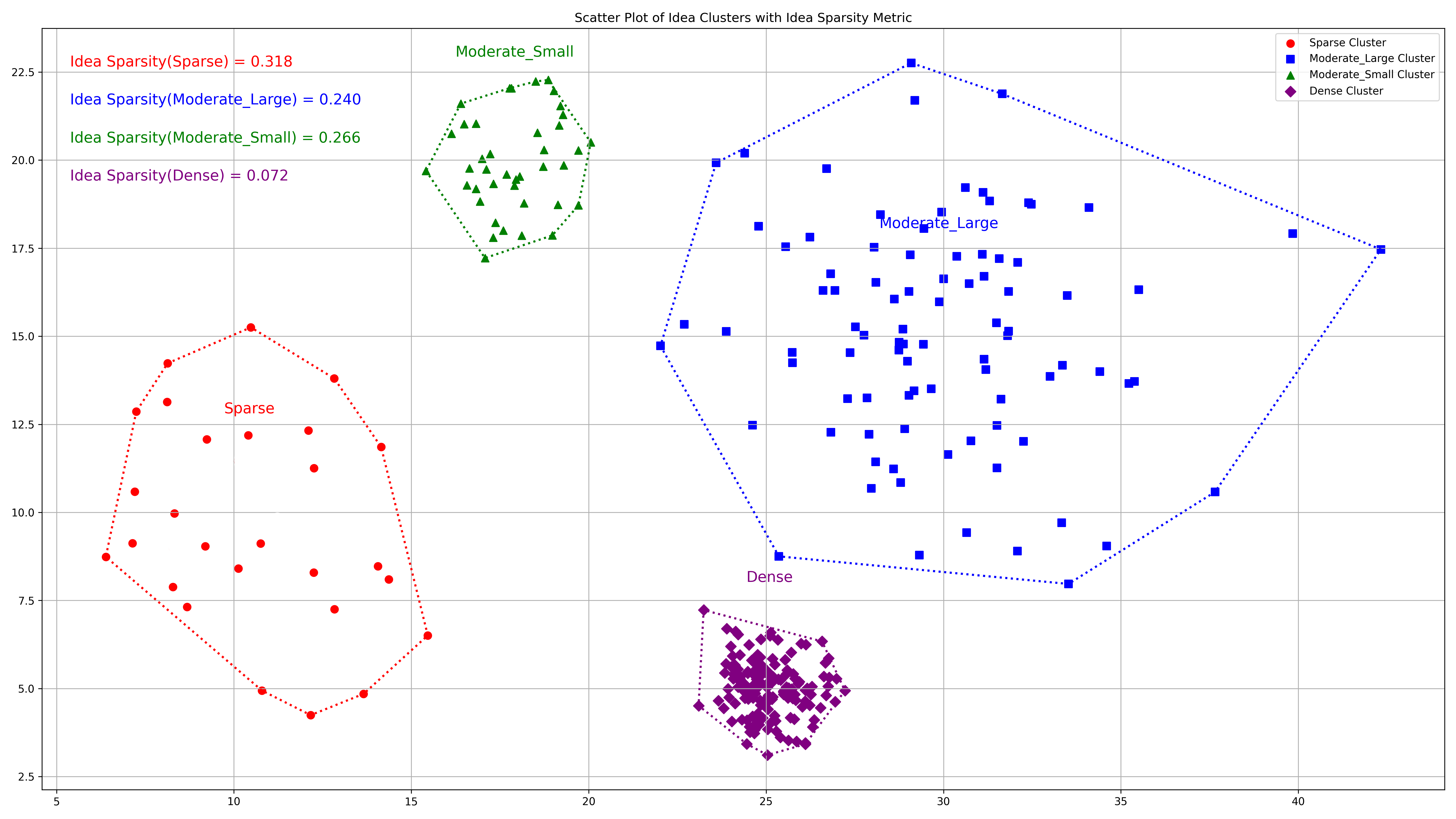}}
\caption{Representative diagram illustrating four types of idea clusters characterized by varying spatial distributions—Sparse, Moderate (Large Spread), Moderate (Small Spread), and Dense. Each cluster is visualized with distinct colors and symbols, enclosed by a dotted convex hull boundary. The corresponding Idea Sparsity (IS) values quantify the spatial spread of the ideas within each cluster.}
\label{fig:rep_diagram_idea_sparsity}
\end{figure*}

\begin{figure*}[h!]
\centering
\begin{subfigure}[b]{0.48\textwidth}
    \centering
     \includegraphics[width=1\textwidth]{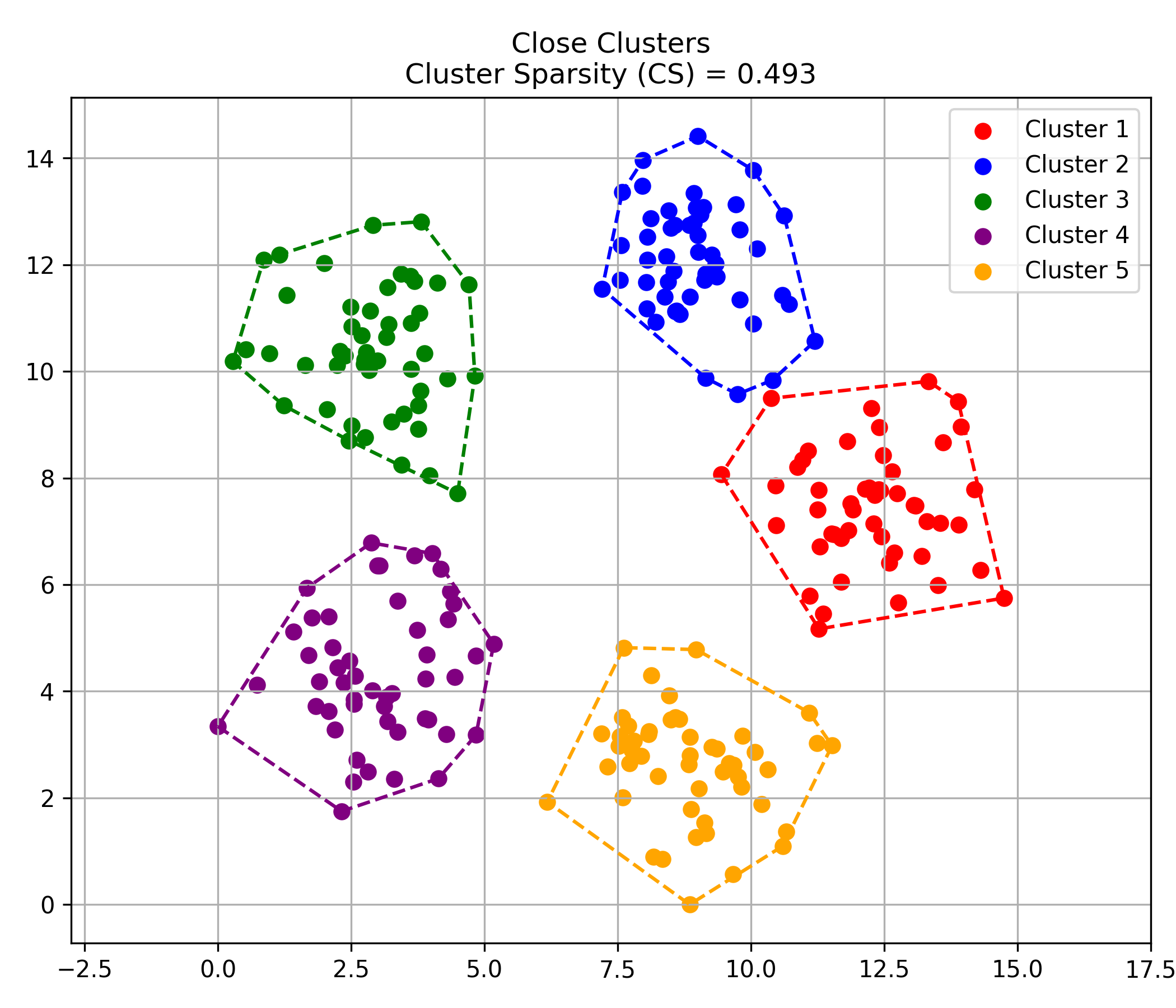}
     \caption{Tight-Knit Clustering Pattern}
     \label{fig:tight_knit_clustering}
 \end{subfigure}
 \hfill
 \begin{subfigure}[b]{0.48\textwidth}
     \centering
     \includegraphics[width=1\textwidth]{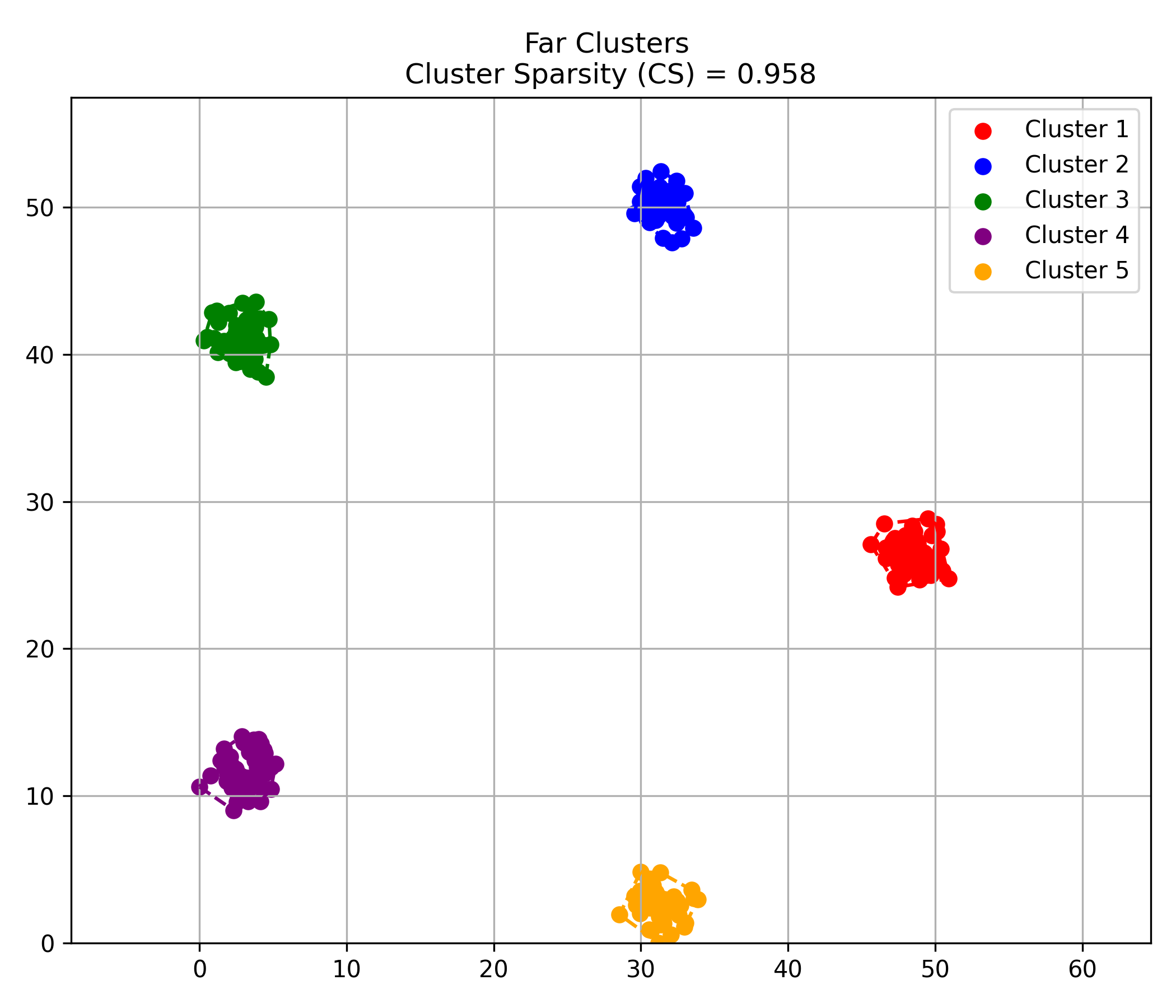}
     \caption{Loosely-Woven Clustering Pattern}
     \label{fig:loose_woven_clustering}
 \end{subfigure}
 \label{}
 \caption{Representative diagram illustrating two clustering scenarios—Tight-Knit Clustering and Loosely-Woven Clustering—with identical intra-cluster spread across five distinct, non-overlapping clusters. Dotted convex hulls outline each cluster, while the Cluster Sparsity (CS) metric captures the degree of global separation. A higher CS in the loosely-woven configuration signifies more distinct and spatially dispersed idea groups. Note: The corresponding clusters are of the same size on both plots.}
\end{figure*}

\subsubsection{Idea Sparsity: Quantifying Local Distribution of Ideas within a Cluster}
An effective ideation exercise would have a set of ideas that cover a large region in the idea space. This means that the mean distance between two ideas in that space is large. We refer to this as a \emph{Sparse} distribution. When the ideas are all diverse, no two ideas would belong to a cluster because each cluster represents a set of similar ideas. However, in the presence of clustering, the mean distance between the ideas in different clusters is usually different. Also, the larger the number of clusters, the greater the number of distinct ideas. The distinctness of ideas within each such cluster can be categorized based on their distribution into \emph{Sparse} (Small No. of ideas over Larger Area), \emph{Moderate} (Large/Small No. of ideas over Larger/Smaller Area respectively) and \emph{Dense} (Large No. of ideas over Smaller Area) (as shown in Figure~\ref{fig:rep_diagram_idea_sparsity}). A sparse cluster is considered locally more effective than moderate, and similarly, a moderate cluster is more effective than dense. We define \emph{Idea Sparsity (IS)} as a measure of how sparsely the ideas are distributed within a cluster in the idea space. It captures the density-independent spatial spread of ideas, which reflects the degree of local diversity among semantically similar ideas.

The equation used to calculate the Idea Sparsity (IS) for a given cluster \textit{i} is as follows:

\begin{displaymath}
Idea \: Sparsity = \frac{A_c}{N_i} \: . (\exp({-\frac{A_c}{N_i}}))
\end{displaymath}

where,\newline
$N_i$ is the Number of Ideas in a cluster\newline
$A_c$ is the Area of the cluster computed as the area of the convex hull of the points in that cluster. 

If the idea sparsity is similar across all the clusters, we say that the idea exploration is uniform. In that case, in a spider plot (Figure~\ref{fig:spider_plot_idea_density}), where each spoke corresponds to a cluster and the length of the spoke corresponds to idea sparsity, a uniform exploration would appear as a regular convex polygon. The larger the polygon, the more number of ideas are distributed uniformly over a larger area within each cluster.


\subsubsection{Cluster Sparsity (CS): Quantifying Global Separation between Clusters in the Idea Space}
Each cluster in an idea space is distributed over an area. Ideally, we desire to have clusters that are sparsely distributed from each other (Loosely-woven clustering - Figure~\ref{fig:loose_woven_clustering}) since they represent more distinct sets of ideas compared to clusters that are close to each other (Tight-knit clustering - Figure~\ref{fig:tight_knit_clustering}), as evidenced by the expert opinion in the section \ref{subsec:meaningfulness}. In order to quantify the nature of the distribution of clusters in the idea space, we define \emph{Cluster Sparsity (CS)} as a measure of how well-separated the clusters are in the overall idea space. It reflects the global distinctness between groups of ideas, each representing different semantic themes.

The Cluster Sparsity (CS) is calculated using the equation as follows:

\begin{displaymath}
Cluster \: Sparsity = 1-\frac{\sum_{i=1}^{N_c} A_i}{A_t}
\end{displaymath}

where, \newline
$N_c$ is the number of clusters in an idea space, \newline
$A_i$ is the area of individual clusters, \newline
$A_t$ is the total area of the idea space computed as the area of the convex hull of all the points in that idea space. 

\begin{figure*}[h!]
\centering
 \begin{subfigure}[b]{0.33\textwidth}
    \centering
     \includegraphics[width=1\textwidth]{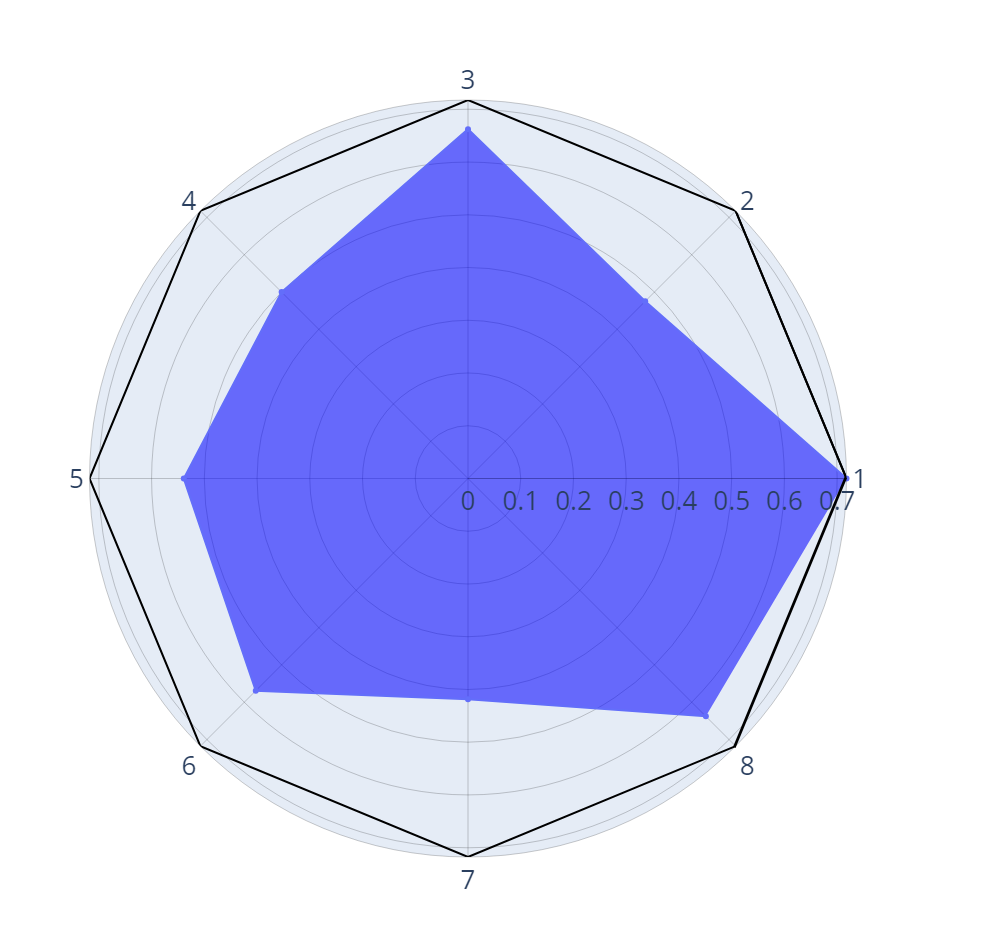}
     \caption{Idea Set 1}
     \label{fig:1_IDSP}
 \end{subfigure}
 \hfill
 \begin{subfigure}[b]{0.33\textwidth}
     \centering
     \includegraphics[width=1\textwidth]{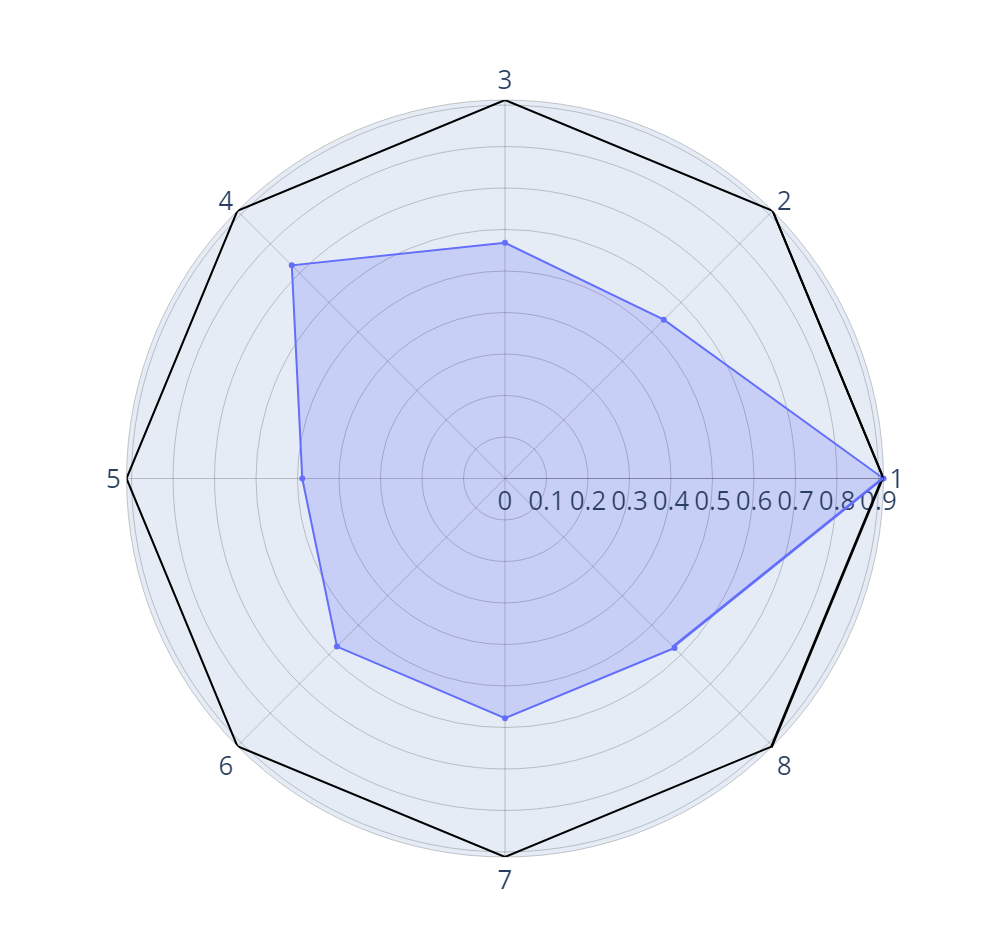}
     \caption{Idea Set 2}
     \label{fig:2_IDSP}
 \end{subfigure}
 \hfill
 \begin{subfigure}[b]{0.33\textwidth}
     \centering
     \includegraphics[width=1\textwidth]{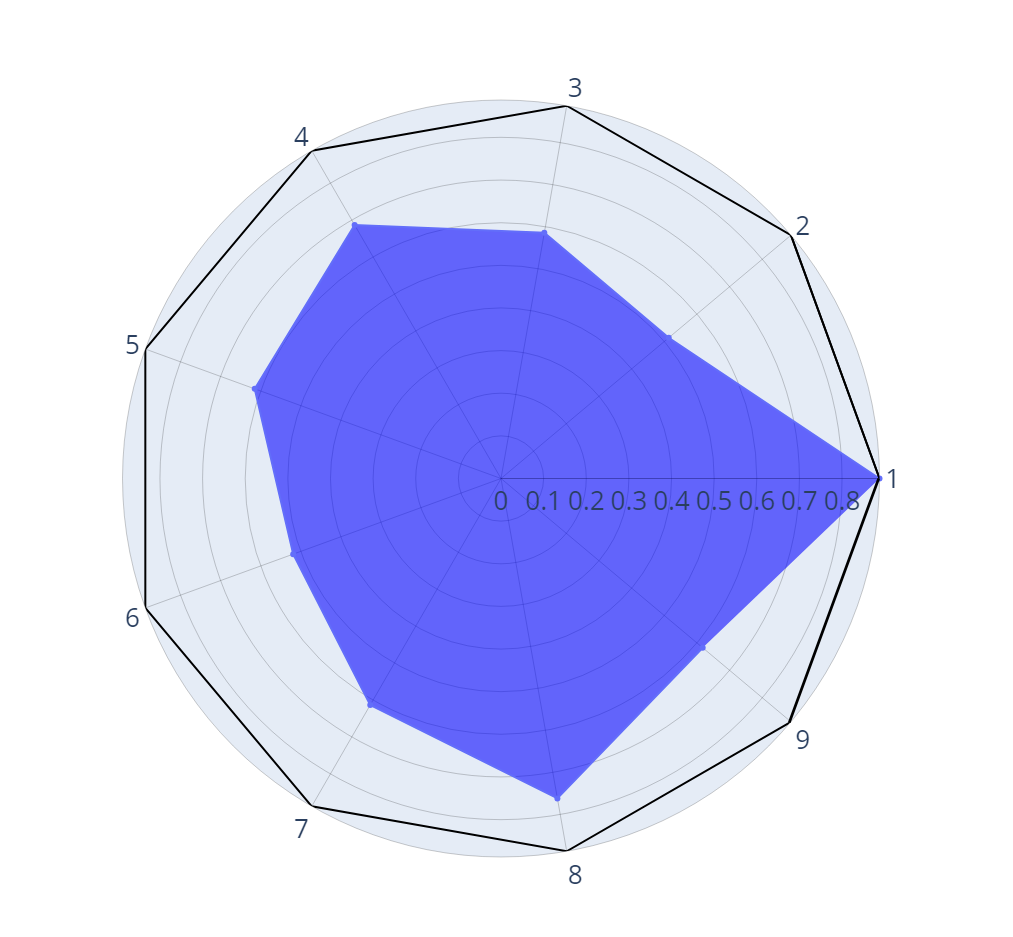}
     \caption{Idea Set 3}
     \label{fig:3_IDSP}
 \end{subfigure}

 \vfill
 
 \begin{subfigure}[b]{0.33\textwidth}
     \centering
     \includegraphics[width=1\textwidth]{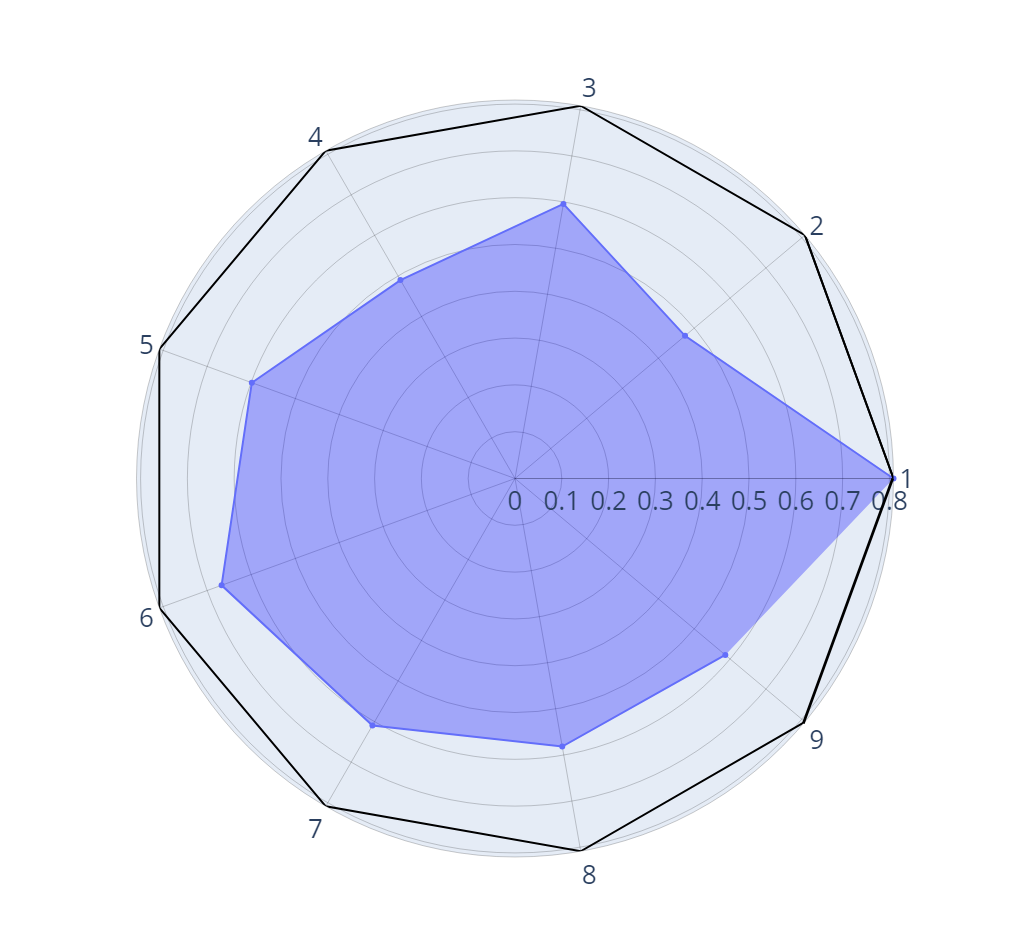}
     \caption{Idea Set 4}
     \label{fig:4_IDSP}
 \end{subfigure}
 \hfill
 \begin{subfigure}[b]{0.33\textwidth}
     \centering
     \includegraphics[width=1\textwidth]{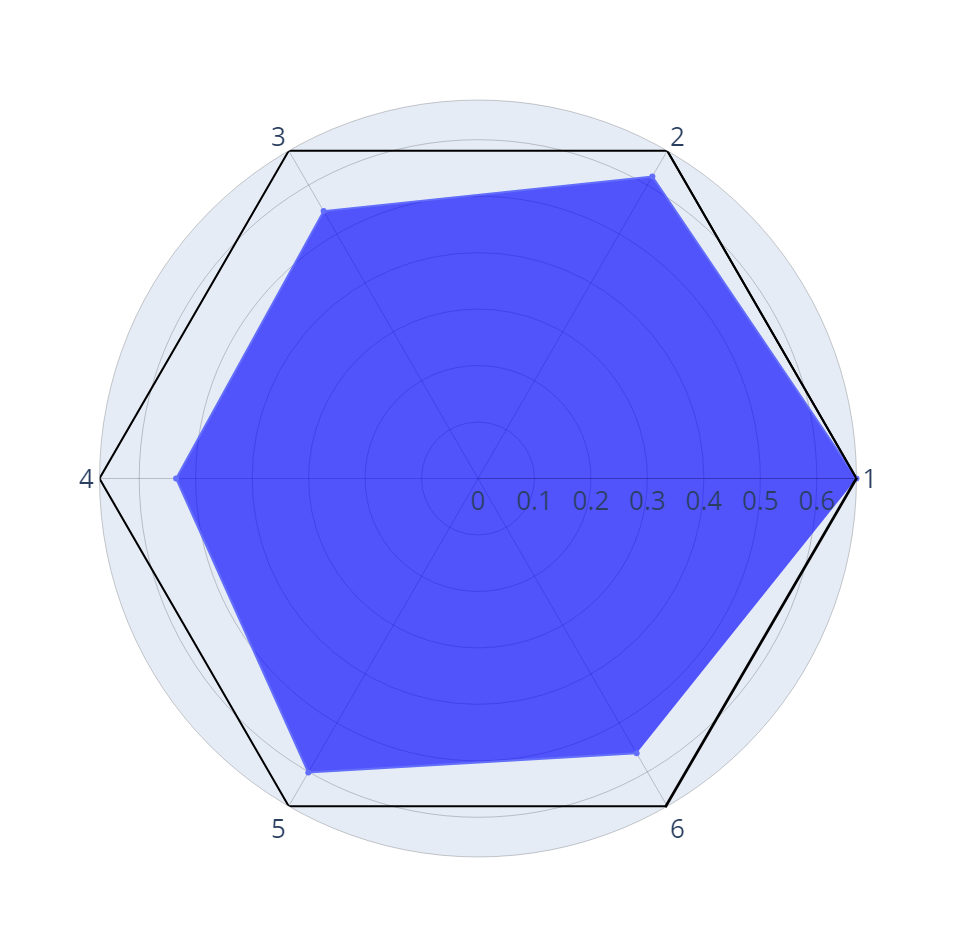}
     \caption{Idea Set 5}
     \label{fig:5_IDSP}
 \end{subfigure}
 \hfill
 \begin{subfigure}[b]{0.33\textwidth}
     \centering
     \includegraphics[width=1\textwidth]{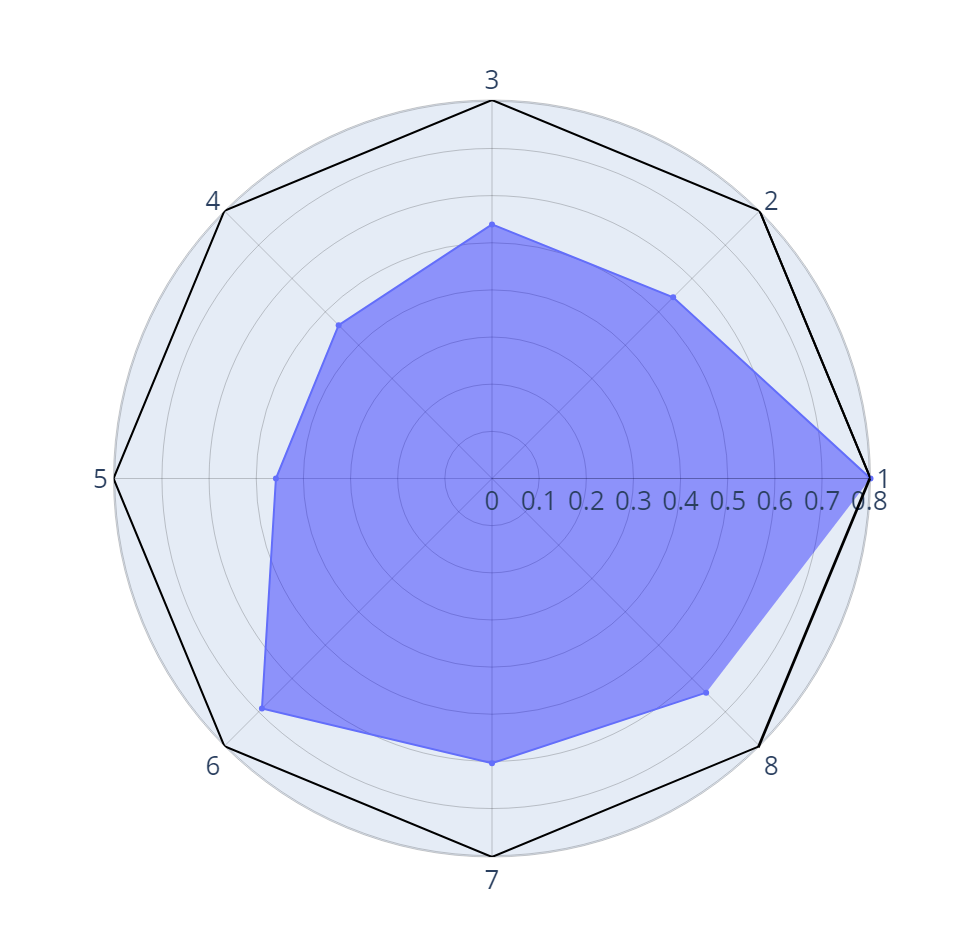}
     \caption{Idea Set 6}
     \label{fig:6_IDSP}
 \end{subfigure}

 \caption{Spider Plot of Idea Density and Cluster Density for Each Idea Set}
 \label{fig:spider_plot_idea_density}
\end{figure*}

An exploration where the clusters are well separated would have high cluster sparsity (desirable), and a compact set of clusters would give low cluster sparsity (undesirable). In Figure~\ref{fig:spider_plot_idea_density}, the brightness of the bounded polygon represents the cluster sparsity.

DBSCAN clusters potentially produce a default cluster with points that could not be included in any of the well-defined clusters; this is usually referred to as the noise cluster. This is the one that is not indicated with a bounding curve. The points in this cluster are important as the associated ideas are distinct from clusters of similar ideas. We compute Idea Sparsity for all clusters, including the noise cluster. However, we observe that (a) noise clusters do not significantly extend the region of explored ideas, and (b) the convex bounding curve for the noise cluster inevitably overlaps multiple identified clusters. Hence, for the computation of cluster sparsity, we discount the noise cluster.

\subsubsection{Results and Inference}

In terms of cluster and idea sparsities, \emph{an effective idea exploration would have a high cluster sparsity with a regular polygon of large size for idea sparsity}.

The bounded polygon for each idea set is compared against the ideal regular polygon. The degree of deviation is calculated as the ratio of the area of the spider plot to the area of the largest regular polygon. This metric is referred to here as \emph{Distribution Score (DS)}. The polygon provides a visual measure, and the distribution score provides a quantitative measure of how well the ideas were distributed across the clusters. A close match to the regular polygon would indicate a uniform distribution, while significant deviations would suggest regions where the idea generation process may have been uneven. The distribution score for each idea set is given in Table~\ref{tab:distribution_score}. It is seen that 4 out of 6 sets have a score close to 0.5, while idea set 1 has a score of 0.61 and idea set 5 has a score of 0.77, indicating that the ideas are potentially distributed uniformly in each cluster. In terms of cluster sparsity, it can be seen that idea sets 1, 3 and 5 are brighter, showing that the clusters are well-separated compared to idea sets 2, 4 and 6, where the clusters are close to each other. Thus, overall, Idea Set 5 represents a uniform exploration compared to other idea sets in terms of idea and cluster sparsity. Thus, exploration in idea set 5 is considered the most effective, and idea set 2 is the least effective per the above metrics.
Thus, the answer for RQ2.1 is, \emph{Yes, the cluster-analysis of embedded ideas provides a reasonable framework for objective assessment of the effectiveness of idea exploration.}

\begin{table} [h!]
    \centering
    \begin{tabular}{p{0.5cm}|p{1.5cm}|p{1.5cm}|p{2.5cm}}
        Idea Set No. & Area of the Regular Convex Polygon (ARP) & Area of the Irregular Polygon (AIP)  & Distribution Score (DS) = AIP/ARP \\
        \hline
        1 & 1.45 & 0.89 & 0.61 \\
        \hline
        2 & 2.35 & 1.07 & 0.46 \\
        \hline
        3 & 2.28 & 1.19 & 0.52 \\
        \hline
        4 & 1.89 & 1.03 & 0.55 \\
        \hline
        5 & 1.16 & 0.90 & 0.77 \\
        \hline
        6 & 1.82 & 0.99 & 0.55 \\
    \end{tabular}
    \caption{Distribution Score}
    \label{tab:distribution_score}
\end{table}





\subsection{Dispersion of Ideas in the Idea Space (RQ2.2)}
Here, we assess whether ideas spanned all dimensions of the idea space effectively. Using the measures described above, comprehensive coverage of all the dimensions of the embedding cannot be assessed as it is based on a 2D map of the original embeddings. To assess the dispersion of ideas, which refers to the extent to which the ideas are spanned in the higher-dimensional space, Principal Component Analysis (PCA) was adopted. PCA provides eigenvectors that describe the direction and relative magnitude of dominant explorations. An effective exploration of all dimensions of idea space would give \emph{comparable} eigenvalues of reasonably \emph{large} magnitude. For the assessment of the comprehensiveness of different ideation exercises, the eigenvalues were computed using PCA for each idea set and sorted in descending order.  

\subsubsection{Results and Inferences}
\begin{figure}[h!tbp]
\centering
 \begin{subfigure}[b]{1\textwidth}
    \centering
     \includegraphics[width=1.0\textwidth]{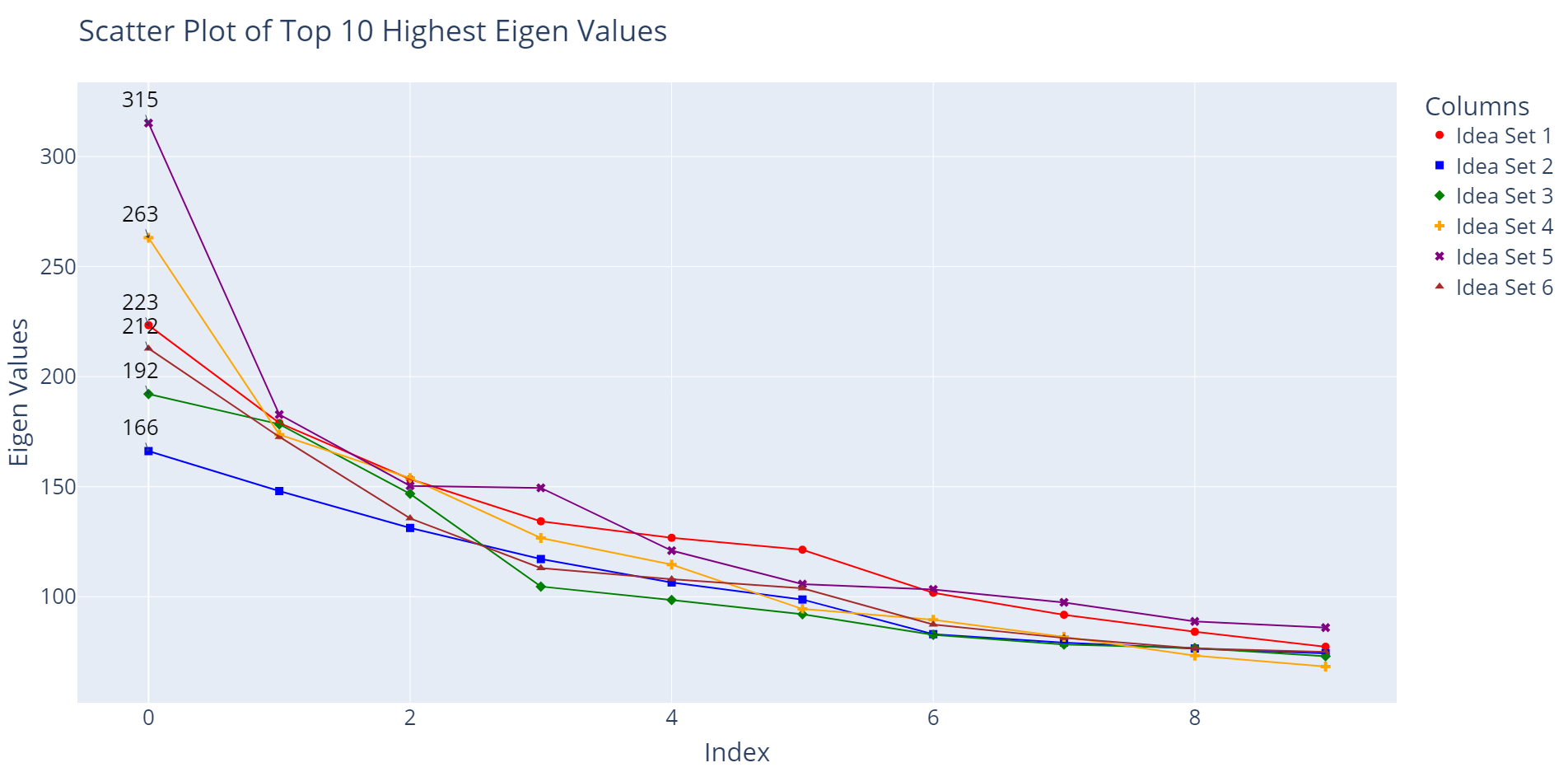}
     \caption{Line Plot of Top 10 Highest Eigen Values of PCA Embedding for Ideas Generated for Six Problem Statements}
     \label{fig:eigen_10}
 \end{subfigure}
 \vfill
 \begin{subfigure}[b]{1\textwidth}
     \centering
     \includegraphics[width=1.0\textwidth]{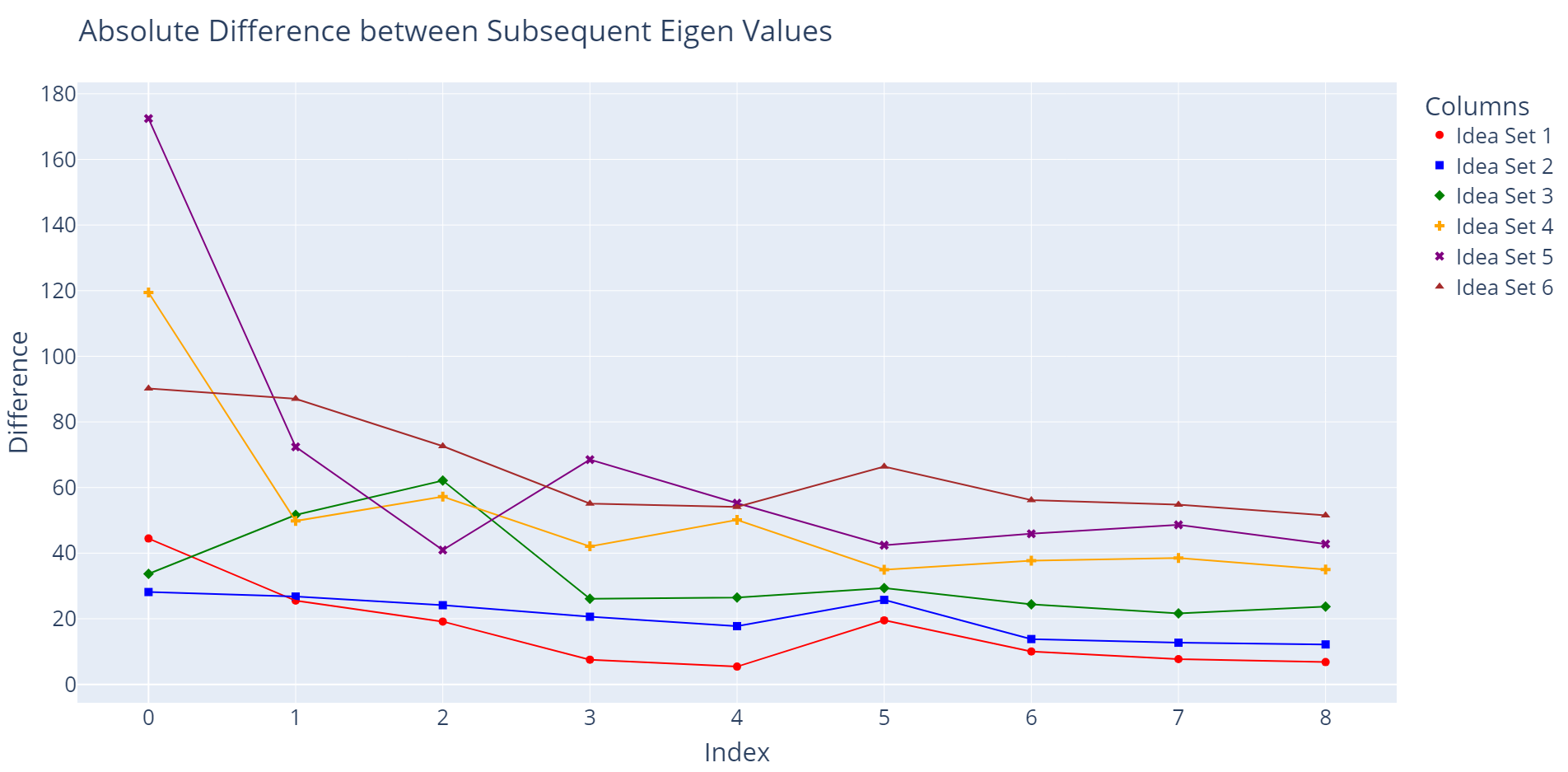}
     \caption{Line Plot of Absolute Different Between the Top 10 Highest Eigen Values}
     \label{fig:eigen_10_diff}
 \end{subfigure}

 \caption{Line Plot of Eigen Values and their Difference of PCA of Embedding for Ideas Generated for Six Problem Statements}
 \label{fig:eigen_value_plot}

\end{figure}

The results are presented in Figure~\ref{fig:eigen_10} and Figure~\ref{fig:eigen_10_diff}. It was observed that for all the dimensions, the idea space is explored more or less uniformly except for the first few dominant eigen directions. The maximum eigenvalue observed was 300, although all eigenvalues eventually saturated in the range of 30 to 50. Hence, the comparison of exploration below uses the first 10 eigenvalues only. 

Figure~\ref{fig:eigen_10} presents the top 10 eigenvalues in descending order, which indicates the spread of ideas in different directions in the high dimensional space of idea embedding. Figure~\ref{fig:eigen_10_diff} shows the differences between subsequent eigenvalues. By examining this difference, the evenness of the spread is assessed. Ideally, for a highly dispersed idea space, the differences between subsequent eigenvalues should be smaller showing a flat line at an elevated position, indicating that the ideas are spread out across multiple dimensions. It is evident from Figure~\ref{fig:eigen_10_diff} that idea sets 6, 2 and 1 approximates a flat line, whereas idea sets 4 and 3 show a short jump between the first three eigenvalues indicating a longer spread in three dimensions and finally the idea set 5 show a significant jump between the first and second eigenvalue indicating a more directed exploration. 
Thus, the answer for RQ2.2 is, \emph{Yes, the analysis of the dispersion of embedded ideas in high dimension provides a reasonable framework for the objective assessment of comprehensiveness of idea exploration.}




\hfill \hrule

\section{Evaluation through existing AI-Based Methods}
Automated evaluation techniques leveraging Artificial Intelligence (AI) have recently emerged as prominent tools for assessing ideas. To explore their applicability in supporting novice designers during ideation, we employed five AI-based methods—namely, GPT-based evaluation, OCSAI (Open Creativity Scoring with Artificial Intelligence), AIDE (AI for Design Evaluation), CLAUS (Cross-Lingual Alternate Uses Scoring), and SemDis (Semantic Distance). These methods predominantly evaluate individual ideas based on a predefined scale (1 to 5), assessing factors such as originality or novelty relative to previously existing ideas. To assess their suitability, the same set of 100 previously generated ideas for each of six distinct product design problems (PS1 through PS6) was independently applied to each of five methods.

\subsection{Results and Observations}
The results of these evaluations were represented through two primary visualization techniques—box and whisker plots (depicting mean, median, standard deviation, Q1, and Q3) and bar plots with individual ratings, mean scores, and standard deviation markers. The plots are given in the appendix. Across the six problem statements, mean scores varied marginally within a relatively narrow range for each method, indicating a moderate degree of consistency at the aggregated statistical level. For instance, the AIDE method yielded a mean novelty score of 3.83 with standard deviations (SD) of 0.31, whereas CLAUS ratings ranged slightly lower (mean score of 3.43, SD of 0.2). Similarly, GPT and SemDis exhibited comparable average novelty scores, maintaining mean values of 3.32 and 3.09. However, OCSAI scores were notably lower, with a mean rating of 2.28 and higher variability (SD of 0.7). Despite similar statistical summaries (mean and SD), significant differences in score distributions and individual assessments were apparent. GPT-based assessments, for instance, displayed wider variability with frequent extreme values, whereas CLAUS and AIDE exhibited narrow distribution clusters. SemDis and OCSAI distributions displayed varied spread with moderate consistency.

\begin{figure*}[t!hbp]
    \centering
    \includegraphics[width=0.95\linewidth]{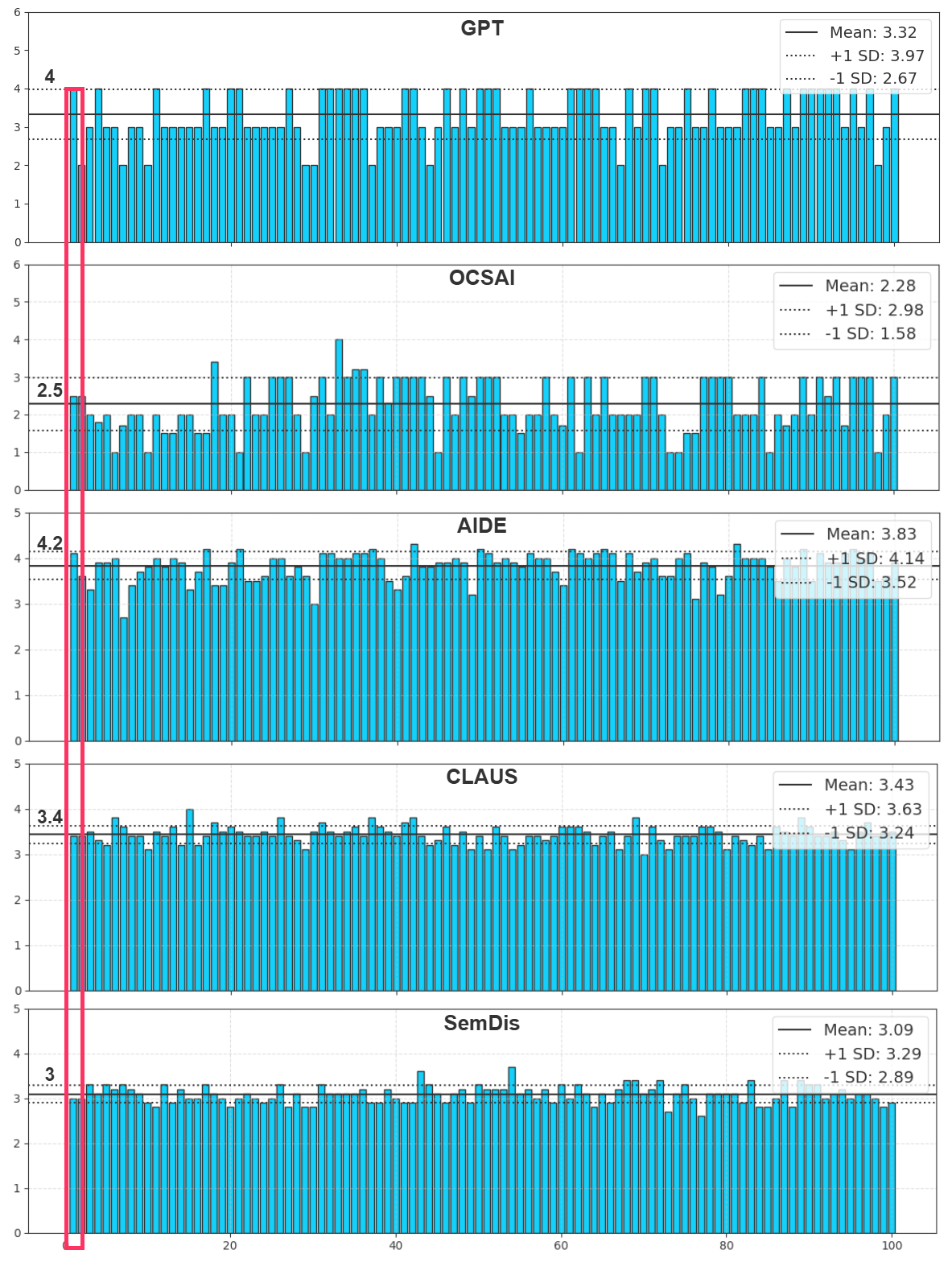}
    \caption{Bar plot comparing the novelty ratings of ideas for Problem Statement 1 as assessed by five different AI-based methods}
    \label{fig:bar_plot_ps1_ai_idea_eval}
\end{figure*}

\subsection{Limitations of the existing AI-based assessment Methods}
Despite the systematic approach and the quantitative nature of these methods, these AI-based methods primarily provide absolute ratings of individual idea novelty, but do not offer mechanisms for comparative or relational evaluation among ideas, nor provide any insights into the nature of the ideation process that led to the generation of these ideas. Importantly, such approaches provide limited insights into the relative diversity or distinctiveness of ideas within a given set. For instance, two ideas scored equally high for novelty by AIDE or GPT might be conceptually very similar, providing little added value to designers looking for distinctly different ideas. Moreover, the assessment outcomes were inconsistent across different methods. Ideas receiving high ratings from GPT-based assessment frequently differed from those highly rated by OCSAI, CLAUS, or SemDis across all ideas. This inconsistency reflects methodological variations in criteria, model training, and sensitivity to input nuances, complicating the designers' task of meaningfully interpreting these results to confidently shortlist promising ideas. Furthermore, none of these methods produced ratings below a certain threshold (e.g., below 2), reducing their discriminative power and limiting their utility for designers seeking genuinely innovative and diverse idea pools. These methods, while effectively assigning novelty ratings of ideas, fail to adequately characterize the ideation process and differentiate ideas based on conceptual diversity.

\subsection{Advantages of our Proposed Embedding-based Clustering Method}
Recognizing the shortcomings inherent in existing AI-based evaluation techniques, our proposed embedding-based clustering method addresses these critical gaps. Rather than assigning individual, isolated scores, our method leverages high-dimensional embeddings to represent ideas spatially. Through clustering techniques, it explicitly focuses on identifying distinctive groups of ideas, thus highlighting conceptual diversity and uniqueness within the idea pool.

\subsubsection{Comparative Evaluation for a Large Set of Ideas}
In contrast to existing AI-based methods that evaluate ideas in isolation, the embedding-based approach evaluates ideas collectively in relation to each other. Ideas are represented as points in a high-dimensional vector space, and semantic similarity or distance among ideas is computed systematically. This representation enables designers to visualize the relative positioning of ideas, making it significantly easier to recognize conceptual redundancy and diversity within the idea set. As a result, designers can make informed and strategic choices, selecting ideas not only for their individual novelty but also for their comparative distinctiveness and collective representativeness.

\subsubsection{Enhanced Support for Novice Designers}
One main benefit of our embedding-based approach is its suitability in assisting novice designers who typically lack expertise in effectively managing and analyzing large ideation sets. Unlike existing methods, which leave designers to interpret and select ideas based solely on individual novelty scores, embedding-based clustering provides an intuitive, visual, and structured representation of ideas. By mapping semantic relationships clearly, it equips novice designers with the capacity to quickly identify distinct clusters of ideas, thereby significantly reducing cognitive load and decision-making complexity associated with selection tasks. Moreover, clustering inherently aids in the identification of "hidden gems", the ideas that might otherwise be overlooked due to less compelling isolated novelty ratings but that distinctly stand out within specific clusters, enriching diversity within the selected shortlist. Such strategic selection from diverse clusters ensures comprehensive exploration of the ideation space, enhancing overall design innovation potential.

\subsubsection{Transparent, Scalable, and Reproducible Outcomes}
Another advantage of our method lies in its transparency and reproducibility. AI-based scoring methods often lack clarity in their reasoning, making it challenging to justify scores or replicate evaluations consistently across contexts. In contrast, embedding-based clustering provides transparent reasoning grounded in semantic relationships and geometric structures, which remain stable and reproducible across multiple assessments. The resultant clusters can be quantitatively and qualitatively justified, lending greater validity and reliability to the assessment process. Furthermore, embedding-based methods scale efficiently to handle large volumes of ideas without sacrificing accuracy or interpretability, thus fitting well into realistic industrial design scenarios characterized by extensive ideation outputs.

In summary, while existing methods effectively quantify individual idea novelty, their isolated and non-comparative evaluations severely limit their practical utility for novice designers tasked with shortlisting promising and diverse ideas from large pools. Our embedding-based clustering approach fundamentally addresses these issues by providing comparative, interpretable, and visually intuitive evaluations of idea sets. The method empowers novice designers by highlighting conceptual uniqueness and relational diversity, thereby supporting more informed, strategic, and efficient idea selection decisions. Consequently, our approach significantly outperforms existing AI-based methods in providing practical, scalable, and meaningful assistance during the crucial selection phase of the design process.

\hfill \hrule

\section*{Discussion}

\begin{figure}[htbp]
    \centering
    \includegraphics[width=1\linewidth]{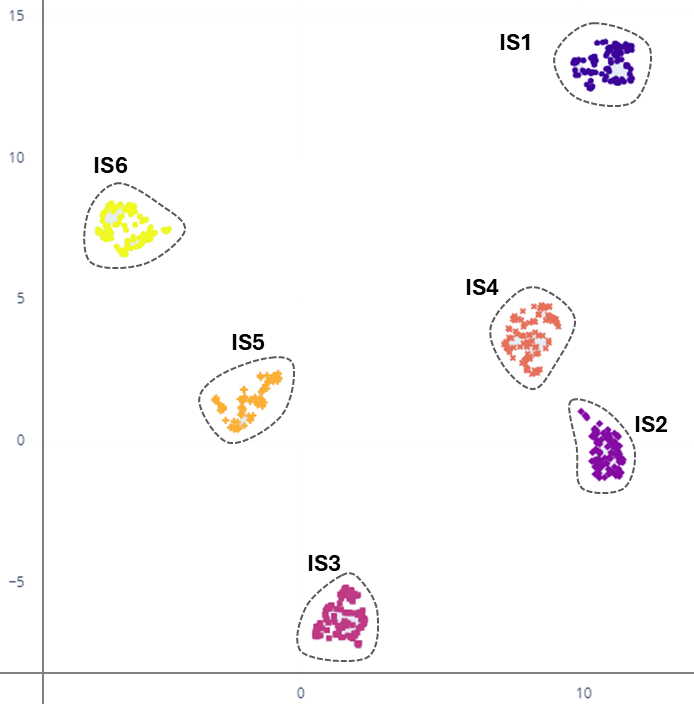}
    \caption{DBSCAN Clustering of UMAP Embeddings of Ideas Generated for All Six Problem Statements}
    \label{fig:dbscan_all_ideas}
    \footnote{IS - Idea Set}
\end{figure}

An intriguing observation emerged when all idea sets were processed through UMAP and DBSCAN simultaneously. The resulting plot, as shown in Figure~\ref{fig:dbscan_all_ideas}, revealed that each idea set formed a dense cluster island, representing a closely-knit group of 100 ideas.  Even though in the context of the present study, the problem statements were not included as a part of the idea embeddings, certain idea sets were positioned closer to each other, suggesting potential similarities in addressing the challenges posed by their respective problems. Therefore, although individual idea sets exhibited diversity and clustering, each set, in turn, appeared as clusters in a larger context; the ideas in the respective sets have something (either object or action) in common, leading to closer clustering. 

Similar plots can be used to elucidate how expert and novice designers explore ideas for various problems or how the ideation pattern changes over the career of a designer.
Merging of problem-specific idea clusters could imply the presence of some \emph{versatile ideas} that address multiple problems. We believe that such versatile ideas are more likely in the ideation by experts. 

The analysis presented above used CAI-generated ideas, as given in ~(\cite{Sankar2024}), using a particular prompt structure and temperature settings. The methodology can be used to explore other prompt structures and temperature settings to derive optimally engineered prompts for design problems. This, in turn, will provide insight into the effect of a problem statement on the efficacy of ideation. Without an objective framework for generation and evaluation, as presented in this paper, the reliability of such comparative studies would always remain in question. The authors' future research will address some of the above issues.


\hfill \hrule

\section*{Conclusion}
This paper presented a novel technique for the objective evaluation and selection of ideas by leveraging advanced techniques such as Vector embedding, UMAP, DBSCAN, and PCA. The findings indicated that the use of embeddings and clustering techniques ensured that the semantic and contextual relationships between ideas were preserved, facilitating a more meaningful and efficient evaluation process. The clustered and visually organized representations of ideas enabled quick and effective decision-making, further streamlining the ideation phase. This objective method allows designers to focus on creativity and innovation rather than the tedious task of manually sifting through large volumes of ideas. This framework not only democratizes the evaluation process but also paves the way for more consistent and reproducible outcomes which otherwise would demand the expertise and experience of the designer. Finally, our research work highlights the use of CAI towards a unified process for the generation, evaluation, and selection of ideas while maintaining the necessary rigour and reduced intervention from designers.









\newpage
\printbibliography

@misc{gonzalez-a,
  citation-number = {1},
  author = {Gonzalez, G.E. and Moran, D.A.S. and Houde, S. and He, J. and Ross, S.I. and Muller, M. and Kunde, S. and Weisz, J.D.},
  title = {Collaborative Canvas: A Tool for Exploring LLM Use in Group Ideation Tasks},
  language = {en}
}

@article{Dean2006,
  title = {Identifying Quality,  Novel,  and Creative Ideas: Constructs and Scales for Idea Evaluation},
  volume = {7},
  ISSN = {1536-9323},
  DOI = {10.17705/1jais.00106},
  number = {10},
  journal = {Journal of the Association for Information Systems},
  publisher = {Association for Information Systems},
  author = {Dean,  Douglas and Hender,  Jillian and Rodgers,  Thomas and Santanen,  Eric},
  year = {2006},
  month = oct,
  pages = {646–699}
}

@article{Sankar2024, title={A novel idea generation tool using a structured conversational AI (CAI) system}, volume={39}, DOI={10.1017/S089006042500006X}, journal={Artificial Intelligence for Engineering Design, Analysis and Manufacturing}, author={Sankar, B. and Sen, Dibakar}, year={2025}, pages={e11}}

@inbook{Sankar2023,
  title = {A Novel Version Control Scheme for Supporting Interrupted Product Concept Sketching},
  ISBN = {9789819904280},
  ISSN = {2190-3026},
  DOI = {10.1007/978-981-99-0428-0_33},
  booktitle = {Design in the Era of Industry 4.0,  Volume 3},
  publisher = {Springer Nature Singapore},
  author = {B., Sankar and Sen,  Dibakar},
  year = {2023},
  pages = {395–411}
}

@misc{shaer2024aiaugmented,
      title={AI-Augmented Brainwriting: Investigating the use of LLMs in group ideation}, 
      author={Orit, Shaer and Angelora, Cooper and Osnat, Mokryn and Andrew, L. Kun and Hagit, Ben Shoshan},
      year={2024},
      eprint={2402.14978},
      archivePrefix={arXiv},
      primaryClass={cs.HC}
}

@inproceedings{Desmond2024,
  series = {IUI ’24},
  title = {EvaluLLM: LLM assisted evaluation of generative outputs},
  DOI = {10.1145/3640544.3645216},
  booktitle = {Companion Proceedings of the 29th International Conference on Intelligent User Interfaces},
  publisher = {ACM},
  author = {Desmond,  Michael and Ashktorab,  Zahra},
  year = {2024},
  month = mar,
  collection = {IUI ’24}
}

@misc{liu2023geval,
      title={G-Eval: NLG Evaluation using GPT-4 with Better Human Alignment}, 
      author={Yang, Liu and Dan ,Iter and Yichong, Xu and Shuohang, Wang and Ruochen, Xu and Chenguang, Zhu},
      year={2023},
      eprint={2303.16634},
      archivePrefix={arXiv},
      primaryClass={cs.CL}
}

@inproceedings{yuhan-etal-2023-unleashing,
    title = "Unleashing the Power of Large Models: Exploring Human-Machine Conversations",
    author = "Yuhan, Liu  and
      Xiuying, Chen  and
      Rui, Yan",
    editor = "Zhang, Jiajun",
    booktitle = "Proceedings of the 22nd Chinese National Conference on Computational Linguistics (Volume 2: Frontier Forum)",
    month = aug,
    year = "2023",
    address = "Harbin, China",
    publisher = "Chinese Information Processing Society of China",
    pages = "16--29",
    language = "English",
}

@article{meyer2023a,
  author = {Meyer, J.G. and Urbanowicz, R.J. and Martin, P.C.N.},
  title = {ChatGPT and large language models in academia: opportunities and challenges},
  volume = {16},
  pages = {20},
  date = {2023},
  doi = {10.1186/s13040-023-00339-9},
  language = {en},
  journal = {BioData Mining}
}

@article{Ben_2010, title={Evaluation framework for the design of an engineering model}, volume={24}, rights={https://www.cambridge.org/core/terms}, ISSN={0890-0604, 1469-1760}, DOI={10.1017/S0890060409000171}, abstractNote={According to both cybernetics and general system theory, a subject develops and uses an adequate model of a system to widen his/her knowledge about the system. Models are then the interface between a subject and a real-world system to solve a problem and to construct knowledge. Hence, evaluating these models is crucial to ensure the quality of the constructed knowledge. We propose here an evaluation framework to assess complex models based on the intrinsic properties of these models as well as the properties of the derived knowledge. A series of 40 evaluation criteria are proposed under the four systemic axes: ontology, functioning, evolution, and teleology. Through a case study, we show how our evaluation model allows both presenting a given model and assessing it.}, number={1}, journal={Artificial Intelligence for Engineering Design, Analysis and Manufacturing}, author={Ben Ahmed, Walid and Mekhilef, Mounib and Yannou, Bernard and Bigand, Michel}, year={2010}, month=feb, pages={107–125}, language={en} }

@article{Boudier_2023, title={Idea evaluation as a design process: understanding how experts develop ideas and manage fixations}, volume={9}, ISSN={2053-4701}, DOI={10.1017/dsj.2023.7}, abstractNote={Idea evaluation is used to identify and select ideas for development as future innovations. However, approaching idea evaluation as a decision gate can limit the role of the person evaluating ideas, create fixation bias, and underutilise the person’s creative potential. Although studies show that during evaluation experts are able to engage in design activities, it is still not clear how they design and develop ideas. The aim of this study was to understand how experts develop ideas during evaluation. Using the think-aloud technique, we identify different ways in which experts develop ideas. Specifically, we show how experts transform initial idea concepts using iterative steps of elaboration and transformation of different idea components. Then, relying on concept-knowledge theory (C-K theory), we identify six types of reasoning that the experts use during idea evaluation. This helps us to distinguish between three different roles that experts can move between during evaluation: gatekeeper, designer managing fixation, and designer managing defixation. These findings suggest that there is value in viewing idea evaluation as a design process because it allows us to identify and leverage the experts’ knowledge and creativity to a fuller extent.}, journal={Design Science}, author={Boudier, Justine and Sukhov, Alexandre and Netz, Johan and Le Masson, Pascal and Weil, Benoit}, year={2023}, pages={e9}, language={en} }

@inproceedings{Bryant_2005, address={Long Beach, California, USA}, title={A Computational Technique for Concept Generation}, ISBN={978-0-7918-4742-8}, DOI={10.1115/DETC2005-85323}, abstractNote={Few computational tools exist to assist designers during the conceptual phase of design, and design success is often heavily weighted on personal experience and innate ability. Many well-known methods (e.g. brainstorming, intrinsic and extrinsic searches, and morphological analysis) are designed to stimulate a designer’s creativity, but ultimately still rely heavily on individual bias and experience. Under the premise that quality designs comes from experienced designers, experience in the form of design knowledge is extracted from existing products and stored for reuse in a web-based repository. This paper presents an automated concept generation tool that utilizes the repository of existing design knowledge to generate and evaluate conceptual design variants. This tool is intended to augment traditional conceptual design phase activities and produce numerous feasible concepts early in the design process.}, booktitle={Volume 5a: 17th International Conference on Design Theory and Methodology}, publisher={ASMEDC}, author={Bryant, Cari R. and McAdams, Daniel A. and Stone, Robert B. and Kurtoglu, Tolga and Campbell, Matthew I.}, year={2005}, month=jan, pages={267–276}, language={en} }

@misc{blagec2022globalanalysismetricsused,
      title={A global analysis of metrics used for measuring performance in natural language processing}, 
      author={Kathrin Blagec and Georg Dorffner and Milad Moradi and Simon Ott and Matthias Samwald},
      year={2022},
      eprint={2204.11574},
      archivePrefix={arXiv},
      primaryClass={cs.CL},
}

@incollection{PUCCIO2012189,
title = {Chapter 9 - Idea Generation and Idea Evaluation: Cognitive Skills and Deliberate Practices},
editor = {Michael D. Mumford},
booktitle = {Handbook of Organizational Creativity},
publisher = {Academic Press},
address = {San Diego},
pages = {189-215},
year = {2012},
isbn = {978-0-12-374714-3},
doi = {10.1016/B978-0-12-374714-3.00009-4},
author = {Gerard J. Puccio and John F. Cabra}
}

@article{Christensen_Ball_2016, title={Dimensions of creative evaluation: Distinct design and reasoning strategies for aesthetic, functional and originality judgments}, volume={45}, ISSN={0142694X}, DOI={10.1016/j.destud.2015.12.005}, abstractNote={We examined evaluative reasoning taking place during expert ‘design critiques’. We focused on key dimensions of creative evaluation (originality, functionality and aesthetics) and ways in which these dimensions impact reasoning strategies and suggestions oﬀered by experts for how the student could continue. Each dimension was associated with a speciﬁc underpinning ‘logic’ determining how these dimensions were evaluated in practice. Our analysis clariﬁed how these dimensions triggered reasoning strategies such as running mental simulations or making design suggestions, ranging from ‘go/kill’ decisions to loose recommendations to continue without directional steer. The ﬁndings advance our theoretical understanding of evaluation behaviour in design and alert practicing design evaluators to the nature and consequences of their critical appraisals. Ó 2016 Elsevier Ltd. All rights reserved.}, journal={Design Studies}, author={Christensen, Bo T. and Ball, Linden J.}, year={2016}, month=jul, pages={116–136}, language={en} }

@inproceedings{Desmond_Ashktorab_Pan_Dugan_Johnson_2024, address={Greenville SC USA}, title={EvaluLLM: LLM assisted evaluation of generative outputs}, DOI={10.1145/3640544.3645216}, abstractNote={With the rapid improvement in large language model (LLM) capabilities, its becoming more difficult to measure the quality of outputs generated by natural language generation (NLG) systems. Conventional metrics such as BLEU and ROUGE are bound to reference data, and are generally unsuitable for tasks that require creative or diverse outputs. Human evaluation is an option, but manually evaluating generated text is difficult to do well, and expensive to scale and repeat as requirements and quality criteria change. Recent work has focused on the use of LLMs as customize-able NLG evaluators, and initial results are promising. In this demonstration we present EvaluLLM, an application designed to help practitioners setup, run and review evaluation over sets of NLG outputs, using an LLM as a custom evaluator. Evaluation is formulated as a series of choices between pairs of generated outputs conditioned on a user provided evaluation criteria. This approach simplifies the evaluation task and obviates the need for complex scoring algorithms. The system can be applied to general evaluation, human assisted evaluation, and model selection problems.}, booktitle={Companion Proceedings of the 29th International Conference on Intelligent User Interfaces}, publisher={ACM}, author={Desmond, Michael and Ashktorab, Zahra and Pan, Qian and Dugan, Casey and Johnson, James M.}, year={2024}, month=mar, pages={30–32}, language={en} }

@article{Fiorineschi_Rotini_2023, title={Uses of the novelty metrics proposed by Shah et al. : what emerges from the literature?}, volume={9}, ISSN={2053-4701}, DOI={10.1017/dsj.2023.9}, abstractNote={Several concepts and types of procedures for assessing novelty and related concepts exist in the literature. Among them, the two approaches originally proposed by Shah and colleagues are often considered by scholars. These metrics rely on well-defined novelty types and a specific concept of novelty; however, more than 20 years after the first publication, it is still not clear whether and to what extent these metrics are actually used, why they are used and how. Through a comprehensive review of the papers citing the main work of Shah, VargasHernandez \& Smith (2003a, 2003b) (the main study where the metrics are comprehensively described and applied), the present work aims to bridge this gap. The results highlight that only a few of the citing papers actually use the assessment approach proposed by Shah et al. and that a nonnegligible number uses a modified or adapted version of the original metrics. Furthermore, several criticalities in the application of the metrics have been uncovered, which are expected to provide relevant information for scholars involved in reliable and repeatable novelty assessments.}, journal={Design Science}, author={Fiorineschi, Lorenzo and Rotini, Federico}, year={2023}, pages={e11}, language={en} }

@article{Gonzalez_Moran_Houde_He_Ross_Muller_Kunde_Weisz, title={Collaborative Canvas: A Tool for Exploring LLM Use in Group Ideation Tasks}, abstractNote={We present the Collaborative Canvas, a prototype tool for exploring ways for groups to interact with large language models (LLMs) in ideation tasks. Collaborative Canvas provides a shared, graphical canvas in which multiple parties – human and LLM – can share ideas in the form of virtual “sticky notes” that can be moved around the canvas. The development of Collaborative Canvas raised numerous issues about the role of an LLM in group interactions: is it useful, what role does it play within the group’s workflow, and how do people interact with generated content? A preliminary examination of the Collaborative Canvas shows that users found the generative capabilities to be useful, although they preferred to review and filter generated content before sharing it with the group. Users also speculated that the role of the AI could extend into facilitating group brainstorming rather than being confined to idea generation. Our work motivates the study of human-AI co-creation in group settings beyond dyadic interactions.}, author={Gonzalez, Gabriel Enrique and Moran, Dario Andres Silva and Houde, Stephanie and He, Jessica and Ross, Steven I and Muller, Michael and Kunde, Siya and Weisz, Justin D}, language={en}, journal={}, year={} }

@article{Han_Shi_Chen_Childs_2018, title={The Combinator – a computer-based tool for creative idea generation based on a simulation approach}, volume={4}, rights={http://creativecommons.org/licenses/by/4.0/}, ISSN={2053-4701}, DOI={10.1017/dsj.2018.7}, abstractNote={Idea generation is significant in design, but coming up with creative ideas is often challenging. This paper presents a computer-based tool, called the Combinator, for assisting designers to produce creative ideas. The tool is developed based on an approach simulating aspects of human cognition in achieving combinational creativity. It can generate combinational prompts in text and image forms through combining unrelated ideas. A case study has been conducted to evaluate the Combinator. The study results indicate that the Combinator, in its current formulation, has assisted the tool users involved in the case study in improving the fluency of idea generation, as well as increasing the originality, usefulness, and flexibility of the ideas generated. The results also indicate that the tool could benefit its users in generating high-novelty and high-quality ideas eﬀectively. The Combinator is considered to be beneficial in expanding the design space, increasing better idea occurrence, improving design space exploration, and enhancing the design success rate.}, journal={Design Science}, author={Han, Ji and Shi, Feng and Chen, Liuqing and Childs, Peter R. N.}, year={2018}, pages={e11}, language={en} }

@article{Karimi_Maher_Davis_Grace_2019, title={Deep Learning in a Computational Model for Conceptual Shifts in a Co-Creative Design System}, abstractNote={This paper presents a computational model for conceptual shifts, based on a novelty metric applied to a vector representation generated through deep learning. This model is integrated into a co-creative design system, which enables a partnership between an AI agent and a human designer interacting through a sketching canvas. The AI agent responds to the human designer’s sketch with a new sketch that is a conceptual shift: intentionally varying the visual and conceptual similarity with increasingly more novelty. The paper presents the results of a user study showing that increasing novelty in the AI contribution is associated with higher creative outcomes, whereas low novelty leads to less creative outcomes.}, note={arXiv:1906.10188 [cs, stat]}, number={arXiv:1906.10188}, publisher={arXiv}, author={Karimi, Pegah and Maher, Mary Lou and Davis, Nicholas and Grace, Kazjon}, year={2019}, month=jun, language={en}, journal={} }

@article{Kim_Maher_2023, title={The effect of AI-based inspiration on human design ideation}, volume={11}, ISSN={2165-0349, 2165-0357}, DOI={10.1080/21650349.2023.2167124}, abstractNote={Computational co-creative systems in design allow users to collaborate with an AI partner on open-ended creative tasks in the design process. Cocreative systems can enhance design creativity by inspiring the explora­ tion of novel design solutions in the initial idea generation. However, there are a lack of studies about the effect of co-creative systems on the cognitive process during ideation. This study examines the effect of an AIbased co-creative design tool that provides inspirations based on con­ ceptual similarity on design ideation. It was hypothesized that concep­ tually similar inspirations have a significant influence on design ideation than random inspirations. The Collaborative Ideation Partner (CIP), a cocreative design system that provides inspirational images based on con­ ceptual similarity, was developed to examine the effect of an AI Model for conceptual similarity on ideation during a design task. We conducted an experiment with a control condition in which the images are selected randomly from a curated database for inspiration and a treatment condi­ tion in which conceptual similarity is the basis for selecting the next inspiring image. Our findings show that the AI model of conceptual similarity used in the treatment condition has a significant effect on the novelty, variety, and quantity of ideas during human design ideation.}, number={2}, journal={International Journal of Design Creativity and Innovation}, author={Kim, Jingoog and Maher, Mary Lou}, year={2023}, month=apr, pages={81–98}, language={en} }

@article{Kurtoglu_Campbell_Linsey_2009, title={An experimental study on the effects of a computational design tool on concept generation}, volume={30}, rights={https://www.elsevier.com/tdm/userlicense/1.0/}, ISSN={0142694X}, DOI={10.1016/j.destud.2009.06.005}, abstractNote={We have developed a computational design tool to help designers create conceptual solutions to detailed functional speciﬁcations. The computational method extracts design knowledge from an expanding online design library in the form of procedural rules, and provides these rules as the building blocks for solving new problems. In this paper, we study how this automated approach would beneﬁt designers during concept generation. Accordingly, we test the eﬀects of using our computational tool as an aid for concept generation in an experiment mimicking real design scenarios. Three metrics (completeness, novelty and variety) are used to evaluate the solutions generated to two separate design problems in order to determine how eﬀective the computational method outputs are in improving conceptual design generation. Ó 2009 Elsevier Ltd. All rights reserved.}, number={6}, journal={Design Studies}, author={Kurtoglu, Tolga and Campbell, Matthew I. and Linsey, Julie S.}, year={2009}, month=nov, pages={676–703}, language={en} }

@article{Linsey_Clauss_Kurtoglu_Murphy_Wood_Markman_2011, title={An Experimental Study of Group Idea Generation Techniques: Understanding the Roles of Idea Representation and Viewing Methods}, volume={133}, ISSN={1050-0472, 1528-9001}, DOI={10.1115/1.4003498}, abstractNote={Advances in innovation processes are critically important as economic and business landscapes evolve. There are many concept generation techniques that can assist a designer in the initial phases of design. Unfortunately, few studies have examined these techniques that can provide evidence to suggest which techniques should be preferred or how to implement them in an optimal way. This study systematically investigates the underlying factors of four common and well-documented techniques: brainsketching, gallery, 6-3-5, and C-sketch. These techniques are resolved into their key parameters, and a rigorous factorial experiment is performed to understand how the key parameters affect the outcomes of the techniques. The factors chosen for this study with undergraduate mechanical engineers include how concepts are displayed to participants (all are viewed at once or subsets are exchanged between participants, i.e., “rotational viewing”) and the mode used to communicate ideas (written words only, sketches only, or a combination of written words and sketches). Four metrics are used to evaluate the data: quantity, quality, novelty, and variety. The data suggest that rotational viewing of sets of concepts described using sketches combined with words produces more ideas than having all concepts displayed in a “gallery view” form, but a gallery view results in more high quality concepts. These results suggest that a hybrid of methods should be used to maximize the quality and number of ideas. The study also shows that individuals gain a significant number of ideas from their teammates. Ideas, when shared, can foster new idea tracks, more complete layouts, and a diverse synthesis. Finally, as teams develop more concepts, the quality of the concepts improves. This result is a consequence of the team-sharing environment and, in conjunction with the quantity of ideas, validates the effectiveness of group idea generation. This finding suggests a way to go beyond the observation that some forms of brainstorming can actually hurt productivity.}, number={3}, journal={Journal of Mechanical Design}, author={Linsey, J. S. and Clauss, E. F. and Kurtoglu, T. and Murphy, J. T. and Wood, K. L. and Markman, A. B.}, year={2011}, month=mar, pages={031008}, language={en} }

@article{Liu_Iter_Xu_Wang_Xu_Zhu_2023, title={G-Eval: NLG Evaluation using GPT-4 with Better Human Alignment},  abstractNote={The quality of texts generated by natural language generation (NLG) systems is hard to measure automatically. Conventional reference-based metrics, such as BLEU and ROUGE, have been shown to have relatively low correlation with human judgments, especially for tasks that require creativity and diversity. Recent studies suggest using large language models (LLMs) as reference-free metrics for NLG evaluation, which have the benefit of being applicable to new tasks that lack human references. However, these LLM-based evaluators still have lower human correspondence than medium-size neural evaluators. In this work, we present G-Eval, a framework of using large language models with chain-of-thoughts (CoT) and a form-filling paradigm, to assess the quality of NLG outputs. We experiment with two generation tasks, text summarization and dialogue generation. We show that G-Eval with GPT-4 as the backbone model achieves a Spearman correlation of 0.514 with human on summarization task, outperforming all previous methods by a large margin. We also propose preliminary analysis on the behavior of LLM-based evaluators, and highlight the potential issue of LLM-based evaluators having a bias towards the LLM-generated texts. The code is at https://github.com/nlpyang/geval}, note={arXiv:2303.16634 [cs]}, number={arXiv:2303.16634}, publisher={arXiv}, author={Liu, Yang and Iter, Dan and Xu, Yichong and Wang, Shuohang and Xu, Ruochen and Zhu, Chenguang}, year={2023}, month=may, language={en}, journal={} }

@article{hunt-earl,
author = {Hunt, Earl},
title = {Creative approches to cognition. Creative cognition: Theory, research and applications. R. A. Finke, T. B. Ward and S. M. Smith. Cambridge, MA: MIT Press (Bradford Books), 1992. No. of pages 240. ISBN 0-262-06150-3. Price \$24.95 (Hard cover)},
journal = {Applied Cognitive Psychology},
volume = {8},
number = {5},
pages = {528-529},
doi = {10.1002/acp.2350080511},
year = {1994}
}

@ARTICLE{Ward2004-mw,
  title     = "Cognition, creativity, and entrepreneurship",
  author    = "Ward, Thomas B",
  journal   = "J. Bus. Venturing",
  publisher = "Elsevier BV",
  volume    =  19,
  number    =  2,
  pages     = "173--188",
  month     =  mar,
  year      =  2004,
  language  = "en"
}

@ARTICLE{Fauconnier2003,
  title     = "Conceptual blending, form and meaning",
  author    = "Fauconnier, Gilles and Turner, Mark",
  journal   = "ReC",
  publisher = "Universite Catholique de Louvain",
  volume    =  19,
  month     =  mar,
  year      =  2003
}

@article{Nelson_Wilson_Rosen_Yen_2009, title={Refined metrics for measuring ideation effectiveness}, volume={30}, rights={https://www.elsevier.com/tdm/userlicense/1.0/}, ISSN={0142694X}, DOI={10.1016/j.destud.2009.07.002}, abstractNote={Idea generation is an important step in the engineering design process, and as a result signiﬁcant research eﬀorts have focused on developing methods to aid designers in exploring design possibilities. Metrics to evaluate design exploration are thus necessary to make conclusions and comparisons among idea generation methods. Metrics have previously been proposed, identifying novelty, variety, quantity, and quality to characterize sets of designs and the degree to which they describe design space exploration. This article describes ﬂaws in the variety metric and proposes a new metric to eliminate the ﬂaws. Additionally, a single metric is proposed to evaluate the quality of design space exploration during concept generation, enabling application of a single metric to compare idea generation processes and methodologies. Ó 2009 Elsevier Ltd. All rights reserved.}, number={6}, journal={Design Studies}, author={Nelson, Brent A. and Wilson, Jamal O. and Rosen, David and Yen, Jeannette}, year={2009}, month=nov, pages={737–743}, language={en} }

@article{Shaer_Cooper_Kun_Mokryn, title={Toward Enhancing Ideation through Collaborative Group-AI Brainwriting}, abstractNote={This paper introduces a collaborative group-AI framework for enhancing ideation through co-creation. The proposed framework integrates LLMs into the creative process to support both the divergence stage of idea generation and the convergence stage of evaluation and selection of a few chosen ideas. We describe the framework and the tools we designed to implement it as well as summarize findings from its evaluation with novice designers - students in an advanced interaction design course. Our findings suggest that the framework could enhance both the ideation process and its outcome through human-AI co-creation.}, author={Shaer, Orit and Cooper, Angelora and Kun, Andrew L and Mokryn, Osnat}, language={en}, journal={}, year={} }

@article{Shaer_Cooper_Mokryn_Kun_Shoshan_2024, title={AI-Augmented Brainwriting: Investigating the use of LLMs in group ideation}, abstractNote={The growing availability of generative AI technologies such as large language models (LLMs) has significant implications for creative work. This paper explores twofold aspects of integrating LLMs into the creative process - the divergence stage of idea generation, and the convergence stage of evaluation and selection of ideas. We devised a collaborative group-AI Brainwriting ideation framework, which incorporated an LLM as an enhancement into the group ideation process, and evaluated the idea generation process and the resulted solution space. To assess the potential of using LLMs in the idea evaluation process, we design an evaluation engine and compared it to idea ratings assigned by three expert and six novice evaluators. Our findings suggest that integrating LLM in Brainwriting could enhance both the ideation process and its outcome. We also provide evidence that LLMs can support idea evaluation. We conclude by discussing implications for HCI education and practice.}, note={arXiv:2402.14978 [cs]}, number={arXiv:2402.14978}, publisher={arXiv}, author={Shaer, Orit and Cooper, Angelora and Mokryn, Osnat and Kun, Andrew L. and Shoshan, Hagit Ben}, year={2024}, month=feb, language={en}, journal={} }

@article{Shah2003,
  title = {Metrics for measuring ideation effectiveness},
  volume = {24},
  ISSN = {0142-694X},
  DOI = {10.1016/s0142-694x(02)00034-0},
  number = {2},
  journal = {Design Studies},
  publisher = {Elsevier BV},
  author = {Shah,  Jami J. and Smith,  Steve M. and Vargas-Hernandez,  Noe},
  year = {2003},
  month = mar,
  pages = {111–134}
}

@misc{Altshuller1984,
  title = {Creativity As an Exact Science},
  ISBN = {9780429073793},
  DOI = {10.1201/9781466593442},
  publisher = {CRC Press},
  author = {Altshuller},
  year = {1984},
  month = jan 
}

@inproceedings{Cheeley2018,
  series = {IDETC-CIE2018},
  title = {A Proposed Quality Metric for Ideation Effectiveness},
  DOI = {10.1115/detc2018-85401},
  booktitle = {Volume 7: 30th International Conference on Design Theory and Methodology},
  publisher = {American Society of Mechanical Engineers},
  author = {Cheeley,  Avery and Weaver,  Morgan B. and Bennetts,  Caleb and Caldwell,  Benjamin W. and Green,  Matthew G.},
  year = {2018},
  month = aug,
  collection = {IDETC-CIE2018}
}

@inproceedings{Srivathsavai2010,
  series = {IDETC-CIE2010},
  title = {Study of Existing Metrics Used in Measurement of Ideation Effectiveness},
  DOI = {10.1115/detc2010-28802},
  booktitle = {Volume 5: 22nd International Conference on Design Theory and Methodology; Special Conference on Mechanical Vibration and Noise},
  publisher = {ASMEDC},
  author = {Srivathsavai,  Ramesh and Genco,  Nicole and Ho¨ltta¨-Otto,  Katja and Seepersad,  Carolyn C.},
  year = {2010},
  month = jan,
  pages = {355–366},
  collection = {IDETC-CIE2010}
}

@article{Milan2016,
  author = {Milan Stevanovic, Dorian Marjanović, Mario ŠtorgaMario},
  title = {Idea Assessment and Selection in Product Innovation - The Empirical Research Results},
  volume = {23},
  ISSN = {1848-6339},
  DOI = {10.17559/tv-20151103120545},
  number = {6},
  journal = {Tehnicki vjesnik - Technical Gazette},
  publisher = {Mechanical Engineering Faculty in Slavonski Brod},
  year = {2016},
  month = nov 
}

@article{Mirabito2021,
  title = {Factors Impacting Highly Innovative Designs: Idea Fluency,  Timing,  and Order},
  volume = {144},
  ISSN = {1528-9001},
  DOI = {10.1115/1.4051683},
  number = {1},
  journal = {Journal of Mechanical Design},
  publisher = {ASME International},
  author = {Mirabito,  Yakira and Goucher-Lambert,  Kosa},
  year = {2021},
  month = jul 
}

@inbook{Plucker2010,
  title = {Assessment of Creativity},
  ISBN = {9780521730259},
  DOI = {10.1017/cbo9780511763205.005},
  booktitle = {The Cambridge Handbook of Creativity},
  publisher = {Cambridge University Press},
  author = {Plucker,  Jonathan A. and Makel,  Matthew C.},
  year = {2010},
  month = aug,
  pages = {48–73}
}

@article{Organisciak2023,
  title = {Beyond semantic distance: Automated scoring of divergent thinking greatly improves with large language models},
  volume = {49},
  ISSN = {1871-1871},
  DOI = {10.1016/j.tsc.2023.101356},
  journal = {Thinking Skills and Creativity},
  publisher = {Elsevier BV},
  author = {Organisciak,  Peter and Acar,  Selcuk and Dumas,  Denis and Berthiaume,  Kelly},
  year = {2023},
  month = sep,
  pages = {101356}
}

@article{Dumas2021,
  title = {Measuring divergent thinking originality with human raters and text-mining models: A psychometric comparison of methods.},
  volume = {15},
  ISSN = {1931-3896},
  DOI = {10.1037/aca0000319},
  number = {4},
  journal = {Psychology of Aesthetics,  Creativity,  and the Arts},
  publisher = {American Psychological Association (APA)},
  author = {Dumas,  Denis and Organisciak,  Peter and Doherty,  Michael},
  year = {2021},
  month = nov,
  pages = {645–663}
}

@book{stevenson2020a,
  author = {Stevenson, C. and Smal, I. and Baas, M. and Dahrendorf, M. and Grasman, R. and Tanis, C. and Scheurs, E. and Sleiffer, D. and Maas, H.},
  date = {2020},
  title = {Automated AUT scoring using a Big Data variant of the Consensual Assessment Technique: Final Technical Report},
  publisher = {Modeling Creativity Project, Universiteit van Amsterdam},
  url = {http://modelingcreativity.org/blog/wpcontent/uploads/2020/07/ABBAS_report_200711_final.pdf},
  language = {en}
}

@ARTICLE{Acar2014,
  title     = "Assessing associative distance among ideas elicited by tests of
               divergent thinking",
  author    = "Acar, Selcuk and Runco, Mark A",
  journal   = "Creat. Res. J.",
  publisher = "Informa UK Limited",
  volume    =  26,
  number    =  2,
  pages     = "229--238",
  month     =  apr,
  year      =  2014,
  language  = "en"
}

@ARTICLE{Acar2023,
  title     = "Applying automated originality scoring to the Verbal Form of
               Torrance Tests of Creative Thinking",
  author    = "Acar, Selcuk and Berthiaume, Kelly and Grajzel, Katalin and
               Dumas, Denis and Flemister, Charles ``tedd'' and Organisciak,
               Peter",
  journal   = "Gift. Child Q.",
  publisher = "SAGE Publications",
  volume    =  67,
  number    =  1,
  pages     = "3--17",
  month     =  jan,
  year      =  2023,
  language  = "en"
}

@ARTICLE{Kudrowitz2013,
  title     = "Assessing the quality of ideas from prolific, early-stage
               product ideation",
  author    = "Kudrowitz, Barry Matthew and Wallace, David",
  journal   = "J. Eng. Des.",
  publisher = "Informa UK Limited",
  volume    =  24,
  number    =  2,
  pages     = "120--139",
  month     =  feb,
  year      =  2013,
  language  = "en"
}

@ARTICLE{Beaty2021,
  title     = "Automating creativity assessment with {SemDis}: An open platform
               for computing semantic distance",
  author    = "Beaty, Roger E and Johnson, Dan R",
  journal   = "Behav. Res. Methods",
  publisher = "Springer Science and Business Media LLC",
  volume    =  53,
  number    =  2,
  pages     = "757--780",
  month     =  apr,
  year      =  2021,
  copyright = "https://creativecommons.org/licenses/by/4.0",
  language  = "en"
}

@ARTICLE{Beaty2022,
  title     = "Semantic distance and the alternate uses task: Recommendations for reliable automated assessment of originality",
  author    = "Beaty, Roger E and Johnson, Dan R and Zeitlen, Daniel C and
               Forthmann, Boris",
  journal   = "Creat. Res. J.",
  publisher = "Informa UK Limited",
  volume    =  34,
  number    =  3,
  pages     = "245--260",
  month     =  jul,
  year      =  2022,
  language  = "en"
}

@ARTICLE{Beaty2023,
  title    = "Semantic memory and creativity: The costs and benefits of
              semantic memory structure in generating original ideas",
  author   = "Beaty, Roger E and Kenett, Yoed N and Hass, Richard W and
              Schacter, Daniel L",
  journal  = "Think. Reason.",
  volume   =  29,
  number   =  2,
  pages    = "305--339",
  year     =  2023,
  language = "en"
}

@ARTICLE{Benedek2013,
  title     = "Assessment of divergent thinking by means of the subjective top-scoring method: Effects of the number of top-ideas and time-on-task on reliability and validity",
  author    = "Benedek, Mathias and M{\"u}hlmann, Caterina and Jauk, Emanuel and Neubauer, Aljoscha C",
  journal   = "Psychol. Aesthet. Creat. Arts",
  publisher = "American Psychological Association (APA)",
  volume    =  7,
  number    =  4,
  pages     = "341--349",
  month     =  nov,
  year      =  2013,
  language  = "en"
}

@ARTICLE{Bossomaier2009,
  title     = "A semantic network approach to the creativity quotient ({CQ})",
  author    = "Bossomaier, Terry and Harr{\'e}, Mike and Knittel, Anthony and Snyder, Allan",
  journal   = "Creat. Res. J.",
  publisher = "Informa UK Limited",
  volume    =  21,
  number    =  1,
  pages     = "64--71",
  month     =  feb,
  year      =  2009,
  language  = "en"
}

@ARTICLE{Carson2005,
  title     = "Reliability, validity, and factor structure of the creative
               achievement questionnaire",
  author    = "Carson, Shelley H and Peterson, Jordan B and Higgins, Daniel M",
  journal   = "Creat. Res. J.",
  publisher = "Informa UK Limited",
  volume    =  17,
  number    =  1,
  pages     = "37--50",
  month     =  feb,
  year      =  2005,
  language  = "en"
}

@ARTICLE{Ceh2021,
  title     = "Assessing raters: What factors predict discernment in novice
               creativity raters?",
  author    = "Ceh, Simon Majed and Edelmann, Carina and Hofer, Gabriela and
               Benedek, Mathias",
  journal   = "J. Creat. Behav.",
  publisher = "Wiley",
  number    = "jocb.515",
  month     =  jul,
  year      =  2021,
  copyright = "http://creativecommons.org/licenses/by/4.0/",
  language  = "en"
}

@article{Cropley11042024,
author = {David H Cropley and Caroline Theurer and A C Sven Mathijssen and Rebecca L Marrone and},
title = {Fit-For-Purpose Creativity Assessment: Automatic Scoring of the Test of Creative Thinking – Drawing Production (TCT-DP)},
journal = {Creativity Research Journal},
volume = {0},
number = {0},
pages = {1--16},
year = {2024},
publisher = {Routledge},
doi = {10.1080/10400419.2024.2339667},
}

@ARTICLE{Cseh2019,
  title     = "A scattered {CAT}: A critical evaluation of the consensual
               assessment technique for creativity research",
  author    = "Cseh, Genevieve M and Jeffries, Karl K",
  journal   = "Psychol. Aesthet. Creat. Arts",
  publisher = "American Psychological Association (APA)",
  volume    =  13,
  number    =  2,
  pages     = "159--166",
  month     =  may,
  year      =  2019,
  language  = "en"
}

@ARTICLE{Dumas2014,
  title     = "Understanding Fluency and Originality: A latent variable
               perspective",
  author    = "Dumas, Denis and Dunbar, Kevin N",
  journal   = "Think. Skills Creat.",
  publisher = "Elsevier BV",
  volume    =  14,
  pages     = "56--67",
  month     =  dec,
  year      =  2014,
  language  = "en"
}

@ARTICLE{Forthmann2017,
  title     = "Missing creativity: The effect of cognitive workload on rater (dis-)agreement in subjective divergent-thinking scores",
  author    = "Forthmann, Boris and Holling, Heinz and Zandi, Nima and Gerwig, Anne and {\c C}elik, P{\i}nar and Storme, Martin and Lubart,
               Todd",
  journal   = "Think. Skills Creat.",
  publisher = "Elsevier BV",
  volume    =  23,
  pages     = "129--139",
  month     =  mar,
  year      =  2017
}

@ARTICLE{Jackson2022,
  title     = "From text to thought: How analyzing language can advance
               psychological science",
  author    = "Jackson, Joshua Conrad and Watts, Joseph and List, Johann-Mattis and Puryear, Curtis and Drabble, Ryan and Lindquist, Kristen A",
  journal   = "Perspect. Psychol. Sci.",
  publisher = "SAGE Publications",
  volume    =  17,
  number    =  3,
  pages     = "805--826",
  month     =  may,
  year      =  2022,
  language  = "en"
}

@ARTICLE{Gunther2019,
  title     = "Vector-space models of semantic representation from a cognitive perspective: A discussion of common misconceptions",
  author    = "G{\"u}nther, Fritz and Rinaldi, Luca and Marelli, Marco",
  journal   = "Perspect. Psychol. Sci.",
  publisher = "SAGE Publications",
  volume    =  14,
  number    =  6,
  pages     = "1006--1033",
  month     =  nov,
  year      =  2019,
  language  = "en"
}

@ARTICLE{Johnson2023,
  title     = "Divergent semantic integration ({DSI)}: Extracting creativity from narratives with distributional semantic modeling",
  author    = "Johnson, Dan R and Kaufman, James C and Baker, Brendan S and Patterson, John D and Barbot, Baptiste and Green, Adam E and van Hell, Janet and Kennedy, Evan and Sullivan, Grace F and Taylor, Christa L and Ward, Thomas and Beaty, Roger E",
  journal   = "Behav. Res. Methods",
  publisher = "Springer Science and Business Media LLC",
  volume    =  55,
  number    =  7,
  pages     = "3726--3759",
  month     =  oct,
  year      =  2023,
  copyright = "https://creativecommons.org/licenses/by/4.0",
  language  = "en"
}

@ARTICLE{Dumas2021f,
  title     = "Four text‐mining methods for measuring elaboration",
  author    = "Dumas, Denis and Organisciak, Peter and Maio, Shannon and
               Doherty, Michael",
  journal   = "J. Creat. Behav.",
  publisher = "Wiley",
  volume    =  55,
  number    =  2,
  pages     = "517--531",
  month     =  jun,
  year      =  2021,
  copyright = "http://onlinelibrary.wiley.com/termsAndConditions\#vor",
  language  = "en"
}

@ARTICLE{Kenett2019,
  title     = "What can quantitative measures of semantic distance tell us about creativity?",
  author    = "Kenett, Yoed N",
  journal   = "Curr. Opin. Behav. Sci.",
  publisher = "Elsevier BV",
  volume    =  27,
  pages     = "11--16",
  month     =  jun,
  year      =  2019,
  language  = "en"
}

@ARTICLE{Patterson2023,
  title     = "Multilingual semantic distance: Automatic verbal creativity assessment in many languages",
  author    = "Patterson, John D and Merseal, Hannah M and Johnson, Dan R and
               Agnoli, Sergio and Baas, Matthijs and Baker, Brendan S and
               Barbot, Baptiste and Benedek, Mathias and Borhani, Khatereh and
               Chen, Qunlin and Christensen, Julia F and Corazza, Giovanni
               Emanuele and Forthmann, Boris and Karwowski, Maciej and
               Kazemian, Nastaran and Kreisberg-Nitzav, Ariel and Kenett, Yoed
               N and Link, Allison and Lubart, Todd and Mercier, Maxence and
               Miroshnik, Kirill and Ovando-Tellez, Marcela and Primi, Ricardo
               and Puente-D{\'\i}az, Rogelio and Said-Metwaly, Sameh and
               Stevenson, Claire and Vartanian, Meghedi and Volle, Emannuelle
               and van Hell, Janet G and Beaty, Roger E",
  journal   = "Psychol. Aesthet. Creat. Arts",
  publisher = "American Psychological Association (APA)",
  volume    =  17,
  number    =  4,
  pages     = "495--507",
  month     =  aug,
  year      =  2023,
  language  = "en"
}

@misc{stevenson2022p,
      title={Putting GPT-3's Creativity to the (Alternative Uses) Test}, 
      author={Claire Stevenson and Iris Smal and Matthijs Baas and Raoul Grasman and Han van der Maas},
      year={2022},
      eprint={2206.08932},
      archivePrefix={arXiv},
      primaryClass={cs.AI},
      url={https://arxiv.org/abs/2206.08932}, 
}

@ARTICLE{ijaz2023,
AUTHOR={Ul Haq, Ijaz  and Pifarré, Manoli },  
TITLE={Dynamics of automatized measures of creativity: mapping the landscape to quantify creative ideation},  
JOURNAL={Frontiers in Education},  
VOLUME={Volume 8 - 2023},
YEAR={2023},
URL={https://www.frontiersin.org/journals/education/articles/10.3389/feduc.2023.1240962},
DOI={10.3389/feduc.2023.1240962},
ISSN={2504-284X}}

@article{Amabile1982,
  title = {Social psychology of creativity: A consensual assessment technique.},
  volume = {43},
  ISSN = {0022-3514},
  url = {http://dx.doi.org/10.1037/0022-3514.43.5.997},
  DOI = {10.1037/0022-3514.43.5.997},
  number = {5},
  journal = {Journal of Personality and Social Psychology},
  publisher = {American Psychological Association (APA)},
  author = {Amabile,  Teresa M.},
  year = {1982},
  month = nov,
  pages = {997–1013}
}

@article{Silvia2008,
  title = {Assessing creativity with divergent thinking tasks: Exploring the reliability and validity of new subjective scoring methods.},
  volume = {2},
  ISSN = {1931-3896},
  url = {http://dx.doi.org/10.1037/1931-3896.2.2.68},
  DOI = {10.1037/1931-3896.2.2.68},
  number = {2},
  journal = {Psychology of Aesthetics,  Creativity,  and the Arts},
  publisher = {American Psychological Association (APA)},
  author = {Silvia,  Paul J. and Winterstein,  Beate P. and Willse,  John T. and Barona,  Christopher M. and Cram,  Joshua T. and Hess,  Karl I. and Martinez,  Jenna L. and Richard,  Crystal A.},
  year = {2008},
  month = may,
  pages = {68–85}
}

@article{Silvia2009,
  title = {A snapshot of creativity: Evaluating a quick and simple method for assessing divergent thinking},
  volume = {4},
  ISSN = {1871-1871},
  url = {http://dx.doi.org/10.1016/j.tsc.2009.06.005},
  DOI = {10.1016/j.tsc.2009.06.005},
  number = {2},
  journal = {Thinking Skills and Creativity},
  publisher = {Elsevier BV},
  author = {Silvia,  Paul J. and Martin,  Christopher and Nusbaum,  Emily C.},
  year = {2009},
  month = aug,
  pages = {79–85}
}

\appendix
\newpage
.
\newpage
\section{Evaluation Results: Novelty Rating Distributions of AI-Based Assessment Methods}
\subsection{Bar Plots}
The subsequent bar plot offers a visual representation of each novelty rating within the output of each AI assessment method. This allows for a more granular understanding of the rating patterns.

\begin{figure*}[!ht]
    \centering
    \begin{subfigure}{0.3\linewidth}
        \centering
        \includegraphics[width=\linewidth]{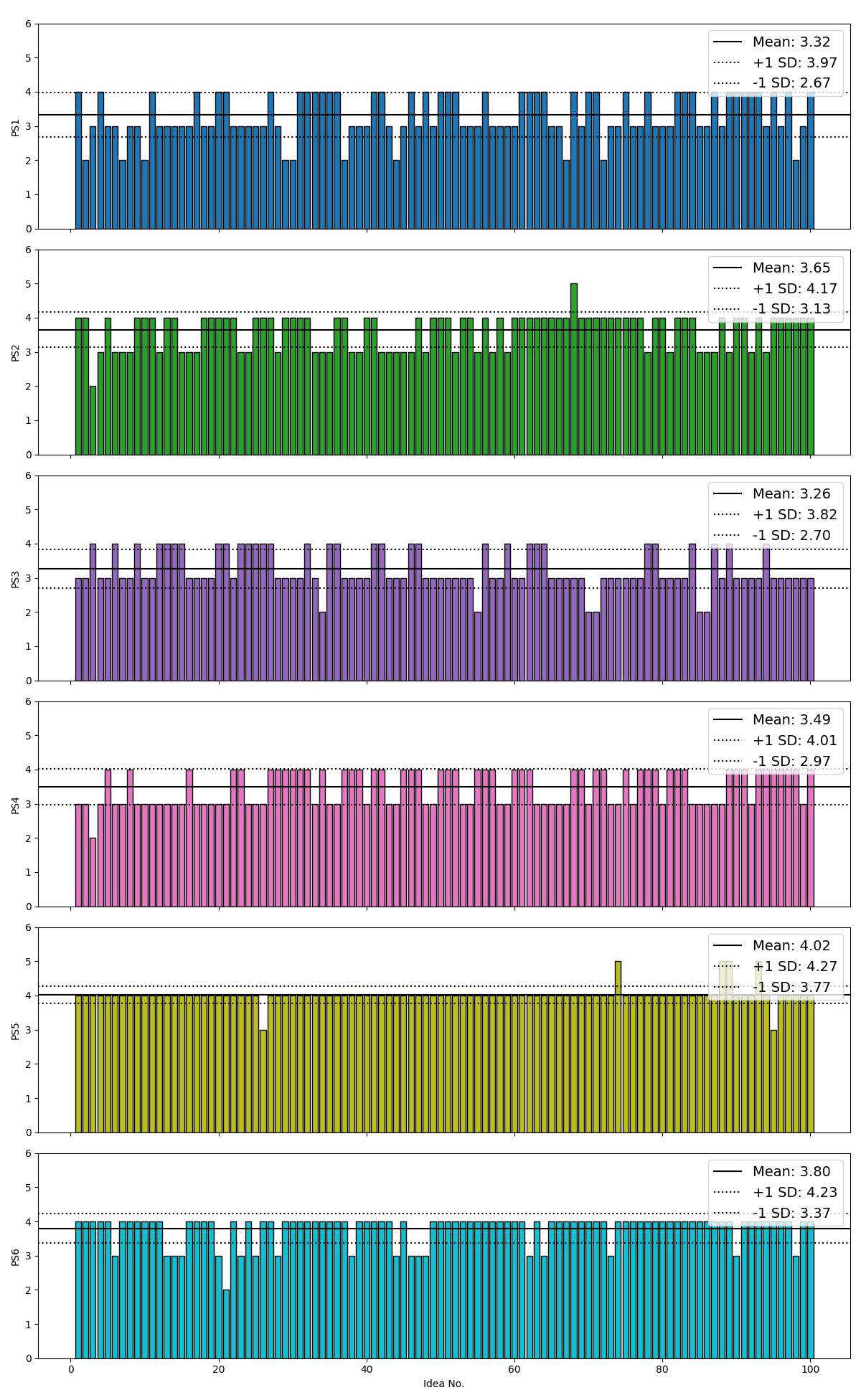}
        \caption{GPT-based Assessment}
        \label{fig:hist_gpt}
    \end{subfigure}\hfill
    \begin{subfigure}{0.3\linewidth}
        \centering
        \includegraphics[width=\linewidth]{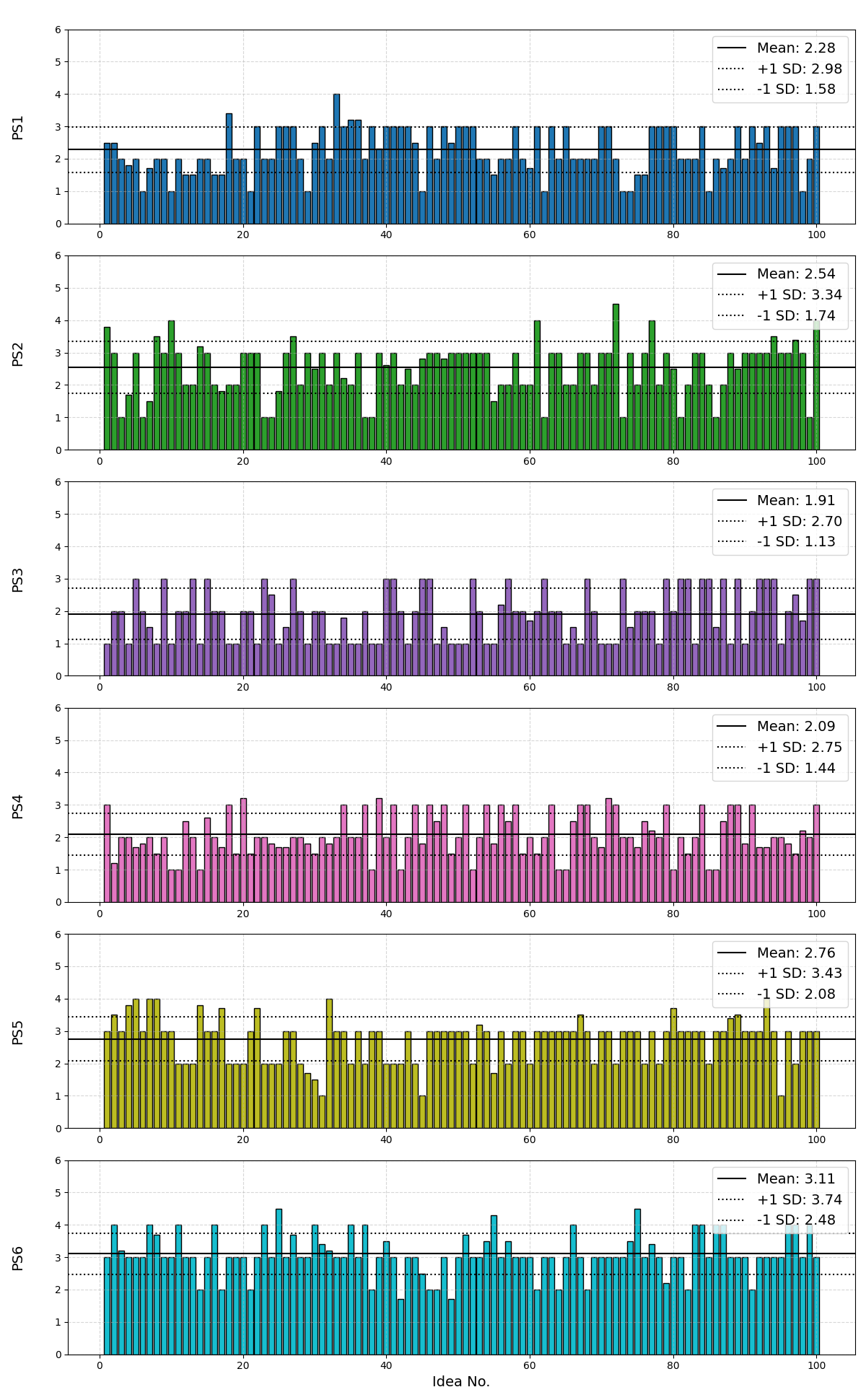}
        \caption{OCSAI-based Assessment}
        \label{fig:hist_ocsai}
    \end{subfigure}\hfill
    \begin{subfigure}{0.3\linewidth}
        \centering
        \includegraphics[width=\linewidth]{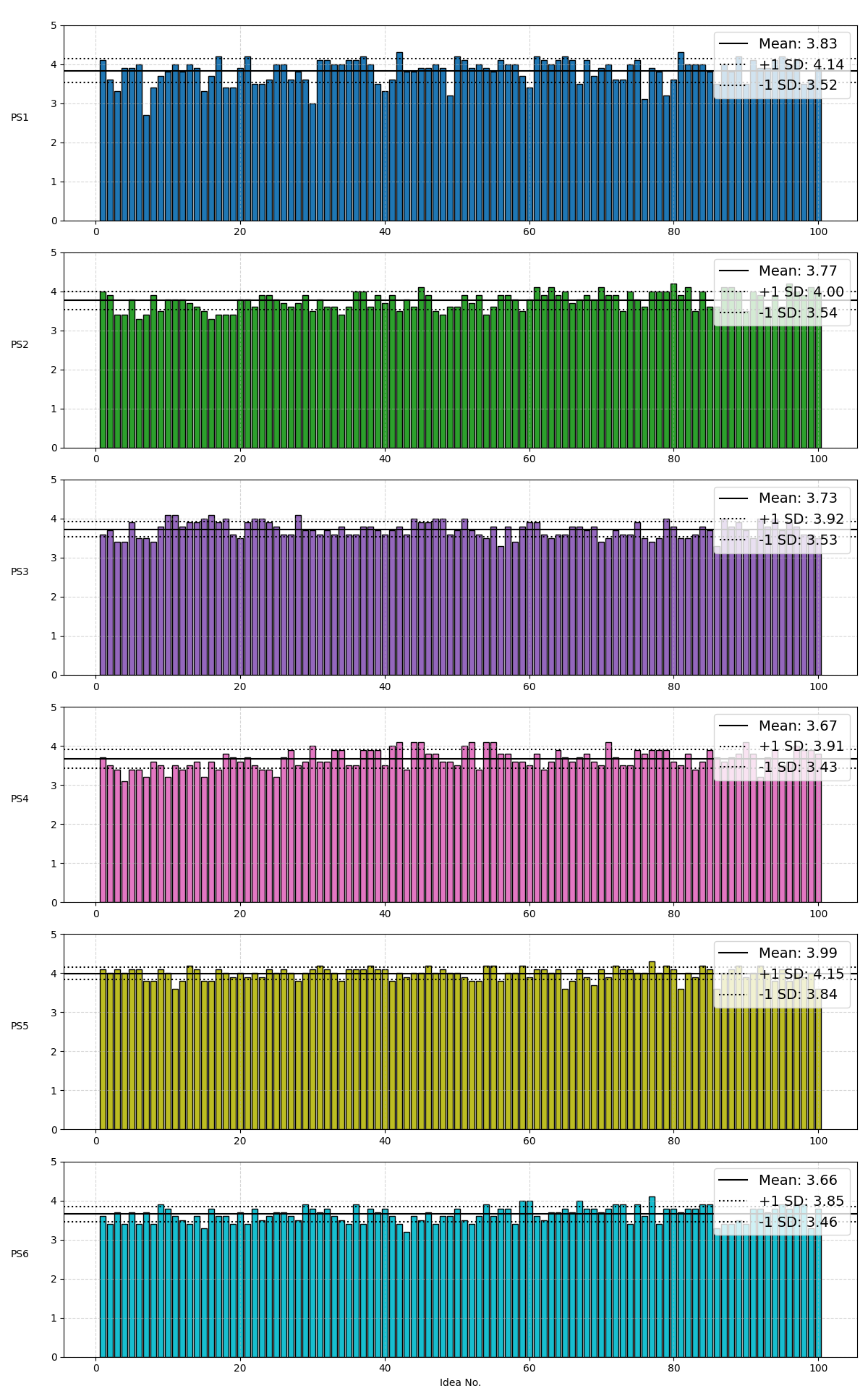}
        \caption{AIDE-based Assessment}
        \label{fig:hist_aide}
    \end{subfigure}
    
    \vspace{0.5em} 
    
    \begin{subfigure}{0.3\linewidth}
        \centering
        \includegraphics[width=\linewidth]{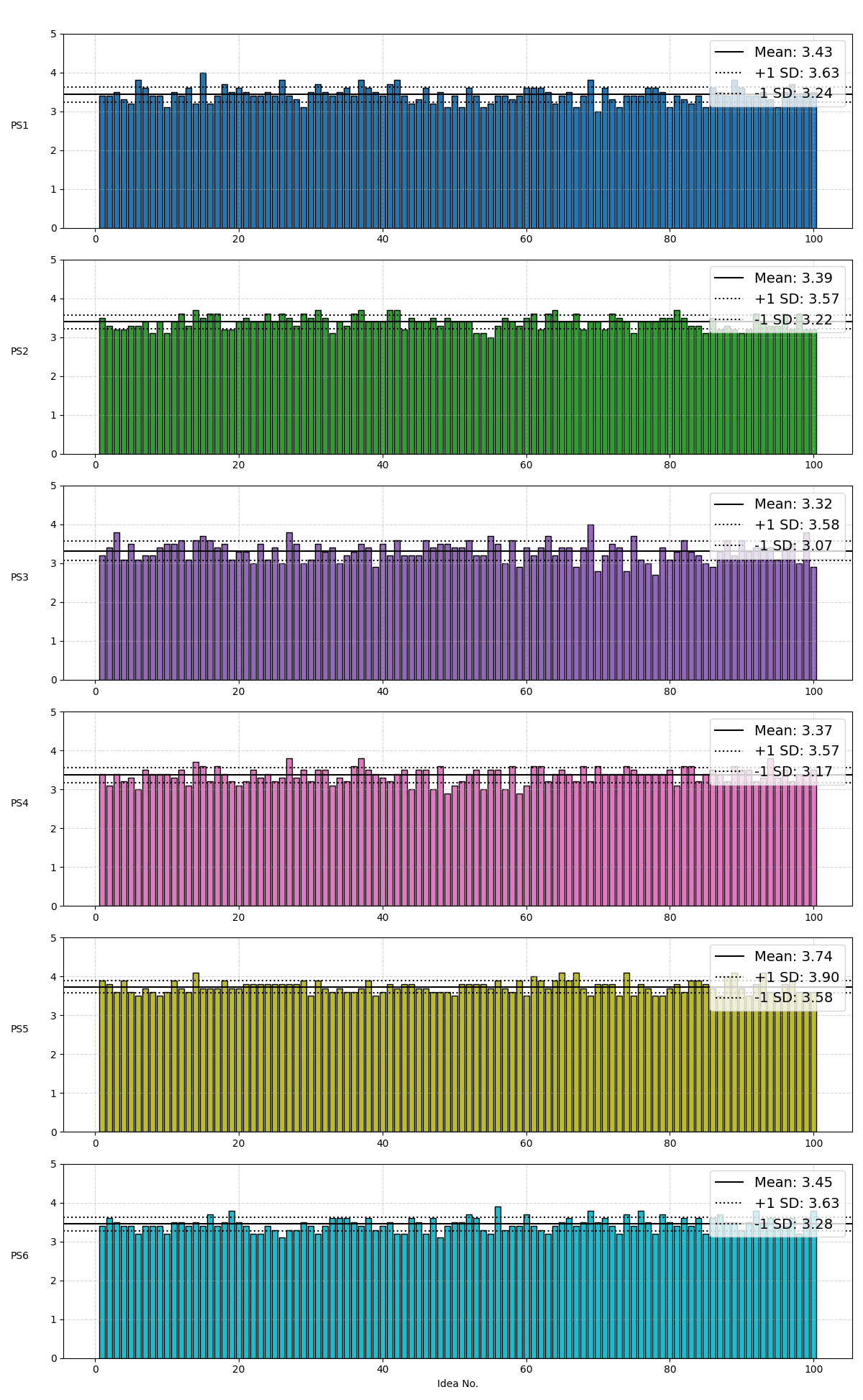}
        \caption{CLAUS-based Assessment}
        \label{fig:hist_claus}
    \end{subfigure}\hfill
    \begin{subfigure}{0.3\linewidth}
        \centering
        \includegraphics[width=\linewidth]{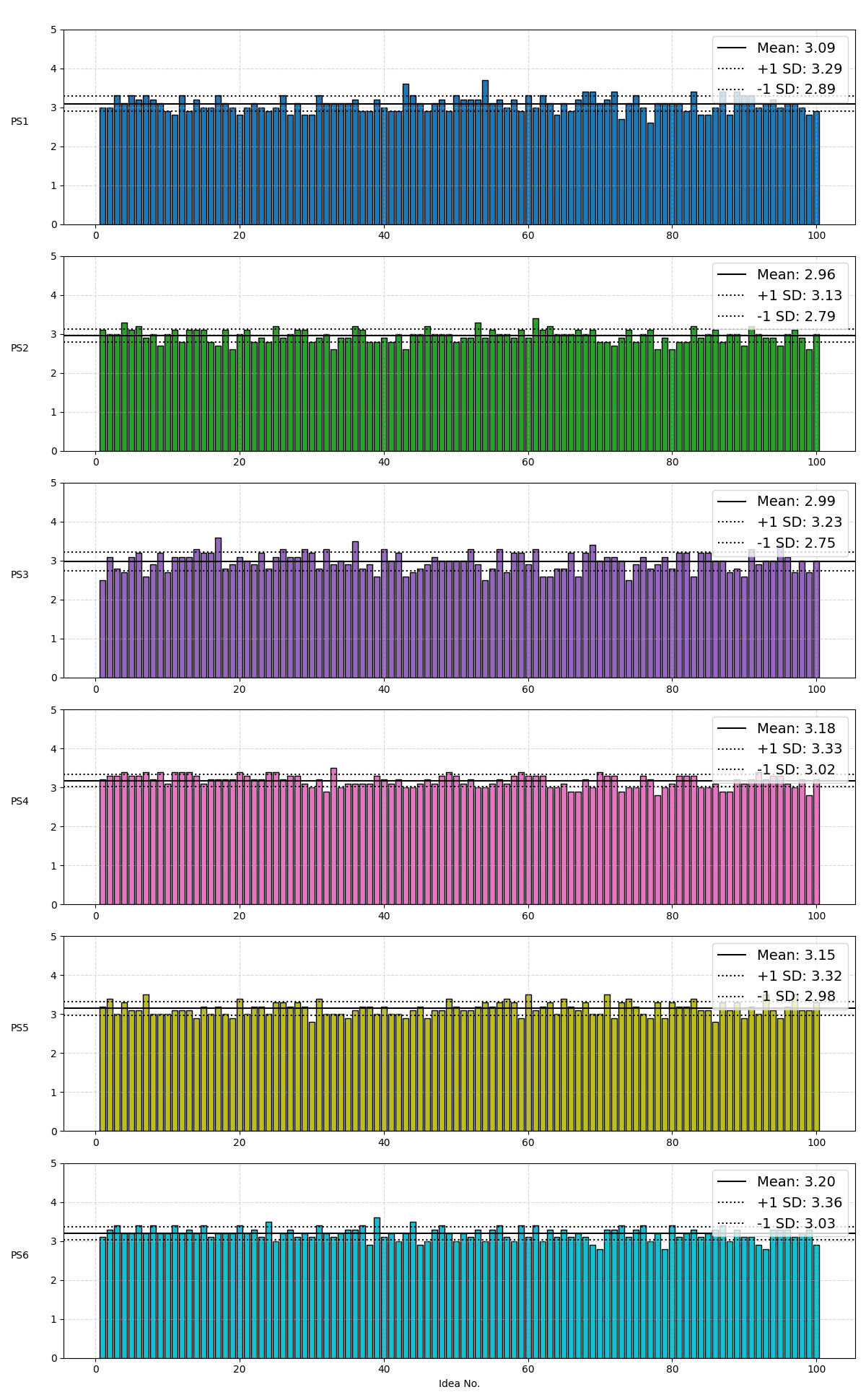}
        \caption{SemDis-based Assessment}
        \label{fig:hist_semdis}
    \end{subfigure}
    
    \caption{Bar plots of novelty ratings for 100 ideas across 6 problem statements}
    \label{fig:hist_ai_ass_methods}
\end{figure*}

\subsection{Box and Whisker Plots}
The box and whisker plots shown below provide a comparative summary of the distribution of novelty scores assigned by the different AI-based assessment methods. Key statistical features such as the mean, standard deviation, median, quartiles, and range are clearly depicted.

\begin{figure*}[h!]
    \centering
    \begin{subfigure}{\linewidth}
        \centering
        \includegraphics[width=0.5\linewidth]{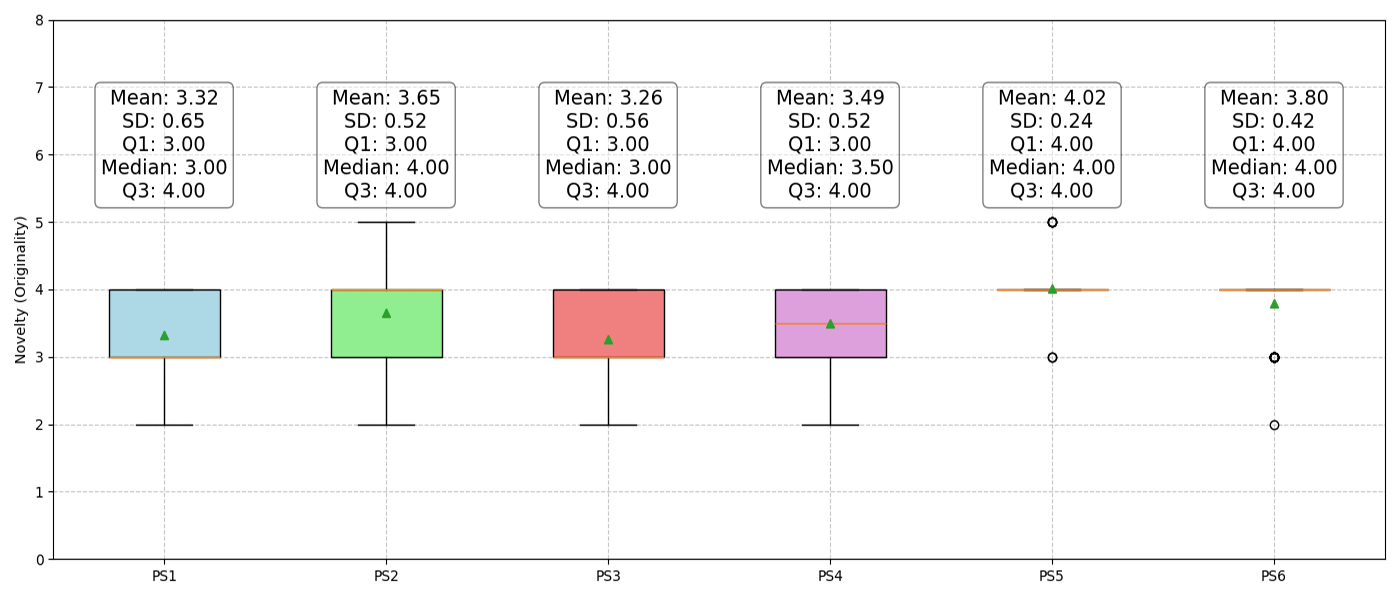}
        \caption{GPT-based Assessment}
        \label{fig:bw_gpt}
    \end{subfigure}%
    \hfill
    \begin{subfigure}{\linewidth}
        \centering
        \includegraphics[width=0.5\linewidth]{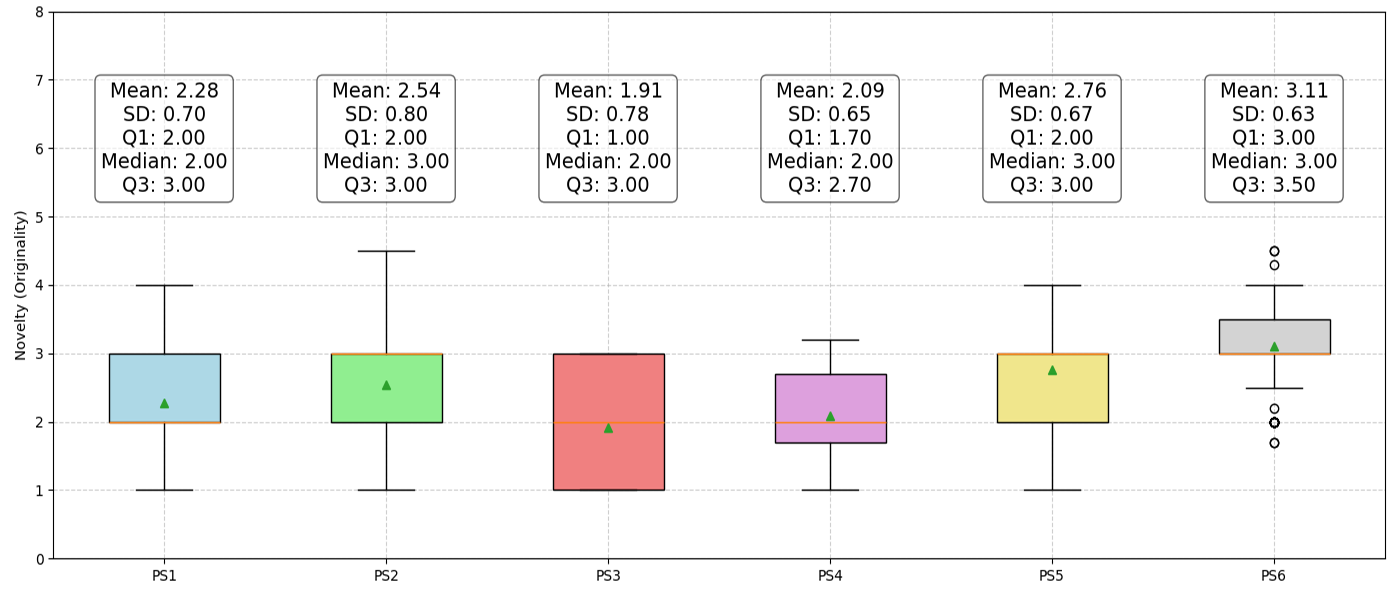}
        \caption{OCSAI-based Assessment}
        \label{fig:bw_ocsai}
    \end{subfigure}%
    \hfill
    \begin{subfigure}{\linewidth}
        \centering
        \includegraphics[width=0.5\linewidth]{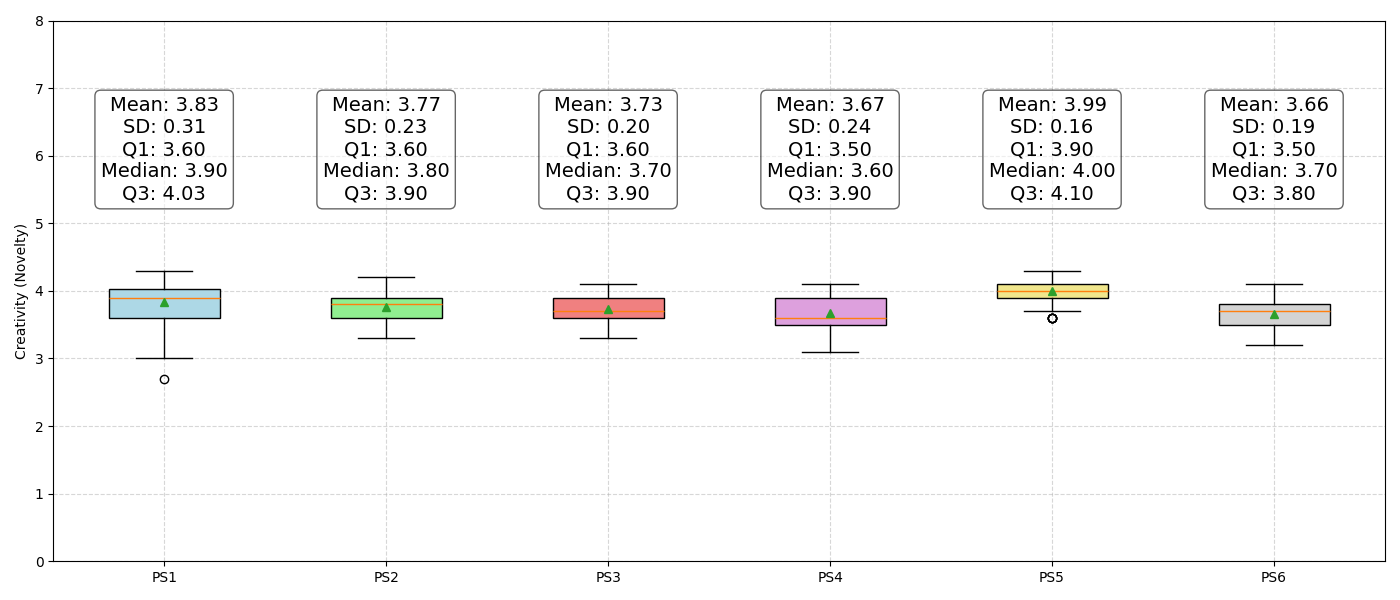}
        \caption{AIDE-based Assessment}
        \label{fig:bw_aide}
    \end{subfigure}%
    \hfill
    \begin{subfigure}{\linewidth}
        \centering
        \includegraphics[width=0.5\linewidth]{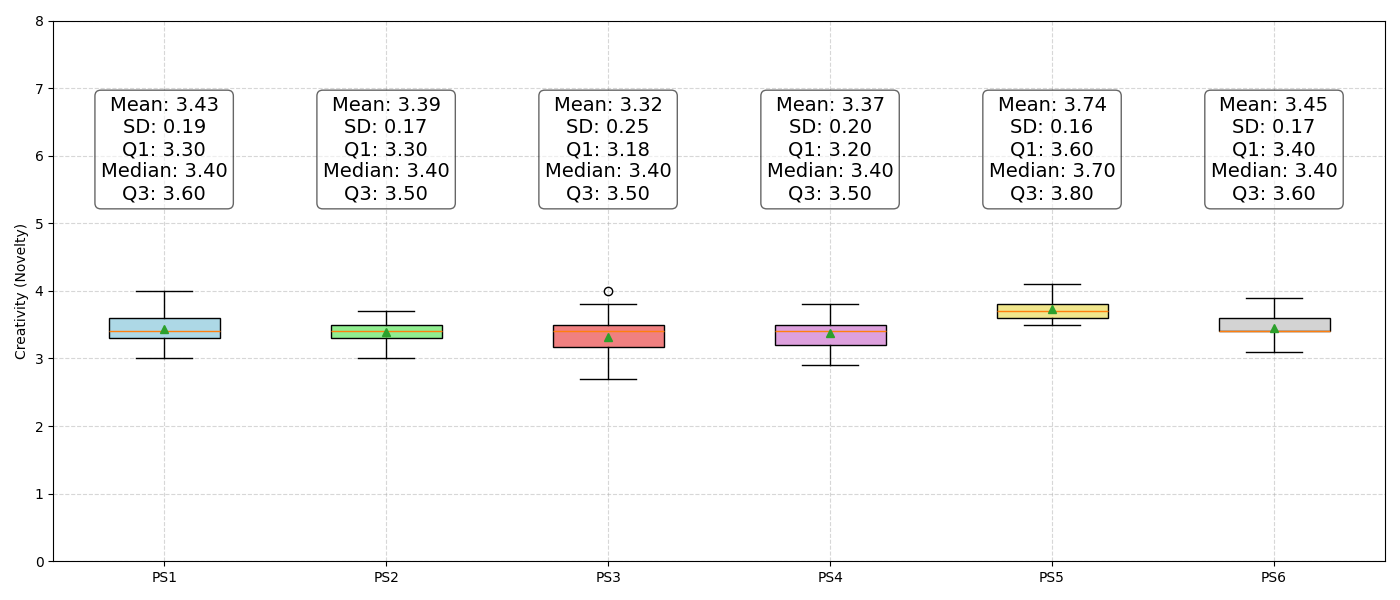}
        \caption{CLAUS-based Assessment}
        \label{fig:bw_claus}
    \end{subfigure}%
    \hfill
    \begin{subfigure}{\linewidth}
        \centering
        \includegraphics[width=0.5\linewidth]{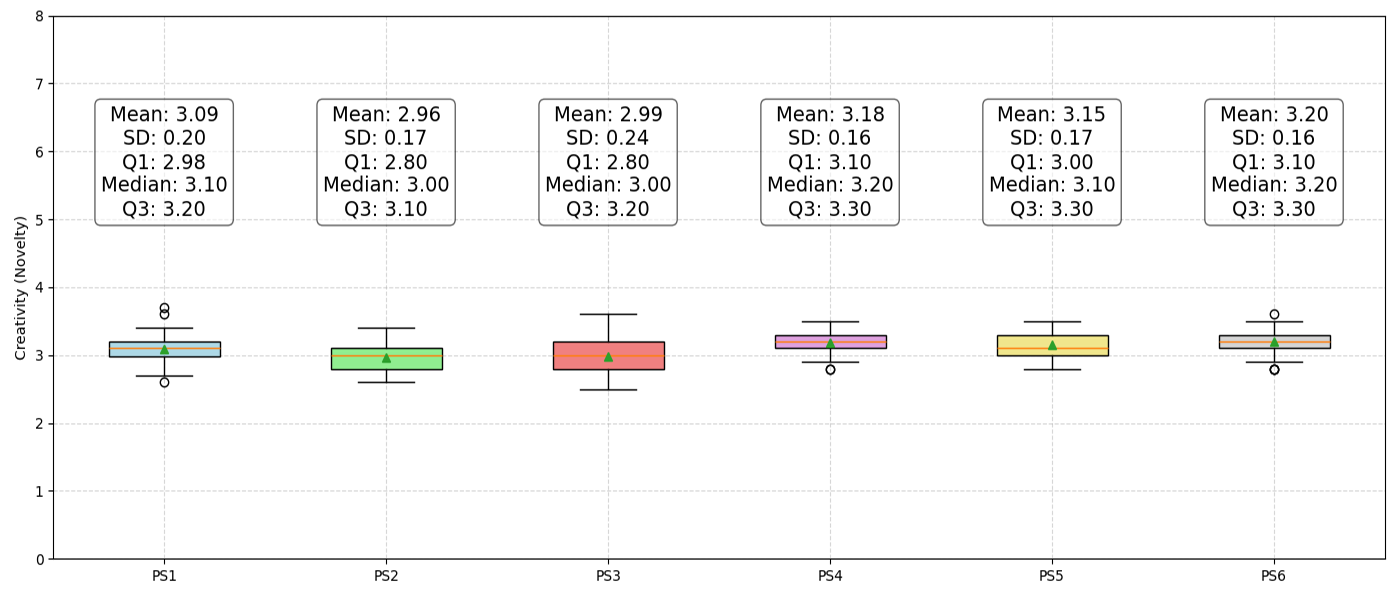}
        \caption{SemDis-based Assessment}
        \label{fig:bw_semdis}
    \end{subfigure}
    \caption{Box and whisker plots illustrating the novelty ratings generated by five distinct AI-based assessment methods.}
    \label{fig:bw_ai_ass_methods}
\end{figure*}

\end{document}